%% file: main.tex
\documentclass[10pt,twocolumn,letterpaper]{article}

\usepackage[pagenumbers]{iccv} 

\input{preamble}

\definecolor{iccvblue}{rgb}{0.21,0.49,0.74}
\usepackage[pagebackref,breaklinks,colorlinks,allcolors=iccvblue]{hyperref}
\usepackage{multirow}
\usepackage{xcolor}    
\usepackage{colortbl}
\usepackage{arydshln}
\usepackage{tabularx}
\newcommand{\blue}[1]{\textbf{\textcolor{mblue}{#1}}}
\usepackage{makecell}

\usepackage{pifont}
\definecolor{mblue}{RGB}{0, 77, 128}
\definecolor{darkgreen}{rgb}{0.0, 0.5, 0.0}
\definecolor{mycolor_blue}{HTML}{E7EFFA}
\definecolor{mycolor_green}{HTML}{E6F8E0}
\definecolor{mycolor_gray}{HTML}{ECECEC}
\newcommand{\cmark}{\textcolor{darkgreen}{\ding{51}}}  
\newcommand{\xmark}{\textcolor{red}{\ding{55}}}        
\def\paperID{1902} 
\def\confName{ICCV}
\def\confYear{2025}
\newcommand{\MYhref}[3][blue]{\href{#2}{\color{#1}{#3}}} 
\title{LMM4LMM: Benchmarking and Evaluating Large-multimodal \\Image Generation with LMMs}

\begin{document}
\twocolumn[{
\vspace{-2em}
\author{
Jiarui Wang$^{1}$,    
Huiyu Duan$^{1,2}$, 
Yu Zhao$^{1}$,
Juntong Wang$^{1}$,
Guangtao Zhai$^{1,2}$,
Xiongkuo Min$^{1*}$,
\\
$^{1}$Institute of Image Communication and Network Engineering, \\
$^{2}$ MoE Key Lab of Artificial Intelligence, AI Institute,\\ Shanghai Jiao Tong University, Shanghai, China\\}
\maketitle
\vspace{-2em}
\input{figures/abstr}
}]
\footnotetext{$^{*}$Corresponding Author}
\input{sec/0_abstract}    
\input{sec/1_intro}

\input{sec/2_related}
\input{sec/3_benchmark}
\input{sec/4_method}

\input{sec/5_experiment}
\input{sec/6_conclusion}

{
    \small
    \bibliographystyle{ieeenat_fullname}
    \bibliography{main}
}
\newpage
\clearpage
\newpage
\appendix
\twocolumn[{\section*{Appendix }}\vspace*{5mm}]

\input{sec2/1_overview}

\input{sec2/2_prompt}

\input{sec2/3_models}
\input{sec2/4_exp}
\input{sec2/5_analyze}
\input{sec2/6_loss}
\input{sec2/7_methods}
\input{sec2/8_results}
\input{sec2/9_figures}

\end{document}

%% file: preamble.tex
\newcommand{\red}[1]{{\color{red}#1}}


%% file: figures/abstr.tex
\centering
\includegraphics[width=1\linewidth]{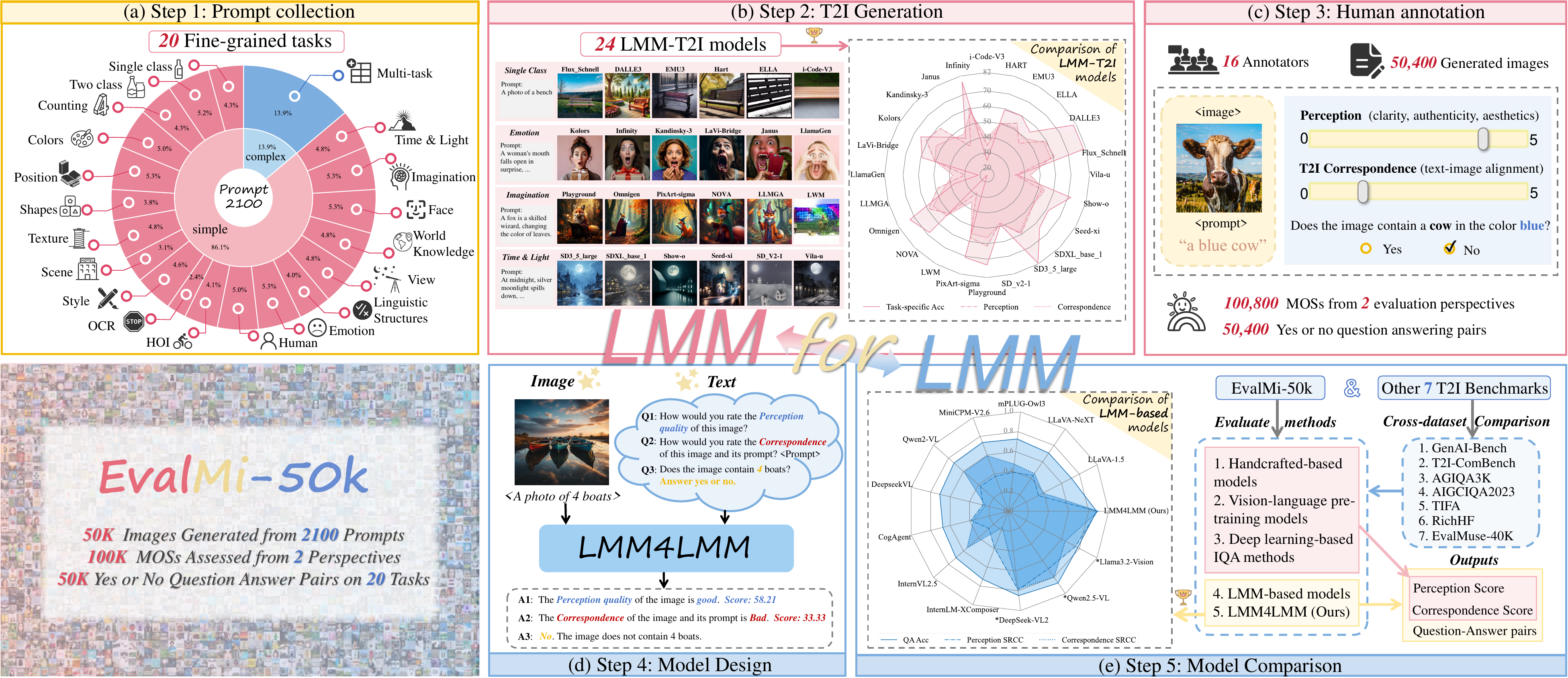}
\vspace{-2em}
\captionof{figure}{
We present the large multimodal image generation evaluation database and model, termed EvalMi-50K and LMM4LMM, respectively.
(a) We first collect 2100 comprehensive prompts across 20 fine-grained tasks.
(b) Then 24 LMM-T2I models are applied to generate 50K images.
(c) 100K MOSs and 50K question-answering pairs are acquired from 16 annotators.
(d) We design LMM4LMM to evaluate LMM-T2I models.
(e) We conduct model comparisons on EvalMi-50K and the other 7 benchmarks.
}
\label{abstract}
\vspace{2em}

%% file: sec/0_abstract.tex
\begin{abstract}
Recent breakthroughs in large multimodal models (LMMs) have significantly advanced both text-to-image (T2I) generation and image-to-text (I2T) interpretation.
However, many generated images still suffer from issues related to perceptual quality and text-image alignment. Given the high cost and inefficiency of manual evaluation, an automatic metric that aligns with human preferences is desirable.
To this end, we present \textbf{EvalMi-50K}, a comprehensive dataset and benchmark for \underline{eval}uating large-\underline{m}ultimodal \underline{i}mage generation,
which features \textbf{(i) comprehensive tasks}, encompassing 2,100 extensive prompts across 20 fine-grained task dimensions, and \textbf{(ii) large-scale human-preference annotations}, including 100K mean-opinion scores (MOSs) and 50K question-answering (QA) pairs annotated on 50,400 images generated from 24 T2I models.
Based on EvalMi-50K, we propose \textbf{LMM4LMM}, an \underline{LMM}-based metric for evaluating \underline{l}arge \underline{m}ulti\underline{m}odal T2I generation from multiple dimensions including perception, text-image correspondence, and task-specific accuracy.
Extensive experimental results show that LMM4LMM achieves state-of-the-art performance on EvalMi-50K, and exhibits strong generalization ability on other AI-generated image evaluation benchmark datasets, manifesting the generality of both the EvalMi-50K dataset and LMM4LMM metric.
Both EvalMi-50K and LMM4LMM will be released at 
\MYhref[magenta]{https://github.com/IntMeGroup/LMM4LMM}{https://github.com/IntMeGroup/LMM4LMM}.
\end{abstract}
\vspace{-3mm}

%% file: sec/1_intro.tex
\section{Introduction}
\label{sec:intro}
\input{tabels/database}

The rapid advancement of large multimodal models (LMMs) has revolutionized the fields of both text-to-image (T2I) generation \cite{betker2023improving,xiao2024omnigen,xie2024show} and image-to-text (I2T) interpretation \cite{liu2023visual,liu2024improved,chen2024expanding}, leading to high-quality AI-generated images (AIGIs) and comprehensive multimodal understanding capabilities.
However, state-of-the-art T2I models may still generate images struggling with perceptual quality and text-image correspondence, thus failing to satisfy human preferences \cite{wang2023aigciqa2023,xu2023imagereward,liang2024rich,han2024evalmuse40kreliablefinegrainedbenchmark}.
Since human evaluation is expensive and inefficient, it is of great significance to develop reliable evaluation metrics that align well with human perception and preference. 

Traditional image quality assessment (IQA) methods \cite{mittal2012making,xue2013learning,kang2014convolutional,su2020blindly,sun2023blind} generally focus on natural images with in-the-wild distortions such as noise, blur, compression \cite{zhai2020perceptual,min2019quality,min2019quality,duan2024finevq}, \textit{etc.}, while ignoring the unique distortions in AIGIs including unrealistic structures, unnatural textures, and text-image inconsistencies \cite{liang2024rich,wang2023aigciqa2023,wang2024quality}.
AIGI evaluation metrics such as Inception Score (IS) \cite{gulrajani2017improved} and Fréchet Inception Distance (FID) \cite{heusel2017gans} cannot evaluate the authenticity of a single image, and cannot take prompts into consideration \cite{wang2024quality}.
Other common metrics such as CLIPScore \cite{hessel2021clipscore} show less alignment with human preferences \cite{ghosh2023geneval}.
As shown in Table \ref{tab:relate}, some recent works such as AGIQA \cite{li2023agiqa} and AIGCIQA2023 \cite{wang2023aigciqa2023} have studied fine-grained mean opinion score (MOS) evaluation for AIGIs, however, the dataset scale or dimension scale is still relatively small.
In addition, the text-image correspondence scores in these works may be affected by the perceptual quality, while they lack task-specific accuracy annotations, which are essential for benchmarking T2I models \cite{ghosh2023geneval}.
Other studies such as GenEval \cite{ghosh2023geneval} and EvalMuse-40K \cite{han2024evalmuse40kreliablefinegrainedbenchmark} have T2I correspondence or task-specific accuracy annotations, but they lack consideration of the perceptual quality dimension and provide limited score annotations per image (about 3-6 per image), which may limit the model generality.

In this paper, we present \textbf{EvalMi-50K}, a large-scale dataset and benchmark towards better \underline{\textbf{eval}}uation of large-\underline{\textbf{m}}ultimodal \underline{\textbf{i}}mage generation, which includes 50,400 images generated by 24 state-of-the-art T2I models using 2,100 diverse prompts across 20 task-specific challenges.
As shown in Figure \ref{abstract}, we collect \textbf{2M+} human annotations from the perception, text-image correspondence, and task-specific accuracy, respectively, and finally obtain 100,800 MOSs and 50,400 question-answering (QA) pairs.
Based on EvalMi-50K, we propose \textbf{LMM4LMM}, a \underline{LMM}-based metric for evaluating \underline{l}arge \underline{m}ulti\underline{m}odal T2I generation from multiple dimensions including perceptual quality, text-image correspondence, and task-specific accuracy, respectively.
Specifically, LMM4LMM adopts an LMM as the backbone and leverages instruction tuning \cite{liu2023visual} techniques by training the visual-language projector to give the right answers.
To extract quality related and text-image aligned features and further refine these features, we apply LoRA adaptation \cite{hulora} to both the vision encoder and the large language model, respectively.
Through extensive experimental validation, we demonstrate that LMM4LMM achieves state-of-the-art performance on the EvalMi-50K dataset and manifests strong zero-shot generalization ability on other benchmarks.
The main highlights of this work include:

\begin{itemize}
    \item We introduce EvalMi-50K, a large-scale dataset that contains 50,400 multimodal generated images with 2M+ subjective ratings from the perception, text-image correspondence, and task-specific accuracy, respectively.

    \item We also use EvalMi-50K to benchmark the ability of LMMs in evaluating the generated images. EvalMi-50K can not only be used to evaluate the \textit{generation ability} of large multimodal (LMM) T2I models, but also the \textit{interpretation ability} of large multimodal models (LMM).

    \item We propose LMM4LMM, a novel LMM-based evaluation model capable of both AIGI perception quality evaluation and T2I correspondence attribution.
    
    \item Extensive experimental results on EvalMi-50K and other AIGI benchmarks manifest the state-of-the-art performance and strong generalization ability of LMM4LMM.
\end{itemize}
It should be noted that LMM4LMM also conveys the concept that we can use LMM interpretation to assess LMM image generation ability, and vice versa use LMM image generation to assess LMM interpretation ability.

%% file: tabels/database.tex
\begin{table*}[tbph]
\centering
\vspace{-11mm}
\caption{Comparision of text-to-image model evaluation benchmarks and image quality evaluation databases. }
\vspace{-4mm}
\renewcommand\arraystretch{0.85}
\label{tab:relate}
\resizebox{0.9\textwidth}{!}{\begin{tabular}{cccccccccc}
\toprule
\textbf{Database} & MOS Granularity  & Images  & Annotations  & Models & T2I Tasks & People per MOS & Dimensions & QA Pairs   \\ \midrule
HPD~\cite{wu2023human} &No MOS&98,807& 98,807&1&3&N/A&Human Preference & \xmark\\
Pick-A-Pic~\cite{kirstain2023pick} &No MOS&10,000& 500,000&6&4&N/A&Human Preference & \xmark\\
\hdashline
TIFA~\cite{hu2023tifa}&Coarse-MOS&800& 1,600&5&12&2&T2I Correspondence & \cmark\\
GenEval~\cite{ghosh2023geneval}&Coarse-MOS&1,200& 6,000&6&6&5&T2I Correspondence & \cmark\\
T2I-CompBench~\cite{huang2023t2i}&Coarse-MOS&2,400& 7,200&6&8&3&T2I Correspondence & \xmark\\
GenAIBench~\cite{li2024genai}&Coarse-MOS&9,600& 40,000&6&8&3&T2I Correspondence & \cmark\\
%
RichHF~\cite{liang2024rich}&Coarse-MOS&18,000&  216,000 & 4&1&3&Plausibility, Alignment, Aesthetics, and Overall &\xmark  \\
EvalMuse-40K~\cite{han2024evalmuse40kreliablefinegrainedbenchmark}        & Coarse-MOS   & 40,000   & 1,000,000  & 20     & 12 & 3-6 & T2I Correspondence & \cmark        \\
\hdashline
AGIQA-1K~\cite{zhang2023perceptual}          & Fine-MOS   & 1,080   & 23,760  & 2      & 4 & 22 & Overall    &\xmark     \\
AGIQA-3K~\cite{li2023agiqa}          & Fine-MOS   & 2,982   & 125,244  & 6      & 5 & 21 & Perception and Alignment   &\xmark      \\
AIGIQA-20K~\cite{li2024aigiqa}              & Fine-MOS   & 20,000   & 420,000 & 15     & 1 & 21 & Overall   &\xmark      \\

AIGCIQA2023~\cite{wang2023aigciqa2023}          & Fine-MOS   & 2,400   & 48,000  & 6      & 10 & 20 & Quality, Authenticity and Correspondence   &\xmark     \\
 \rowcolor{gray!20} \textbf{EvalMi-50K (Ours)}           & \textbf{Fine-MOS}   & \textbf{50,400}  & \textbf{2,419,200}  & \textbf{24 }     & \textbf{20} & \textbf{16} & \textbf{ Perception and T2I Correspondence}   &\textbf{\cmark}    \\
\bottomrule
\end{tabular}}
\vspace{-5mm}
\end{table*}

%% file: sec/2_related.tex
\section{Related Works}\label{sec:related}
\subsection{Benchmarks for T2I Generation}

\input{figures/datamodel}
\input{figures/MOS}
As shown in Table \ref{tab:relate}, the development of T2I generation has spawned many T2I model evaluation benchmarks and AIGI IQA databases, which can be categorized into three groups based on the presence and granularity of the human Mean Opinion Scores (MOS).
No-MOS and coarse-MOS databases contain large datasets, with a limited number of annotators.
Fine-MOS databases offer more reliable assessments derived from more than 15 annotators, following the guidelines of ITU-R BT.500 \cite{series2012methodology}. HPD~\cite{wu2023human} and Pick-A-Pic~\cite{kirstain2023pick} focus on image pairs comparison, but lack precise quality assessment for each AIGI.
While TIFA~\cite{hu2023tifa}, GenAIBench~\cite{li2024genai}, and T2I-CompBench~\cite{huang2023t2i} focus on T2I correspondence, they overlook AIGI's visual perception. While AGIQA-3K~\cite{li2023agiqa}  and AIGCIQA2023~\cite{wang2023aigciqa2023} consider both perceptual quality and T2I correspondence, they lack task-specific QA pairs, limiting their ability to assess T2I generation across diverse tasks. EvalMi-50K stands out by providing fine-grained MOSs across both perceptual quality and T2I correspondence, along with task-specific QA pairs.
\subsection{Evaluation Metrics for T2I Generation}
Many image quality assessment models have been proposed in the literature, including handcrafted IQA models (\textit{e.g.}, NIQE~\cite{mittal2012making}, QAC~\cite{xue2013learning}, BRISQUE~\cite{mittal2012no}) and deep learning-based IQA models (\textit{e.g.}, CNNIQA~\cite{kang2014convolutional}, DBCNN~\cite{quality:DBCNN}, HyperIQA~\cite{su2020blindly}). These models characterize
quality-aware information to predict IQA scores but can not evaluate T2I correspondence, which is crucial for assessing the relationship between the generated image and its corresponding text prompt.
CLIPScore~\cite{hessel2021clipscore}, PickScore~\cite{kirstain2023pick}, and VQAScore~\cite{li2024evaluating} improve the evaluation of the T2I correspondence, but they struggle to assess the quality of image perception.
LMMs with visual understanding capabilities perform well in QA tasks but their ability to assess image perceptual quality remains limited and often fail to give precise quality scores. 
GenEval~\cite{ghosh2023geneval} and T2I-CompBench~\cite{huang2023t2i} employ various detection models for task-specific accuracy, but this approach is quite complex. 
To address this gap, our proposed LMM-based model complies with an \textit{\textbf{all-in-one}} manner, which predicts task-specific accuracies for all tasks using one model.

%% file: figures/datamodel.tex
\begin{figure*}
    \centering
         \vspace{-8mm}
    \includegraphics[width=1\textwidth]{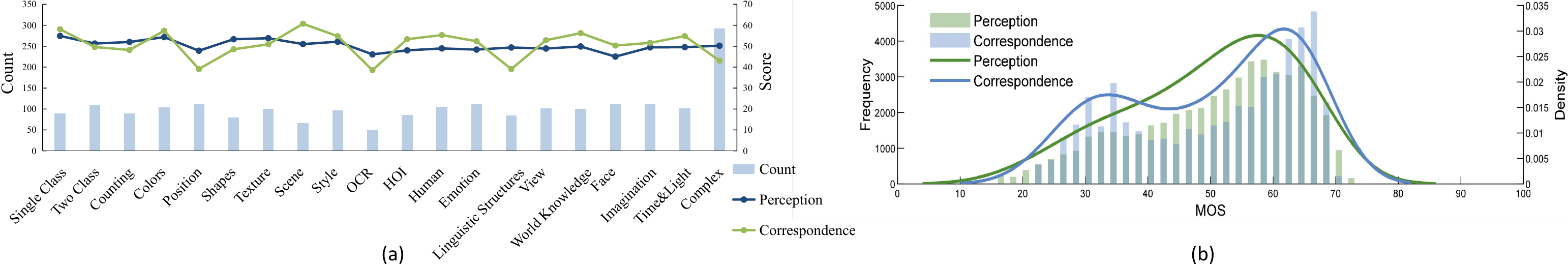}
     \vspace{-7mm}
    \caption{(a) Distribution of task counts and scores across different tasks. (b) Distribution of perception and correspondence MOSs.} 
    \label{distribution}
\end{figure*}

%% file: figures/MOS.tex
\begin{figure*}
    \centering
         \vspace{-2mm}
    \includegraphics[width=1\textwidth]{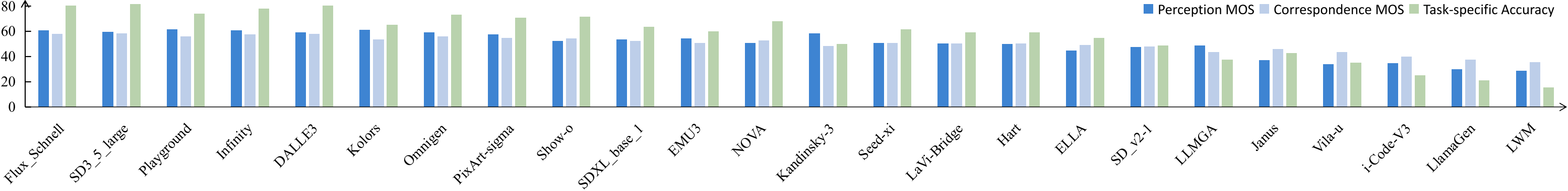}
     \vspace{-7mm}
    \caption{Comparison of T2I generation models regarding the perception MOSs, correspondence MOSs, and task-specific accuracy.} 
     \vspace{-2mm}
    \label{MOS}
\end{figure*}

%% file: sec/3_benchmark.tex
\section{EvalMi-50K Dataset \& Benchmark}

\subsection{Data Collection}

\vspace{-1mm}

Our prompt design focuses on 20 different tasks as shown in Figure \ref{abstract}(a). 
The complex tasks are designed by combining simpler task components, such as color, counting, and shape, into more complex challenges. 
The prompts are initially crafted based on the requirements of each task and then further refined using DeepSeek R1~\cite{guo2025deepseek} to expand and modify them, ensuring clarity and diversity. In total, we collect 2,100 prompts, each corresponding to a specific task. 
To generate the AIGIs, we utilize 24 of the latest LMM-T2I models, as shown in Figure \ref{abstract}(b).
We leverage open-source website APIs or the default weights of these models to generate images. For each prompt, each model generates a subset of images, and one of them is randomly selected from each model’s output. With 2,100 distinct prompts, this process results in a total of 50,400 images (24 models × 2,100 prompts).
More details of the database can be found in the \textit{supplementary material}.

\input{figures/leida}
\vspace{-1mm}
\subsection{Subjective Experiment Setup and Procedure}
Due to the unique distortions in AIGIs and varying elements determined by different text prompts, relying solely on an overall score for evaluation is inadequate. In this paper, we propose to evaluate AIGIs across two dimensions.
(1) \textbf{Perceptual quality} focuses on visual perception, evaluating factors such as detail richness, color vibrancy, distortion levels, and authenticity.
(2) \textbf{Text-image correspondence} evaluates how accurately the generated image reflects the objects, scenes, styles, and details described in the text prompt.
We use a 1-5 Likert scale to score the images based on the perception and T2I correspondence. For the correspondence evaluation, in addition to the rating, annotators are instructed to answer 20 task-specific yes/no questions to determine whether the image consistently aligns with the prompt.
 Finally, we obtain a total of 2,419,200 human annotations including 
 1,612,800 reliable score ratings (16 annotators $\times$ 2 dimensions $\times$ 50,400 images), and 806,400 task-specific QA pairs (16 annotators $\times$ 50,400 images).
 
\vspace{-1mm}
\subsection{Subjective Data Processing}
In order to obtain the MOS for an AIGI, we 
first convert the raw ratings into Z-scores, and then 
linearly scale them to the range $[0,100]$ as follows:
$$z_i{}_j=\frac{r_i{}_j-\mu_i{}_j}{\sigma_i},\quad z_{ij}'=\frac{100(z_{ij}+3)}{6},$$
$$\mu_i=\frac{1}{N_i}\sum_{j=1}^{N_i}r_i{}_j, ~~ \sigma_i=\sqrt{\frac{1}{N_i-1}\sum_{j=1}^{N_i}{(r_i{}_j-\mu_i{}_j)^2}},$$ 
where $r_{ij}$ is the raw rating given by the $i$-th subject to the $j$-th image. $N_i$ is the number of images judged by subject $i$. 
Next, the MOS of the $j$-th image is computed by averaging the rescaled z-scores across all subjects as follows:
$$MOS_j=\frac{1}{M}\sum_{i=1}^{M}z_{ij}',$$
where $MOS_j$ indicates the MOS for the $j$-th AIGI, $M$ is the number of subjects, and $z'_i{}_j$ are the rescaled z-scores. 
The task-specific accuracy is determined by the most votes.
Therefore, a total of 100,800 MOSs (2 dimensions $\times$ 50,400 images) and 50,400 question answering pairs are obtained.

\subsection{Subjective Data Analysis}
Figure \ref{distribution}(a) demonstrates the distribution of task counts and scores, highlighting the diversity and performance variations across different tasks.
Figure \ref{distribution}(b) illustrates the distribution of MOSs for both perceptual quality and T2I correspondence. 
We launch comparisons of LMM-T2I generation models based on perceptual quality MOSs, T2I correspondence MOSs, and task-specific accuracy, as shown in Figure \ref{MOS}. Kandinsky-3~\cite{arkhipkin2024kandinsky} excels in perceptual quality but performs poorly in correspondence, while NOVA~\cite{deng2024nova} exhibits the opposite trend. This contrast highlights the necessity of evaluating the perception and correspondence as separate dimensions.
We further analyze the MOSs and task-specific accuracies across different prompt categories. As shown in Figure
\ref{leidat}(a), perception MOS is particularly sensitive to tasks such as optical character recognition (OCR) and face, as high-quality images are crucial for accurately recognizing characters and face identifications. Figure \ref{leidat}(b) and (c) display similar trends in correspondence evaluations, with task-specific accuracy results exhibiting sharper distinctions. 
While task-specific accuracy provides binary (0/1) assessments, MOS offers continuous scoring, enabling more granular evaluation of T2I correspondence.
For tasks involving linguistic structures, most models perform poorly, suggesting that T2I models struggle to understand words such as “without” or “no”. Additionally, models show weak performance in tasks requiring position understanding, indicating that these models may not fully grasp spatial relationships or the positioning of objects within the scene. 
\input{figures/model}

%% file: figures/leida.tex
\begin{figure*}[!t]
    \centering
     \vspace{-4mm}
    \includegraphics[width=1\textwidth]{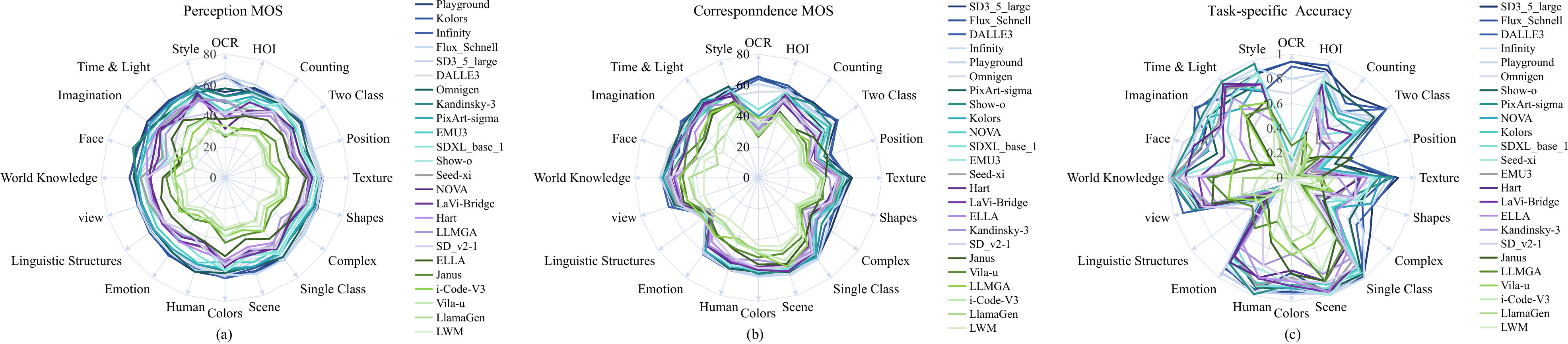}
     \vspace{-7mm}
    \caption{Comparison of MOSs and task-specific accuracy of 24 generation models across 20 tasks with descending order arranged in legend. (a) Results across perception MOSs. (b) Results across correspondence MOSs. (c) Results across task-specific accuracy. } 
     \vspace{-2mm}
    \label{leidat}
\end{figure*}

%% file: figures/model.tex
\begin{figure*}
    \centering
    \vspace{-5mm}
    \includegraphics[width=1\textwidth]{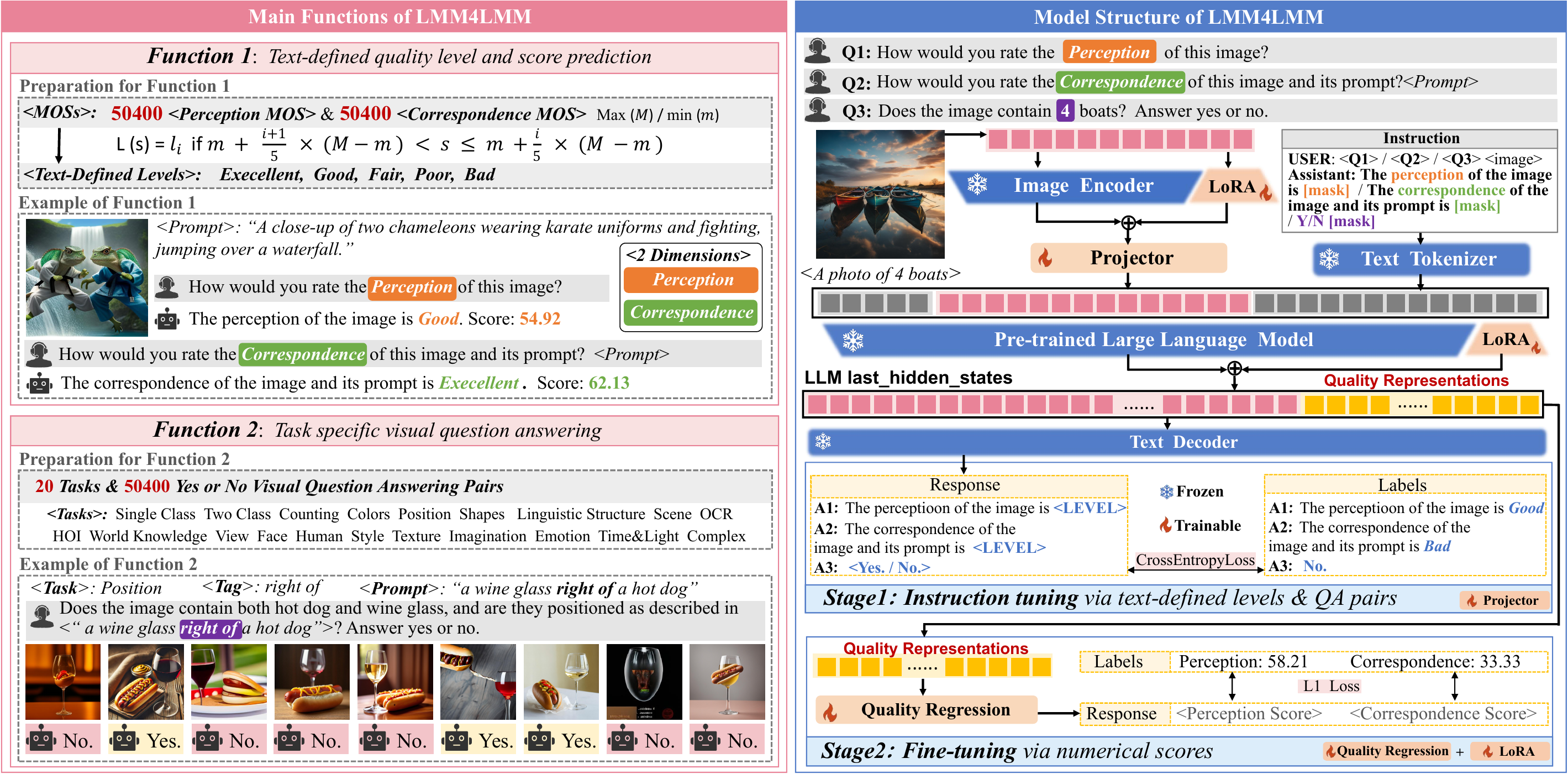}
    \vspace{-6mm}
    \caption{Overview of the LMM4LMM architecture. The model includes two functions: (1) text-defined quality level and score prediction, (2) task-specific visual question answering. The training process consists of two stages: instruction tuning of the model via text-defined levels, and then fine-tuning the vision encoder and LLM via numerical scores. The model incorporates an image encoder and a text encoder for extracting visual and
textual features, which are fed into a pre-trained LLM to generate results. LoRA  \cite{hulora} weights are introduced to the pre-trained image encoder and the LLM to adapt the models to perception quality evaluation and T2I correspondence attribution tasks.} 
     \vspace{-1mm}
    \label{modell}
\end{figure*}

%% file: sec/4_method.tex
\section{The LMM4LMM Approach}
In this section, we introduce our \textit{\textbf{all-in-one}} image quality assessment method, \textbf{LMM4LMM}, towards giving text-defined quality levels,
predicting perception and T2I correspondence scores, and providing visual question answers for its correspondence assessments, depicting quality attributes from 20 task-specific challenges using one model.

\subsection{Model Structure}
\paragraph{Visual Encoding.}
As shown in Figure \ref{modell}, the visual encoding part includes an image encoder for feature extraction and a projector for feature alignment between the image features and the input of the large language model (LLM).
To enhance scalability for processing high-resolution images, we employ a pixel unshuffle operation, which reduces the number of visual tokens to one-quarter of the original size.
Specifically, for an input AIGI $I$, we first resize it to 1024×1024 and then
 divide images into tiles of 448×448 pixels based on the aspect ratio and resolution of the input images.
 The image encoder $E_I$ is built on a pre-trained vision transformer (ViT), \textit{i.e.}, InternViT \cite{chen2024expanding}, which is pre-trained on the LAION-en dataset \cite{schuhmann2022laion} using text-image contrastive learning.
To align the extracted features with the input space of the LLM, a projector $P_I$ with two multilayer perceptron (MLP) layers is applied.
The process can be formulated as:
\begin{equation}
    T_i = P_I(E_I(I)),
\end{equation}
where $T_i$ is the mapped image feature tokens.
\vspace{-2mm}
\paragraph{Feature Fusion and Quality Regression.}
We utilize the LMM (InternVL2.5-8B~\cite{chen2024expanding}) to integrate  the visual tokens and text instruction tokens to perform the following two tasks.
(1) Quality level descriptions: the model generates a descriptive quality level evaluation of the input image, such as “\textit{The perception quality of the image is (bad, poor, fair, good, excellent)}.”
Since LLMs have a better understanding of textual data than numerical data, this initial categorization provides a preliminary classification of the image's quality, which is valuable for guiding subsequent quality regression tasks. 
(2) Regression score output: the model takes the quality representations from the last hidden states of the LLM to perform a regression task through a quality regression module, outputting numerical quality scores.

\subsection{Training and Fine-tuning Strategy}
The training process of LMM4LMM follows a two-stage approach to address two tasks: (1) perception quality and text-image correspondence score prediction, (2) task-specific visual question answering.  We first perform instruction tuning via text-defined quality levels and QA pairs. We then fine-tune the vision encoder and LLM with LoRA \cite{hulora}, and train the quality regression module via numerical scores to enable accurate score generation.
\vspace{-2mm}
\paragraph{Instruction Tuning.}
Achieving an \textit{\textbf{all-in-one}} image quality assessment model is of great significance for enabling multi-dimensional quality evaluation in a single model. 
Benefiting from the generalization ability of LMMs, our model verifies the effectiveness of using the instruction tuning strategy for all-in-one task-specific question answering. 
We train the projector to align textual and visual semantics for joint reasoning and then use language loss during the instruction tuning phase. 
As a result, LMM4LMM can give visual question answers across the 20 task-specific challenges using one model weight.
For score prediction, since LMMs have a better understanding of textual data than numerical data, directly generating numerical scores might be challenging for LMMs.
Therefore, we first convert the continuous scores into categorical text-based quality levels. 
Specifically, we uniformly divide the range between the highest score ($\mathrm{M}$) and the lowest score ($\mathrm{m}$) into five distinct intervals, assigning the scores in each interval to respective levels:
\begin{equation}
    {L(s)} = l_i \text{  if } \text{m} + \frac{i-1}{5}  \times \mathrm{(M-m)} < s \leq \mathrm{m} + \frac{i}{5} \times \mathrm{(M-m)},
\end{equation}
where \{$l_i|_{i=1}^{5}\}=\{\textit{bad, poor, fair, good, excellent}\}$ are the standard text rating levels as defined by ITU~\cite{series2012methodology}. This step provides the LMM with a more accessible way to grasp the concept of image quality by initially framing it in terms of text-defined quality levels.
\vspace{-2mm}
\paragraph{Quality Regression Fine-tuning.}

To further improve the performance of LMM4LMM and enable it to produce more precise quality scores, we introduce a quality regression module, which takes the last-hidden-state features from the LMM as input and generates scores from both perception quality and T2I correspondence perspectives.
Fine-tuning LMMs is generally resource consuming but can lead to better performance.
To make the fine-tuning process more efficient, we adopt the LoRA technique \cite{hulora}.
The LoRA-based approach ensures that the model adapts effectively to the regression task with numerical scores to adjust the model’s predictions and produce more accurate, fine-grained IQA results. During the fine-tuning stage, we employ L1 loss for the quality regression task to minimize the difference between the predicted scores and the groundtruth values.


%% file: sec/5_experiment.tex
\section{Experiments}

In this section, we conduct extensive experiments to evaluate the performance of our proposed model.
We first present the experimental setups in detail. Then we launch experiments to evaluate the performance of our model compared to current state-of-the-art IQA and LMM-based models in predicting scores and task-specific visual question answering based on EvalMi-50K and other seven AI-generated image evaluation datasets.
We launch further cross-dataset experiments to verify the generalizability of the proposed model.
Finally, we conduct ablation experiments to evaluate the efficiency of our proposed components. 

\subsection{Experiment Setup}
To evaluate the correlation between the predicted scores and the ground-truth MOSs, we utilize three evaluation criteria: Spearman Rank Correlation Coefficient (SRCC), Pearson Linear Correlation Coefficient (PLCC), and Kendall’s Rank Correlation Coefficient (KRCC).  For visual question answering, we adopt the average accuracy as the metric.
Traditional handcrafted IQA models are directly evaluated on the corresponding databases.
We load the pre-trained weights for inference for vision-language pre-training and LLM-based models. We fine-tuned three of the LLM-based models using the same fine-tuning approach as our model's backbone. 
For deep learning-based models, we use the same training and testing split (4:1) as the previous literature. 
The models are implemented with PyTorch and trained on a 40GB NVIDIA RTX A6000 GPU with batch size of 8. The initial learning rate is set to 1e-5, and decreased using the cosine annealing strategy. We employ Adam optimizer with $\beta_1 =0.9$ and $\beta_2 = 0.999$. During pre-training, the
number of training epochs is set to 1. For fine-tuning, the
number of training epochs is set to 5.
All experiments for each method are averaged using 5-fold cross-validation.

 \subsection{Evaluation on the EvalMi-50K Database}
   \input{tabels/SRCC2}
  \input{tabels/SRCC_ACC}
\input{tabels/model}

\input{figures/leida22}
\input{figures/example}
\input{tabels/cross}
\input{tabels/ablation}
 As shown in Table \ref{mos}, handcrafted IQA models such as NIQE~\cite{mittal2012making} and QAC \cite{xue2013learning}, show poor performance, indicating their features handcrafted mainly for natural images are ineffective for evaluating AIGIs. 
Vision-language pre-training models such as CLIPScore~\cite{hessel2021clipscore} and PickScore~\cite{kirstain2023pick} perform poorly in perception quality due to their focus on T2I correspondence and overlook AIGI's visual perception.  While LMM-based models are effective in handling complex visual question-answering tasks, their interpretation of image perception quality remains insufficient. 
Deep learning-based IQA methods achieve relatively better results, but still fall short in the T2I correspondence dimension.
Table \ref{lmm} and Figure \ref{leidat22} compare the performances of LMM-based models across the 20 task-specific challenges derived from the EvalMi-50K dataset.
LMMs excel in tasks that require the interpretation of complex visual-textual interactions, such as OCR and World Knowledge, but they struggle with low-level quality features, such as texture and style, as their focus is on semantic understanding rather than perceptual quality. When fine-tuned using our proposed methods, their performance improves significantly, which verifies the effectiveness of our approach in enhancing the evaluation and interpretation capabilities of the LMMs. 
Our model achieves superior performance in both score prediction and visual question answering, making it a more comprehensive method for evaluating AIGIs.

 \subsection{Evaluation on T2I Model Performance}
We further conduct comparisons of the alignment between different metric results and human annotations in evaluating T2I model performance, as shown in Table \ref{model}. 
Our model achieves the highest SRCC with human ratings and the lowest relative Root Mean Square Error (RMSE) in score differences. This demonstrates our model's ability to accurately assess and rank the performance of T2I generative models closest to human judgment. We also provide examples with model prediction scores at the image level. As shown in Figure \ref{example}, LMM4LMM generates scores that are more consistent with human annotations and achieves the highest accuracy in question-answering, which further demonstrates its effectiveness in both image perception evaluation and task-specific T2I correspondence attribution.

\subsection{Zero-shot Cross-dataset Evaluation}
\vspace{-2mm}
As shown in Table \ref{tab:method}, we present zero-shot cross-dataset performance comparisons on multiple benchmark. LMM4LMM achieves the best performance on EvalMi-50K and other 7 AIGI evaluation benchmarks. To further validate the generalization capability of our approach, we fine-tuned our method on EvalMuse-40K~\cite{han2024evalmuse40kreliablefinegrainedbenchmark}. The results demonstrate that fine-tuning on EvalMuse-40K yields slightly lower generalization, likely due to the scoring in EvalMuse-40K is coarser compared to our dataset, which highlights the importance of fine-grained MOS annotations in improving the model's generalization abilities.

\subsection{Ablation Study}
To validate the contributions of the different modules in LMM4LMM, we conduct comprehensive ablation studies, with results summarized in Table \ref{ablation}. 
Our analysis reveals three key findings: First, experiments (1), (2), and (5) demonstrate the effectiveness of quality-level initialization in model performance. Second, through experiments (3)-(7), we validate the significant performance gains achieved by LoRA fine-tuning. Third, experiments (7)-(10) compare different backbone architectures, confirming the effectiveness of our combined approach, which leverages the right balance of modules and model architecture to achieve state-of-the-art performance in IQA. 


%% file: tabels/SRCC2.tex
\begin{table}[t]
\renewcommand\arraystretch{1}
\caption{Performance comparisons of the state-of-the-art quality evaluation methods on the EvalMi-50K from perspectives of perception and T2I correspondence. $\spadesuit$ Handcrafted IQA models, $\diamondsuit$ vision-language pre-training models, $\clubsuit$ LMM-based models,  $\heartsuit$ deep learning-based IQA models. *Refers to finetuned models. 
}
  \label{t2}
  \centering
\vspace{-3mm}

\resizebox{0.47\textwidth}{!}{
  \begin{tabular}{l||ccc|ccc}
  \Xhline{1px}
    \multicolumn{1}{l}{\bf Dimension} &
    \multicolumn{3}{c}{\textbf{Perception}} &
    \multicolumn{3}{c}{\textbf{Correspondence}} 
    \\
    \cmidrule(lr){2-4}\cmidrule(lr){5-7}
\textbf{Methods / Metrics}  
& \textbf{SRCC}  & \textbf{PLCC}  & \textbf{KRCC}    & \textbf{SRCC}  & \textbf{PLCC}  & \textbf{KRCC} \\

\hline
$\spadesuit$ \textbf{NIQE} \cite{mittal2012making}
 & 0.3818 & 0.3885 & 0.2589 
 & 0.2430 & 0.2505 & 0.1643  

\\
$\spadesuit$ \textbf{QAC} \cite{xue2013learning} 
& 0.0376  & 0.0855 & 0.0246 
& 0.0511  & 0.0680 & 0.0337 

\\
$\spadesuit$ \textbf{BRISQUE} \cite{mittal2012no} 
 & 0.0157 & 0.0334 & 0.0104 
& 0.0467 & 0.0543 & 0.0313 

\\
$\spadesuit$ \textbf{BPRI}  \cite{min2017blind}
& 0.0329 & 0.0196 & 0.0207 
& 0.0068 & 0.0022 & 0.0045 

\\
 
$\spadesuit$ \textbf{HOSA} \cite{xu2016blind} 
 & 0.1480 & 0.1690 & 0.0985
 & 0.1355 & 0.1471 & 0.0905 

\\
$\spadesuit$ \textbf{BMPRI} \cite{quality:BMPRI} 
& 0.1519 & 0.1245 & 0.1011
& 0.0611 & 0.0415 & 0.0410

\\

$\spadesuit$ \textbf{Higrade-2} \cite{kundu2017large}& 0.0393 & 0.0260 & 0.0275 & 0.0326  & 0.0224  & 0.0223\\
\hline
$\diamondsuit$ \textbf{CLIPScore}~\cite{hessel2021clipscore}
& 0.2031 & 0.2561 & 0.1369 
& 0.2607 & 0.3072 & 0.1772 

\\
$\diamondsuit$ \textbf{BLIPScore}~\cite{li2022blip}
& 0.1575 & 0.2166 & 0.1060
& 0.2900 & 0.3468 & 0.1970 

\\

$\diamondsuit$ \textbf{ImageReward}~\cite{xu2023imagereward}
& 0.4105 & 0.4676 & 0.2815 
& 0.4991 & 0.5523 & 0.3470

\\
$\diamondsuit$ \textbf{PickScore}~\cite{kirstain2023pick}
& 0.5623 & 0.5905 & 0.3939 
& 0.4611 & 0.4692 & 0.3214 
\\
$\diamondsuit$ \textbf{HPSv2}~\cite{wu2023human}
& 0.6404 & 0.6751 & 0.4556 
& 0.5336 & 0.5525 & 0.3747 
\\
$\diamondsuit$ \textbf{VQAScore}~\cite{li2024evaluating}
& 0.3314 & 0.3172 & 0.2253
& 0.6062 & 0.6118 & 0.4304 
\\
$\diamondsuit$ \textbf{FGA-BLIP2}~\cite{han2024evalmuse40kreliablefinegrainedbenchmark}
& 0.5275 & 0.5604 & 0.3694 
& 0.6755 & 0.6916 & 0.4901 
\\
\hline
$\clubsuit$ \textbf{LLaVA-1.5 (7B)}~\cite{liu2024improved}
& 0.3372 & 0.3525 & 0.2577 & 0.3887 & 0.3716 & 0.3149
\\
$\clubsuit$ \textbf{LLaVA-NeXT (8B)}~\cite{liu2024llavanext}
& 0.4333 & 0.4164 & 0.3442 & 0.4568 & 0.4803 & 0.3535

\\
$\clubsuit$ \textbf{mPLUG-Owl3 (7B)}~\cite{ye2024mplug}
& 0.3918 & 0.3569 & 0.3018 & 0.4744 & 0.5430 & 0.3657
\\
$\clubsuit$ \textbf{MiniCPM-V2.6 (8B)}~\cite{yao2024minicpm}
& 0.3733 & 0.1053 & 0.2839 & 0.5916 & 0.5971 & 0.4597
\\
$\clubsuit$ \textbf{Qwen2-VL (7B)}~\cite{Qwen2-VL}
& 0.3760 & 0.3625 & 0.3061 & 0.5899 & 0.5954 & 0.4658
\\

$\clubsuit$ \textbf{DeepSeekVL (7B)}~\cite{wu2024deepseekvl2mixtureofexpertsvisionlanguagemodels}
& 0.2611 & 0.3010 & 0.1988
& 0.2356 & 0.3457 & 0.1872
 
\\

 
$\clubsuit$ \textbf{CogAgent (18B)}~\cite{hong2024cogagentvisuallanguagemodel}
& 0.3861 & 0.4235 & 0.2927 & 0.3575 & 0.3601 & 0.2888
\\
$\clubsuit$ \textbf{InternVL2.5 (8B)}~\cite{chen2024expanding}
& 0.2597 & 0.3669 & 0.1859 & 0.5511 & 0.5908 & 0.4039
\\
$\clubsuit$ \textbf{InternLM-XComposer (7B)}~\cite{internlmxcomposer}
& 0.3918 & 0.3569 & 0.3018 & 0.1728 & 0.1659 & 0.1401
\\
 $\clubsuit$ \textbf{DeepSeekVL2 (1B)*}~\cite{wu2024deepseekvl2mixtureofexpertsvisionlanguagemodels}
 & 0.7899 & 0.8253 & 0.6511
& 0.7817 & 0.7991 & 0.6457
\\
 $\clubsuit$ \textbf{Qwen2.5-VL (8B)*}~\cite{Qwen2.5-VL}
& 0.6990 & 0.7495 & 0.5715 & \textbf{\blue{0.8008}} & \textbf{\blue{0.8219}} & \textbf{\blue{0.6657}}
\\
$\clubsuit$ \textbf{Llma3.2-Vision (11B)*}~\cite{meta2024llama}
& 0.7555 & 0.7891 & 0.6155
& 0.6403 & 0.6461 & 0.5168

\\
\hline
$\heartsuit$ \textbf{CNNIQA*}~\cite{kang2014convolutional} 
& 0.4348 & 0.5583 & 0.3383 
& 0.1186 & 0.0791 & 0.1067

\\

$\heartsuit$ \textbf{DBCNN*}~\cite{quality:DBCNN}
& 0.5525 & 0.6181 & 0.3802 
& 0.3301 & 0.3515 & 0.2216 

\\
$\heartsuit$ \textbf{HyperIQA*}~\cite{su2020blindly}
& 0.5872 & 0.6768 & 0.4335 
& 0.5348 & 0.5447 & 0.3742

\\

$\heartsuit$ \textbf{TReS*}~\cite{golestaneh2021no}
& 0.3935 & 0.4301 & 0.2695 
& 0.1406 & 0.1520 & 0.0946 

\\

$\heartsuit$ \textbf{MUSIQ*}~\cite{ke2021musiq}
& 0.7985 & 0.8379 & 0.6032 
& 0.5310 & 0.5510 & 0.3789

\\
$\heartsuit$ \textbf{StairIQA*}~\cite{sun2023blind}
&0.8268 & \textbf{\blue{0.8645}} & 0.6346 
&0.5890 & 0.6089 & 0.4199 

\\
$\heartsuit$ \textbf{Q-Align*}~\cite{wu2023qalign}
& \textbf{\blue{0.8311}} & 0.8505 & \textbf{\blue{0.6383}}
 & 0.4547 & 0.4640 & 0. 3096

\\
$\heartsuit$ \textbf{LIQE*} \cite{zhang2023liqe}
& 0.8106 & 0.8268 & 0.6163 
 & 0.5617 & 0.5777 & 0.4013 \\

\hline
\rowcolor{gray!20}\textbf{LMM4LMM (Ours)} 

&\bf\textcolor{red}{0.8863} &\bf\textcolor{red}{0.9094}
&\bf\textcolor{red}{0.7137} 

&\bf\textcolor{red}{0.8969} &\bf\textcolor{red}{0.9162}  
&\bf\textcolor{red}{0.7332}




\\

\rowcolor{gray!20}\textit{Improvement} &+5.5\% & +4.49\% &+7.54\% &+9.61\% &+9.43\% &+6.75\%



\\
\Xhline{1px}
\end{tabular}}\label{mos}
 \vspace{-5mm}
\end{table}

%% file: tabels/SRCC_ACC.tex
\begin{table*}[tbph]
\vspace{-5mm}
\setlength{\belowcaptionskip}{-0.01cm}
\centering
\belowrulesep=0pt
\aboverulesep=0pt
\renewcommand\arraystretch{0.95}
\caption{Performance comparisons of LMMs on the EvalMi-50K across different task-specific challenges. We report the correlation between automatic evaluation metrics and human groundtruth annotations in terms of perception quality SRCC ($\rho_p$), correspondence SRCC ($\rho_c$), and QA accuracy (Acc\%). 
The best results are marked in {\textcolor{red}{RED}} and the second-best in {\blue{BLUE}}. 
*Refers to finetuned models.
}
\vspace{-3mm}
   \resizebox{\linewidth}{!}{\begin{tabular}{l||cc:c|cc:c|cc:c|cc:c|cc:c|cc:c|cc:c}
    \toprule[1pt]
     Dimension  &\multicolumn{3}{c}{\textbf{Single Class}}&\multicolumn{3}{c}{\textbf{Two Class}}&\multicolumn{3}{c}{\textbf{Counting}}&\multicolumn{3}{c}{\textbf{Colors}}&\multicolumn{3}{c}{\textbf{Position}}&\multicolumn{3}{c}{\textbf{Shapes}}&\multicolumn{3}{c}{\textbf{Texture}}\\
  \cmidrule(lr){2-4} \cmidrule(lr){5-7} \cmidrule(lr){8-10} \cmidrule(lr){11-13} \cmidrule(lr){14-16} \cmidrule(lr){17-19}\cmidrule(lr){20-22}
 Methods / Metrics
&$\rho_p$$\uparrow$&$\rho_c$$\uparrow$&Acc$\uparrow$&$\rho_p$$\uparrow$&$\rho_c$$\uparrow$&Acc$\uparrow$&$\rho_p$$\uparrow$&$\rho_c$$\uparrow$&Acc$\uparrow$&$\rho_p$$\uparrow$&$\rho_c$$\uparrow$&Acc$\uparrow$&$\rho_p$$\uparrow$&$\rho_c$$\uparrow$&Acc$\uparrow$&$\rho_p$$\uparrow$&$\rho_c$$\uparrow$&Acc$\uparrow$&$\rho_p$$\uparrow$&$\rho_c$$\uparrow$&Acc$\uparrow$\\
    \midrule
     LLaVA-1.5 (7B)~\cite{liu2024improved} & 0.295  & 0.232 & 84.2 & 0.371  & 0.544 & 78.1 & 0.130 & 0.387 & 61.4 & 0.302  & 0.322 & 84.9 & 0.372 & 0.474 & 45.7 & 0.357 & 0.183 & 53.2 & 0.190 & 0.215 & 63.8  \\
    LLaVA-NeXT (8B)~\cite{liu2024llavanext} & 0.393 & 0.244  & 83.5 & 0.459 & 0.451 & 79.1 & 0.319 & 0.460 & 68.4 & 0.308 & 0.424 & 83.5 & 0.434 & 0.408  & 46.3 & 0.403 & 0.433 & 59.7 & 0.369 & 0.494 & 62.7  \\
    mPLUG-Owl3 (7B)~\cite{ye2024mplug} & 0.468 & 0.238  & 85.1 & 0.430 & 0.526 & 81.1 & 0.433 & 0.581 & 82.3 & 0.449 & 0.360 & 84.3 & 0.426 & 0.498  & 52.4 & 0.414 & 0.320 & 58.0 & 0.400 & 0.289 & 67.6 \\ 
   MiniCPM-V2.6 (8B)~\cite{yao2024minicpm}& 0.423 & 0.350  & 85.1 & 0.311 & 0.697 & 77.9 & 0.361 & 0.711 & 83.9 & 0.353 & 0.547 & 85.5 & 0.445 & 0.607  & 55.6 & 0.452 & 0.533 & 69.2 & 0.301 & 0.545 & 74.6\\
    Qwen2-VL (7B)~\cite{Qwen2-VL} & 0.425 & 0.163  & 84.4 & 0.200 & 0.673 & 79.1 & 0.314 & 0.697 & 78.2 & 0.357 & 0.441 & 84.1 & 0.224 & 0.552  & 59.7 & 0.265 & 0.500 & 59.0 & 0.234 & 0.405 & 66.1\\
     Qwen2.5-VL(7B)~\cite{Qwen2.5-VL}& 0.507  & 0.394 & 18.6 & 0.495  & 0.716 & 44.6 & 0.493 & 0.705 & 53.0 & 0.524  & 0.505 & 21.9 & 0.569 & 0.648 & 74.9 & 0.602 & 0.484 & 51.7 & 0.426 & 0.592 & 46.0\\
    Llama3.2-Vision (11B)~\cite{meta2024llama}
    &0.265&0.275&85.1&0.197&0.185&73.7&0.402&0.284&73.6&0.213&0.316 &87.9&0.233&0.154&48.9&0.252&0.301&66.6&0.257&0.359&72.3\\
    DeepseekVL (7B)~\cite{lu2024deepseekvl}&0.137&0.043&82.5&0.192&0.447&74.1&0.222&0.194&78.9&0.094&0.134&80.3&0.275&0.332&59.7&0.213&0.030&61.2&0.111&0.100&71.7 \\
    DeepseekVL2 (1B)~\cite{wu2024deepseekvl2mixtureofexpertsvisionlanguagemodels}&0.140 &0.032 &18.6 &0.157 &0.051 &44.6 &0.035 &0.028&53.0 &0.070 &0.046 &21.9 &0.048 &0.049 &74.9 &0.136 &0.035 &51.7 &0.078 &0.038 &46.0\\
  CogAgent (18B)~\cite{hong2024cogagentvisuallanguagemodel}& 0.341 & 0.316  & 85.3 & 0.292 & 0.536 & 81.3 & 0.319 & 0.389 & 71.1 & 0.433 & 0.378 & 84.3 & 0.407 & 0.410  & 35.2 & 0.492 & 0.317 & 57.5 & 0.309 & 0.303 & 62.3 \\
    InternVL2.5 (8B)~\cite{chen2024expanding}& 0.233  & 0.225 & 83.5 & 0.253  & 0.625 & 79.1 & 0.205 & 0.574 & 71.6 & 0.185  & 0.355 & 83.9 & 0.306 & 0.565 & 56.2 & 0.197 & 0.436 & 58.5 & 0.162 & 0.497 & 69.6 \\
   InternLM-XComposer (7B)~\cite{internlmxcomposer}& 0.467 & 0.137  & 85.1 & 0.430 & 0.134 & 81.1 & 0.433 & 0.337 & 82.3 & 0.449 & 0.152 & 84.3 & 0.426 & 0.205  & 52.4 & 0.414 & 0.039 & 58.0 & 0.400 & 0.031 & 67.6\\
   \hline
   *DeepseekVL2 (1B)~\cite{wu2024deepseekvl2mixtureofexpertsvisionlanguagemodels}&\bf\blue{0.772} & \bf\blue{0.658}& 89.1 &\bf\blue{0.784} &0.825 &87.8 &\bf\blue{0.766} &\bf\blue{0.820}& 86.9 &\bf\blue{0.774} & \bf\blue{75.1} &\bf\blue{89.5} &\bf\blue{0.795} & 0.604 & 84.8&\bf\blue{0.799}  &0.604& 76.9&\bf\blue{0.738} &0.701& 81.5\\
   *Qwen2.5-VL (7B)~\cite{Qwen2.5-VL}& 0.708  & 0.609 & \bf\blue{89.5} & 0.705  & \bf\blue{0.828} & \bf\blue{89.2} & 0.699& 0.763 & \bf\blue{87.4} & 0.681  & 0.661 & 89.2 & 0.690& \bf\blue{0.777} & \bf\blue{88.3} & 0.725 & \bf\blue{0.788} & \bf\blue{81.6} & 0.585 & \bf\blue{0.788} & \bf\blue{86.2}\\
   *Llama3.2-Vision (11B)~\cite{meta2024llama}&0.706 & 0.558& 84.2& 0.734& 0.613& 72.6&0.729 &0.510&66.1&0.723 & 0.460& 80.9&0.747 & 0.357& 77.9&0.732 & 0.518&70.2&0.711 & 0.570&70.1\\
   
   \rowcolor{gray!20} \textbf{LMM4LMM (Ours)} &\bf\red{0.850}&\bf{\red{0.799}}&\bf\red{89.5}&\bf\red{0.861}&\bf\red{0.899}&\bf\red{89.3}&\bf\red{0.868}&\bf\red{0.867}&\bf\red{87.5}&\bf\red{0.860}&\bf\red{0.826}&\bf\red{89.5}&\bf\red{0.851}&\bf\red{0.841}&\bf\red{88.8}&\bf\red{0.863}&\bf\red{0.817}&\bf\red{81.8}&\bf\red{0.805}&\bf\red{0.852}&\bf\red{87.1}\\
    \bottomrule[1pt]
  \end{tabular}}

  \vspace{0.6mm}
    \resizebox{\linewidth}{!}{\begin{tabular}{l||cc:c|cc:c|cc:c|cc:c|cc:c|cc:c|cc:c}
    \toprule[1pt]
     Dimension  &\multicolumn{3}{c}{\textbf{Scene}}&\multicolumn{3}{c}{\textbf{Style}}&\multicolumn{3}{c}{\textbf{OCR}}&\multicolumn{3}{c}{\textbf{HOI}}&\multicolumn{3}{c}{\textbf{Human}}&\multicolumn{3}{c}{\textbf{Emotion}}&\multicolumn{3}{c}{\textbf{Linguistic Structure}}\\
  \cmidrule(lr){2-4} \cmidrule(lr){5-7} \cmidrule(lr){8-10} \cmidrule(lr){11-13} \cmidrule(lr){14-16} \cmidrule(lr){17-19}\cmidrule(lr){20-22}
 Methods / Metrics
&$\rho_p$$\uparrow$&$\rho_c$$\uparrow$&Acc$\uparrow$&$\rho_p$$\uparrow$&$\rho_c$$\uparrow$&Acc$\uparrow$&$\rho_p$$\uparrow$&$\rho_c$$\uparrow$&Acc$\uparrow$&$\rho_p$$\uparrow$&$\rho_c$$\uparrow$&Acc$\uparrow$&$\rho_p$$\uparrow$&$\rho_c$$\uparrow$&Acc$\uparrow$&$\rho_p$$\uparrow$&$\rho_c$$\uparrow$&Acc$\uparrow$&$\rho_p$$\uparrow$&$\rho_c$$\uparrow$&Acc$\uparrow$\\
    \midrule
     LLaVA-1.5 (7B)~\cite{liu2024improved}& 0.337  & 0.298 & 92.2 & 0.445  & 0.209 & 74.0 & 0.454 & 0.666 & 83.4 & 0.457  & 0.428 & 71.8 & 0.447 & 0.428 & 83.9 & 0.288 & 0.423 & 63.4 & 0.427 & 0.278 & 77.2 \\
    LLaVA-NeXT (8B)~\cite{liu2024llavanext}& 0.517 & 0.336  & 87.3 & 0.502 & 0.326 & 77.6 & 0.560 & 0.629 & 92.6 & 0.521 & 0.596 & 75.8 & 0.486 & 0.310  & 85.3 & 0.444 & 0.502 & 55.9 & 0.565 & 0.565 & 80.7 \\
    mPLUG-Owl3 (7B)~\cite{ye2024mplug}& 0.448 & 0.013 & 87.3 & 0.209 & 0.202 & 76.7 & 0.361 & 0.669 & 86.0 & 0.460 & 0.498 & 75.3 & 0.460 & 0.443  & 83.2 & 0.333 & 0.500 & 62.9 & 0.499 & 0.555 & 84.6 \\ 
       MiniCPM-V2.6 (8B)~\cite{yao2024minicpm}& 0.369 & 0.126  & 87.3 & 0.475 & 0.467 & 77.6 & 0.525 & 0.727 & 85.6 & 0.403 & 0.545 & 74.6 & 0.450 & 0.478  & 85.7 & 0.313 & 0.515 & 64.1 & 0.436 & 0.657 & 82.5\\
    Qwen2-VL (7B)~\cite{Qwen2-VL}& 0.398 & 0.297  & 87.7 & 0.525 & 0.439 & 75.6 & 0.511 & 0.720 & 92.1 & 0.293 & 0.567 & 75.1 & 0.250 & 0.574  & 82.6 & 0.417 & 0.574 & 62.3 & 0.485 & 0.693 & 82.5\\
     Qwen2.5-VL (7B)~\cite{Qwen2.5-VL}& 0.548  & 0.435 & 91.9 & 0.537  & 0.424 & 74.4 & 0.658 & 0.773 & 90.4 & 0.519  & 0.543 & 77.1 & 0.560 & 0.454 & 80.1 & 0.441 & 0.493 & 51.0 & 0.610 & 0.701 & 84.4\\
    Llama3.2-Vision (11B)~\cite{meta2024llama}
    &0.355&0.047&92.2&0.409&0.342&75.1&0.178&0.258&88.8&0.277&0.086 &72.3&0.196&0.162&69.2& 0.237& 0.128& 59.0& 0.322 &0.442 &59.0\\
    DeepseekVL (7B)~\cite{lu2024deepseekvl}
    &0.333&0.109&89.0&0.301&0.012&75.8&0.352&0.199&74.2&0.360&0.286&70.3&0.456&0.141&78.5&0.249&0.330&66.0&0.444&0.502&83.9 \\ 
    DeepseekVL2 (1B)~\cite{wu2024deepseekvl2mixtureofexpertsvisionlanguagemodels}&0.140 &0.031 &16.9 &0.157 &0.006 &26.3 &0.070 &0.067&68.1 &0.048 &0.015 &31.4 &0.136 &0.029 &32.5 &0.078 &0.057 &33.7&0.257 &0.013 &76.5\\
  CogAgent (18B)~\cite{hong2024cogagentvisuallanguagemodel}& 0.357 & 0.024  & 87.3 & 0.586 & 0.252 & 77.4 & 0.493 & 0.383 & 65.1 & 0.460 & 0.307 & 71.6 & 0.356 & 0.163  & 80.3 & 0.298 & 0.256 & 61.6 & 0.482 & 0.404 & 83.0 \\
    InternVL2.5 (8B)~\cite{chen2024expanding}& 0.218  & 0.216 & 89.9 & 0.344  & 0.406 & 74.4 & 0.349 & 0.666 & 72.1 & 0.355  & 0.473 & 73.6 & 0.249 & 0.446 & 82.0 & 0.237 & 0.517 & 58.2 & 0.357 & 0.613 & 74.6 \\
   InternLM-XComposer (7B)~\cite{internlmxcomposer}& 0.448 & 0.034 & 87.3 & 0.209 & 0.026 & 76.7 & 0.361 & 0.060 & 86.0 & 0.460 & 0.275 & 75.3 & 0.460 & 0.080  & 83.2 & 0.333 & 0.157 & 62.9 & 0.499 & 0.346 & 84.6\\
   \hline
   *DeepseekVL2 (1B)~\cite{wu2024deepseekvl2mixtureofexpertsvisionlanguagemodels}&\bf\blue{0.763} & \bf\blue{0.591}& 93.2 &\bf\blue{0.730} &\bf\blue{0.677} & \bf\blue{83.8} &0.855 &0.825& 91.2 &\bf\blue{0.800} & \bf\blue{0.667} & 80.3 &\bf\blue{0.846} & \bf\blue{0.719} & 88.4 &\bf\blue{0.790}  &0.670& 79.8 &\bf\blue{0.812} &0.375& 87.2\\
   *Qwen2.5-VL (7B)~\cite{Qwen2.5-VL}& 0.680  & 0.521 & \bf\blue{93.2} & 0.668  & 0.642 & 83.7 & \bf\blue{0.881}& \bf\blue{0.850} & \bf\blue{92.9} & 0.751  & 0.611 & \bf\blue{83.5} & 0.693& 0.562 & \bf\blue{92.0} & 0.633 & \bf\blue{0.690} & \bf\blue{82.9} & 0.754 & \bf\blue{0.795} & \bf\blue{88.3}\\
    *Llama3.2-Vision (11B)~\cite{meta2024llama}&0.760 & 0.456& 92.5& 0.720& 0.481& 79.0&0.842 &0.526&80.4&0.790 & 0.469& 74.6&0.817 & 0.635& 84.5&0.759 & 0.463&77.5&0.801& 0.273&73.7\\
   \rowcolor{gray!20} \textbf{LMM4LMM (Ours)} &\bf\red{0.856}&\bf{\red{0.755}}&\bf\red{93.3}&\bf\red{0.860}&\bf\red{0.804}&\bf\red{86.1}&\bf\red{0.938}&\bf\red{0.882}&\bf\red{93.0}&\bf\red{0.921}&\bf\red{0.864}&\bf\red{83.5}&\bf\red{0.916}&\bf\red{0.851}&\bf\red{92.0}&\bf\red{0.907}&\bf\red{0.864}&\bf\red{84.7}&\bf\red{0.878}&\bf\red{0.837}&\bf\red{88.4}\\
    \bottomrule[1pt]
  \end{tabular}}
  
  \vspace{0.6mm}
      \resizebox{\linewidth}{!}{\begin{tabular}{l||cc:c|cc:c|cc:c|cc:c|cc:c|cc:c|cc:c}
    \toprule[1pt]
     Dimension  &\multicolumn{3}{c}{\textbf{View}}&\multicolumn{3}{c}{\textbf{World Knowledge}}&\multicolumn{3}{c}{\textbf{Face}}&\multicolumn{3}{c}{\textbf{Imagination}}&\multicolumn{3}{c}{\textbf{Time \& Light}}&\multicolumn{3}{c}{\textbf{Complex}}&\multicolumn{3}{c}{\textbf{Overall}}\\
  \cmidrule(lr){2-4} \cmidrule(lr){5-7} \cmidrule(lr){8-10} \cmidrule(lr){11-13} \cmidrule(lr){14-16} \cmidrule(lr){17-19}\cmidrule(lr){20-22}
 Methods / Metrics
&$\rho_p$$\uparrow$&$\rho_c$$\uparrow$&Acc$\uparrow$&$\rho_p$$\uparrow$&$\rho_c$$\uparrow$&Acc$\uparrow$&$\rho_p$$\uparrow$&$\rho_c$$\uparrow$&Acc$\uparrow$&$\rho_p$$\uparrow$&$\rho_c$$\uparrow$&Acc$\uparrow$&$\rho_p$$\uparrow$&$\rho_c$$\uparrow$&Acc$\uparrow$&$\rho_p$$\uparrow$&$\rho_c$$\uparrow$&Acc$\uparrow$&$\rho_p$$\uparrow$&$\rho_c$$\uparrow$&Acc$\uparrow$\\
    \midrule
     LLaVA-1.5 (7B)~\cite{liu2024improved}& 0.389  & 0.255 & 73.6 & 0.309  & 0.450 & 80.1 & 0.401 & 0.375 & 65.6 & 0.395  & 0.504 & 68.4 & 0.396 & 0.413 & 73.1 & 0.336 & 0.471 & 66.3 & 0.337 & 0.389 & 71.0 \\
    LLaVA-NeXT (8B)~\cite{liu2024llavanext}& 0.534 & 0.284  & 70.5 & 0.522 & 0.313 & 84.7 & 0.475 & 0.438 & 64.7 & 0.579 & 0.568 & 69.0 & 0.504 & 0.448  & 70.5 & 0.403 & 0.413 & 59.7 & 0.433 & 0.457 & 70.7\\
    mPLUG-Owl3 (7B)~\cite{ye2024mplug}& 0.448 & 0.399  & 71.3 & 0.458 & 0.369 & 80.3 & 0.077 & 0.340 & 68.0 & 0.354 & 0.493 & 66.8 & 0.385 & 0.539  & 69.2 & 0.402 & 0.517 & 64.7 & 0.392 & 0.474 & 72.7 \\ 
   MiniCPM-V2.6 (8B)~\cite{yao2024minicpm} & 0.407 & 0.404  & 76.1 & 0.382 & 0.468 & 61.7 & 0.323 & 0.510 & 70.7 & 0.475 & 0.625 & 71.5 & 0.451 & 0.464  & 65.7 & 0.333 & 0.538 & 66.1 & 0.373 & 0.592 & 73.4\\
    Qwen2-VL (7B)~\cite{Qwen2-VL} & 0.518 & 0.488  & 77.1 & 0.590 & 0.540 & 79.7 & 0.492 & 0.577 & 69.1 & 0.385 & 0.581 & 67.9 & 0.517 & 0.543  & 67.4 & 0.350 & 0.605 & 62.2 & 0.376 & 0.590 & 72.6\\
     Qwen2.5-VL (7B)~\cite{Qwen2.5-VL}& 0.537  & 0.457 & 76.1 & 0.562  & 0.496 & 70.5 & 0.626 & 0.549 & 72.1 & 0.598  & 0.602 & 72.6 & 0.556 & 0.504 & 66.7 & 0.549 & 0.685 & 74.7 & 0.528 & 0.640 & 76.2\\
    Llama3.2-Vision (11B)~\cite{meta2024llama}
    &0.311&0.146&73.0&0.235&0.140&66.4&0.274&0.087&71.3&0.233&0.227 &70.4&0.370&0.130&61.5&0.284&0.155&60.9&0.293&0.260&70.8\\
    DeepseekVL (7B)~\cite{lu2024deepseekvl}
    &0.357&0.228&70.1&0.355&0.233&76.7&0.293&0.177&65.2&0.311&0.502&70.1&0.400&0.252&67.8&0.205&0.325&56.8&0.261&0.236&70.7 \\ 
    DeepseekVL2 (1B)~\cite{wu2024deepseekvl2mixtureofexpertsvisionlanguagemodels}&0.200 &0.084 &30.5 &0.175&0.013 &16.9&0.168 &0.039 & 41.2 &0.172 &0.021 &39.8 &0.179 &0.001 &33.9 &0.048 &0.015 &62.2 &0.125 &0.032 &42.7\\
  CogAgent (18B)~\cite{hong2024cogagentvisuallanguagemodel}& 0.349 & 0.308  & 70.9 & 0.435 & 0.312 & 79.3 & 0.488 & 0.366 & 68.2 & 0.395 & 0.385 & 62.2 & 0.428 & 0.314  & 67.3 & 0.413 & 0.280 & 48.0 & 0.386 & 0.358 & 67.5 \\
    InternVL2.5 (8B)~\cite{chen2024expanding}& 0.307  & 0.359 & 70.3 & 0.315  & 0.426 & 78.5 & 0.301 & 0.490 & 66.6 & 0.347  & 0.489 & 71.4 & 0.333 & 0.373 & 56.4 & 0.228 & 0.590 & 60.8 & 0.260 & 0.551 & 70.1 \\
   InternLM-XComposer (7B)~\cite{internlmxcomposer}& 0.448 & 0.227  & 71.3 & 0.458 & 0.205 & 80.3 & 0.077 & 0.113 & 68.0 & 0.354 & 0.188 & 66.8 & 0.385 & 0.020 & 69.2 & 0.402 & 0.170 & 64.7 & 0.392 & 0.173 & 72.7\\
   \hline
      *DeepseekVL2 (1B)~\cite{wu2024deepseekvl2mixtureofexpertsvisionlanguagemodels}&\bf\blue{0.771} & 0.670&83.9 &\bf\blue{0.815} &0.678 & 89.8 &0.837 &\bf\blue{0.689}& 77.2&\bf\blue{0.767} & \bf\blue{0.764} & 81.8 &\bf\blue{0.771} & 0.666 & 82.6&\bf\blue{0.769}  &0.778 &83.7&\bf\blue{0.790} &0.782& 84.9\\
      *Qwen2.5-VL (7B)~\cite{Qwen2.5-VL}& 0.712  & \bf\blue{0.782} & \bf\blue{85.1} & 0.734  & \bf\blue{0.713} & \bf\blue{90.4} & 0.760& 0.648 & \bf\blue{82.2} & 0.594  & 0.736 & \bf\blue{88.0} & 0.707& \bf\blue{0.718} & \bf\blue{82.7} & 0.668 & \bf\blue{0.811} & \bf\blue{89.2} & 0.699 & \bf\blue{0.801} & \bf\blue{87.2}\\
    *Llama3.2-Vision (11B)~\cite{meta2024llama}&0.718 & 0.258& 81.4& 0.759& 0.658& 88.0&\bf\blue{0.837} &0.598&77.3&0.723 & 0.624& 79.6&0.745 & 0.653& 82.2&0.714 & 0.532&75.9&0.756& 0.640&78.1 \\
   \rowcolor{gray!20} \textbf{LMM4LMM (Ours)} &\bf\red{0.870}&\bf{\red{0.814}}&\bf\red{85.0}&\bf\red{0.885}&\bf\red{0.814}&\bf\red{90.5}&\bf\red{0.949}&\bf\red{0.906}&\bf\red{82.4}&\bf\red{0.886}&\bf\red{0.882}&\bf\red{88.5}&\bf\red{0.878}&\bf\red{0.829}&\bf\red{83.0}&\bf\red{0.877}&\bf\red{0.901}&\bf\red{89.2}&\bf\red{0.886}&\bf\red{0.895}&\bf\red{87.9}\\
    \bottomrule[1pt]
  \end{tabular}}
  \label{lmm}
\end{table*}

%% file: tabels/model.tex
\begin{table*}[tbph]
\vspace{-2mm}
\centering
\renewcommand\arraystretch{0.7}
\caption{Comparisons of the alignment between different metric results and human annotations in evaluating T2I model performance.
}
\vspace{-3mm}
   \resizebox{\linewidth}{!}{\begin{tabular}{l||c:>{\columncolor{mycolor_gray}}cccc|c:>{\columncolor{mycolor_gray}}cccc|c:>{\columncolor{mycolor_gray}}cccc|c:>{\columncolor{mycolor_gray}}c}
  \Xhline{1px}
    Dimension  &\multicolumn{5}{c|}{\textbf{Perception Score}}&\multicolumn{5}{c|}{\textbf{Correspondence Score}}&\multicolumn{5}{c|}{\textbf{Question Answering Accuracy (\%)}}&\multicolumn{2}{c}{\textbf{Overall Rank}}\\
  \cmidrule(lr){2-6}  \cmidrule(lr){7-11} \cmidrule(lr){12-16} \cmidrule(lr){17-18}
 Models
&$\text{Human}$&$\text{Ours}$&$\text{Q-Align}$&$\text{StairIQA}$&$\text{LIQE}$ &$\text{Human}$&$\text{Ours}$&$\text{Q-Align}$&$\text{FGA}$&$\text{VQAScore}$&$\text{Human}$&$\text{Ours}$&$\text{Qwen2.5}$&$\text{Llama3.2}$&$\text{Deepseek2}$&$\text{Human}$&$\text{Ours}$\\
    \midrule
    Flux\_schnell~\cite{flux2024} & 60.63   & 61.51   & 92.69 &61.45&4.38& 58.10 & 58.48 &80.08 &3.50 &81.79& 80.29 & 78.64 & 77.23 &76.53 &71.36& 1 & 1 \\
    SD3\_5\_large~\cite{esser2024scaling} & 59.50 & 59.77&  88.78 & 59.00&4.34  & 58.35 & 59.04 & 74.80 & 3.58 &85.28  & 81.43  &  82.18 & 82.98  & 77.39 &77.93 & 2 & 3  \\
    Playground~\cite{li2024playground} & 61.64 & 62.89 & 96.40 &62.12&4.59 & 56.06 & 57.46 & 85.27  & 3.56 & 82.50 & 73.86 &  74.13 & 73.88 & 78.61 & 70.65& 3 & 2  \\ 
   Infinity~\cite{Infinity}& 60.86 &61.50 & 95.73 &61.38& 4.56& 57.43 &  58.17 & 84.37 & 3.45 & 81.93 & 78.10 & 77.32 & 77.05 &78.45 & 73.30& 4 & 4 \\
    DALLE3~\cite{betker2023improving} & 59.35  & 60.27 & 94.72 & 60.02& 4.40&  57.97 & 58.38 &85.87  & 3.52 & 81.82 & 80.24  & 79.15  & 80.05 & 82.29& 77.80& 5 & 5 \\
    Kolors~\cite{kolors}& 61.14 & 62.29 & 95.70 &62.44 & 4.78 & 53.53  & 55.07 & 85.30 & 3.24  & 77.09& 65.05 & 69.00  & 83.18 &78.73 &66.29 &  6 & 6 \\
    Omnigen~\cite{xiao2024omnigen}
    &59.12&60.47&91.04&59.92&4.44&55.81&57.00& 79.78&3.38 &80.10&73.29&72.75&73.97&80.05&68.86&7&7\\
    PixArt-sigma~\cite{chen2024pixart}
    &57.43&59.19&91.11&59.97&4.04&54.72&56.07&80.90&3.39 &79.98&70.71 &70.49 &71.90 &74.24&66.28&8&8 \\
    Show-o~\cite{xie2024show}&52.31 &52.74  &83.61&54.48&3.32   &54.21&54.54 &72.91&3.38&80.58&71.71 &71.74 &71.96 &79.69&67.55&9&9\\
    SDXL\_base\_1~\cite{podell2023sdxl}&53.50 &54.45  &87.28 &54.59&3.60&52.23 &53.51 &75.71&3.29 &81.45&63.67 &65.82 &65.82 &72.15&62.03&10&10\\
  EMU3~\cite{wang2024emu3}& 54.29   & 54.86  & 87.58 &57.78& 3.54&  50.97 &52.61 &78.50&3.12 &76.53&59.90 &61.56 &57.67 & 67.05&58.35&11&12\\
  NOVA~\cite{deng2024nova}& 50.69 & 51.35 & 79.61  & 53.16&3.27 & 52.73 &52.77 &71.39 &3.29 & 78.17 &68.19 &66.89 &62.93 
  & 73.65&61.26&12&14\\
   Kandinsky-3~\cite{arkhipkin2024kandinsky}& 58.21  & 58.74 & 93.58 &60.58&4.21 & 48.37 &51.60 &79.72&2.84 &72.79&50.14 &57.60 &61.27 &69.36&55.15&13&13\\
   Seed-xi~\cite{ge2024seed}& 50.73 & 51.49 & 79.74  &53.59&3.07 & 50.96 &53.93 &70.08&3.18 &81.95&61.43 &66.28 &66.28 &72.94&60.32&14&11\\
   LaVi-Bridge~\cite{zhao2024bridging}& 50.56 & 51.16  & 66.27 & 50.66& 3.09& 50.19 &51.01 &60.83&3.08 &69.22&59.10 &62.40 &57.70 &73.63&55.09&15&15\\
   Hart~\cite{tang2024hart}&49.80&49.85&88.87&53.75&3.20& 50.30 &53.04 &81.24&3.14 &76.10&59.29 &61.99 &67.07 &72.40&60.53&16&16\\
   ELLA~\cite{hu2024ella}&44.61&45.17&57.68&44.30&2.24&49.07 &50.14 &54.29&3.10 &75.19&54.90 &56.71 &58.35 &71.29&49.65&17&18\\
   SD\_v2-1~\cite{Rombach_2022_CVPR}&47.68&49.23&75.27&50.71&2.69&47.96 &50.41 &64.80&3.02 &77.42&48.86 &54.39 &60.33 &65.80&52.49&18&17\\
   LLMGA~\cite{xia2024llmga}&48.67&50.54&81.63&51.16&2.90&43.43 &46.21 &73.96&2.59 &59.66&37.67 &44.91 &40.04 &65.27&43.58&19&19\\
   Janus~\cite{wu2024janus}&36.98&37.34&41.82&37.81&1.55&45.94 &47.16 &41.31&2.82 &78.62&42.95 &48.67 &49.39 &48.18&36.56&20&20\\
   Vila-u~\cite{wu2024vila}&33.80&33.18&19.54&33.80&1.23&43.47 &44.32 &33.75&2.61 &71.08&35.24 &35.85 &35.85 &37.32&28.05&21&21\\
   i-Code-V3~\cite{tang2023any}&34.70&35.14&20.76&32.62&2.41 &39.80 &40.49 &31.80&1.68&60.11&25.00 &30.98 &26.70&26.95 &21.66&22&22\\
   LlamaGen~\cite{sun2024autoregressive}&29.96&30.74&12.18&29.80&1.25&37.73 &39.09 &27.46&2.31 &61.35&21.19 &22.88 &22.88 &20.82&19.54&23&23\\
   LWM~\cite{liu2024world}&28.88&29.26&11.54&29.14&1.45&35.46 &36.52 &24.42&2.09 &58.52&15.48 &15.88 &18.12 &17.45&13.20&24&24\\
   \hline
   SRCC   to  human $\uparrow$& - & \textbf{\red{0.979}}& 0.940&0.959&\textbf{\blue{0.978}}&-&\textbf{\red{0.983}}&0.777 &\textbf{\blue{0.982}}&0.695& - &\textbf{\red{0.993}}&0.924&0.915 &\textbf{\blue{0.985}} &-&\textbf{\red{0.992}}\\
   RMSE to human $\downarrow$& -  & \textbf{\red{1.24}} & 28.90&\textbf{\blue{2.01}}&47.80&-&\textbf{\red{1.60}} &\textbf{\blue{21.85}}& 47.50&26.52&-&\textbf{\red{3.34}}&5.75&10.73 &\textbf{\blue{4.44}}&-&\textbf{\red{0.866}}\\
    \Xhline{1px}
  \end{tabular}}\label{model}

 \vspace{-3mm}
\end{table*}

%% file: figures/leida22.tex
\begin{figure*}[h]
    \centering
         \vspace{-3mm}
    \includegraphics[width=1\textwidth]{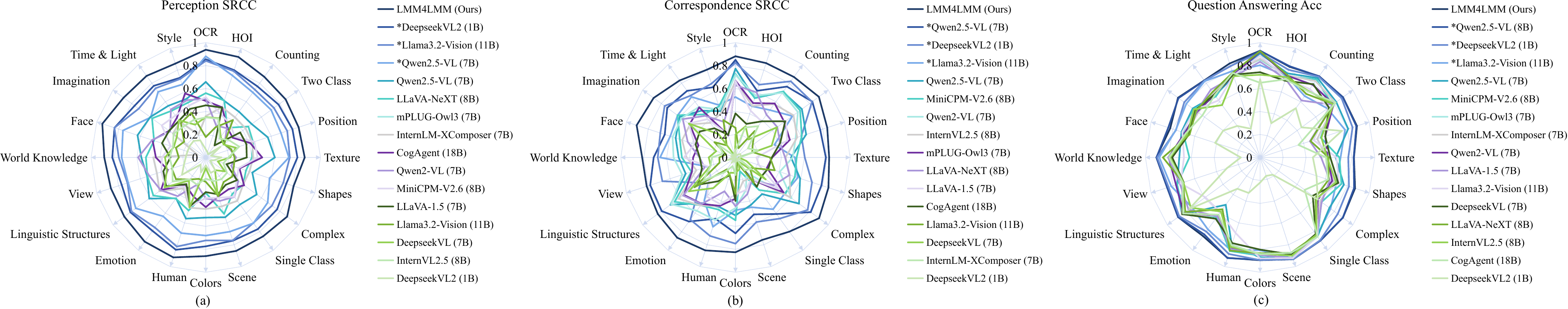}
     \vspace{-7mm}
    \caption{Comparison of MOSs and QA accuracy of different LMM models across different prompt challenges. (a) Results across perceptual quality MOS. (b) Results across T2I correspondence MOS. (c) Results across question answering accuracy.} 
    \label{leidat22}
\end{figure*}

%% file: figures/example.tex
\begin{figure*}[h]
    \centering
    \includegraphics[width=1\textwidth]{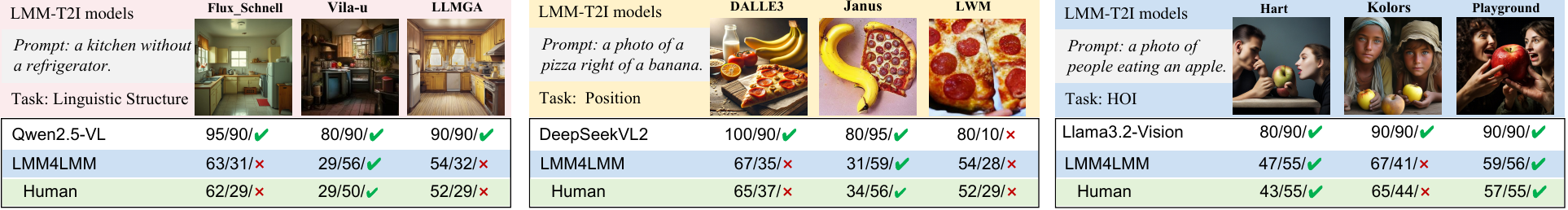}
     \vspace{-7mm}
    \caption{Visualization of the Perception/Correspondence/QA prediction from different methods compared to human annotation. } 
     \vspace{-1mm}
    \label{example}
\end{figure*}

%% file: tabels/cross.tex
\begin{table*}[!h]
  \centering
    \caption{Zero-shot cross-dataset performance comparison on multiple benchmarks. We finetune our model on EvalMi-50K/EvalMuse-40K respectively. FGA-BLIP2~\cite{han2024evalmuse40kreliablefinegrainedbenchmark} is finetuned on EvalMuse-40K~\cite{han2024evalmuse40kreliablefinegrainedbenchmark}. *Refers to scores finetuned on the specific dataset.} 
  \label{tab:method}%
    \vspace{-3mm}
  \renewcommand\arraystretch{0.98}
  \resizebox{\textwidth}{!}{
    \begin{tabular}{l|cc|cc|cc|cc|cc|cc|cc|cc}
    \Xhline{1px}
    \multirow{2}[1]{*}{Method}& \multicolumn{2}{c|}{\textbf{EvalMi-50K (Ours)}} & \multicolumn{2}{c|}{EvalMuse~\cite{han2024evalmuse40kreliablefinegrainedbenchmark}} & \multicolumn{2}{c|}{GenAI-Bench~\cite{li2024genai}} & \multicolumn{2}{c|}{TIFA~\cite{hu2023tifa}} & \multicolumn{2}{c|}{RichHF~\cite{liang2024rich}} & \multicolumn{2}{c|}{AGIQA3K~\cite{liang2024rich}} & \multicolumn{2}{c|}{AIGCIQA~\cite{liang2024rich}} & \multicolumn{2}{c}{CompBench~\cite{huang2023t2i}}\\
\cline{2-17}     & SRCC  & PLCC     & SRCC  & PLCC  & SRCC  & PLCC  & SRCC  & PLCC  & SRCC  & PLCC & SRCC  & PLCC & SRCC  & PLCC & SRCC  & PLCC\\
    \hline
    CLIPScore~\cite{hessel2021clipscore} & 0.2607 & 0.3072 & 0.2993 & 0.2933 & 0.1676 & 0.2030 & 0.3003 & 0.3086 & 0.0570 & 0.3024 & 0.5972 & 0.6839 & 0.2337 & 0.6839&0.2044&0.1944\\
    BLIPScore~\cite{li2022blip} & 0.2900 & 0.3468 & 0.3583 & 0.3348 & 0.2734 & 0.2979 & 0.4287 & 0.4543 & 0.1425 & 0.3105 & 0.6230 & 0.7380 & 0.3784 & 0.2576&0.3967&0.3940\\
    ImageReward~\cite{xu2023imagereward} & 0.4991 & 0.5523 & 0.4655 & 0.4585 & 0.3400  & 0.3786 & 0.6211 & 0.6336 & 0.2747 & 0.3291 & 0.7298 & 0.7862 & 0.5870 & 0.5911&0.4367 &0.4307\\
    PickScore~\cite{kirstain2023pick} & 0.4611 & 0.4692 & 0.4399 & 0.4328 & 0.3541 & 0.3631 & 0.4279 & 0.4342 & 0.3916 & 0.4133 & 0.6977 & 0.7633 & 0.5045 & 0.4998&0.1115&0.0955\\
    HPSv2~\cite{wu2023human} & 0.5336 & 0.5525 & 0.3745 & 0.3657 & 0.1371 & 0.1693 & 0.3647 & 0.3804 & 0.1871 & 0.2577 & 0.6349 & 0.7000 & 0.6068 & 0.5989 & 0.2844 & 0.2761\\
    VQAScore~\cite{li2024evaluating} & 0.6062 & 0.6118 & 0.4877 & 0.4841 & 0.5534 & 0.5175 & 0.6951 & 0.6585 & 0.4826 & 0.4094 & 0.6273 & 0.6677 & 0.6394 & 0.5869 &0.5832 & 0.5328\\
    
    FGA-BLIP2~\cite{han2024evalmuse40kreliablefinegrainedbenchmark} & 0.6755 & 0.6916  & \textbf{\blue{0.7723*}} & \textbf{\blue{0.7716*}} & 0.5638 & 0.5684 & \textbf{\blue{0.7657}} & \textbf{\blue{0.7508}} & 0.4576 & 0.4967 & 0.7793 & 0.8042 & \textbf{\blue{0.7432}} & \textbf{\blue{0.7367}}& \textbf{\blue{0.6231}} &\textbf{\blue{0.6007}} \\
    \hline
    \rowcolor{gray!20}Ours (Train on EvalMi)  &  \textbf{\red{0.8702*}} & \textbf{\red{0.8924*}}  & 0.6560  & 0.6503  & \textbf{\red{0.7082}} & \textbf{\red{0.7019}} & \textbf{\red{0.7734}} & \textbf{\red{0.7604}} & \textbf{\red{0.6231}} & \textbf{\red{0.6259}} & \textbf{\red{0.8011}} & \textbf{\red{0.8205}} & \textbf{\red{0.7514}} & \textbf{\red{0.7473}} & \textbf{\red{0.6911}} & \textbf{\red{0.6726}} \\
     \rowcolor{gray!20}Ours (Train on EvalMuse) &  \textbf{\blue{0.6764}} & \textbf{\blue{0.6928}}  & \textbf{\red{0.7852*}}  & \textbf{\red{0.7958*}}  & \textbf{\blue{0.6523}} & \textbf{\blue{0.6363}} &  0.7390 & 0.7264 & \textbf{\blue{0.5836}} & \textbf{\blue{0.5972}} & \textbf{\blue{0.7797}} & \textbf{\blue{0.8118}} & 0.6823 & 0.6782 &0.5090 &0.5020  \\


    \Xhline{1px}
    \end{tabular}%
  }
\end{table*}%

%% file: tabels/ablation.tex
\begin{table*}[!h]
\vspace{-1mm}
\renewcommand\arraystretch{0.95}
  \caption{Ablation study on the quality-level initialization, LoRA fine-tuning strategy, and the different backbone of LMM4LMM.}

  \vspace{-3mm}
  \resizebox{1\textwidth}{!}{
  \begin{tabular}{ccccc|ccc:ccc:c|ccc|ccc}
    \toprule
                                   & \multicolumn{4}{c}{Backbone \& Strategy}                                  & \multicolumn{3}{c}{Quality (ours)}                   & \multicolumn{3}{c}{Correspondence (ours)}                  & \multicolumn{1}{c}{QA} &
                                   \multicolumn{3}{c}{GenAI-Bench} &
                                   \multicolumn{3}{c}{AGIQA3K}\\
    \cmidrule(r){2-5} \cmidrule(r){6-8} \cmidrule(r){9-11}\cmidrule(r){12-12} \cmidrule(r){13-15}\cmidrule(r){16-18}
    \multicolumn{1}{c}{No.}        & Backbone   & quality level & LoRA$_{r=8}$ (vision)    & LoRA$_{r=8}$ (llm)    & SRCC    & PLCC        &KRCC  & SRCC       & PLCC    & KRCC  &  Acc    &    SRCC     & PLCC   & KRCC          & SRCC      & PLCC   & KRCC      \\
    \hline
    \multicolumn{1}{c}{(1)}          & InternVL2.5 (8B)   &    \ding{52}       &          &             & 0.828    &  0.857    &  0.700   &  0.870   & 0.892     & 0.742     &   86.1\%   & 0.660   & 0.653   &   0.535  &   0.757  &  0.741 & 0.613  \\
    \multicolumn{1}{c}{(2)}          & InternVL2.5 (8B)   &         &      \ding{52}      &    \ding{52}         &  0.865   & 0.895    &  0.687   &    0.888  &  0.906   &  0.721   &  87.9\%    & 0.707    & 0.701         & 0.530   & 0.799 &0.817 &0.605  \\
    \multicolumn{1}{c}{(3)}          &   InternVL2.5 (8B)          &  \ding{52} &           &       \ding{52}     & 0.872    & 0.900     & 0.695     & 0.897    & 0.911     & 0729     &   87.3\%  &  0.689   & 0.680    & 0.515&   0.790   &   0.768  &  0.607 \\
    \multicolumn{1}{c}{(4)}          & InternVL2.5 (8B)   &  \ding{52} & \ding{52} &     &  0.871  & 0.900   &  0.694      & 0.893     & 0.913     & 0.723     & 86.9\%  &  0.688   &  0.680   & 0.514  &  0.778    & 0.810     &  0.593  \\
    \rowcolor{gray!20} \multicolumn{1}{c}{(5)}          &   InternVL2.5 (8B)        &  \ding{52}         & \ding{52} &      \ding{52}         & 0.886     & 0.909    & 0.714     & 0.897    & 0.916    & 0.733    & 87.9\%     & 0.708     & 0.702     & 0.532    & 0.801 & 0.821    & 0.608  \\
    \multicolumn{1}{c}{(6)}          & InternVL2.5 (26B)  & \ding{52} &        &         & 0.834     & 0.867     & 0.704    & 0.848     & 0.866     & 0.718     &  86.6\%    &  0.671    &  0.663    &  0.550    & 0.770    & 0.793 & 0.634   \\
     \multicolumn{1}{c}{(7)}          & InternVL2.5 (26B)   & \ding{52} & \ding{52} & \ding{52}  & 0.882    & 0.906     & 0.709     & 0.897     & 0.906   & 0.727     & 86.9\%&0.726   &  0.741  &  0.548  &0.811    &  0.814 & 0.627\\
      \multicolumn{1}{c}{(8)}          & DeepseekVL2 (1B)  & \ding{52} &  \ding{52}      &   \ding{52}      &   0.790   &  0.825   &  0.651    &  0.782    &   0.799   &  0.646   &  84.9\%  & 0.613  &0.616   & 0.500 & 0.782 &0.712& 0.558  \\
        \multicolumn{1}{c}{(9)}          & Qwen2.5VL (8B)   & \ding{52} & \ding{52} & \ding{52}  &  0.699  &   0.750   &  0.572   &  0.801 &  0.822 & 0.666&87.2\% & 0.626  &  0.616   &   0.505  & 0.767  & 0.786 & 0.619 \\
         \multicolumn{1}{c}{(10)}          & Llama3.2VL (11B)  &  \ding{52}&   \ding{52}    &  \ding{52}    &  0.756  & 0.789   &   0.616 &   0.640   &  0.646   &  0.517  &  78.1\%   &  0.397   & 0.418   & 0.315   & 0.678   & 0.747 & 0.5500  \\

    \bottomrule
  \end{tabular}\label{ablation}
   }
  \vspace{-4mm}
  \centering
\end{table*}

%% file: sec/6_conclusion.tex
\section{Conclusion}
\vspace{-4mm}
In this paper, we introduce EvalMi-50K, a large-scale dataset and benchmark consisting of 50,400 images generated by 24 T2I models using 2,100 prompts across 20 task-specific challenges and 2M+ subjective ratings from the perception, text-image correspondence, and task-specific accuracy, respectively. We use EvalMi-50K to benchmark and evaluate both the generation ability of T2I models and the interpretation ability of LMMs. 
Based on EvalMi-50K, we propose LMM4LMM, an LMM-based evaluation model that leverages instruction tuning and LoRA adaptation to achieve AIGI perceptual quality evaluation and T2I correspondence attribution. Extensive experiments demonstrate that LMM4LMM achieves state-of-the-art performance on the EvalMi-50K dataset and manifests strong zero-shot generalization ability on the other seven benchmarks.

%% file: sec2/1_overview.tex
\section{Overview}
In this supplementary material, we provide additional details on the data collection, methodology, experiments, and results discussed in the main paper. In data collection, we detail the 20 distinct tasks in Section \ref{tasks} and the overview of the 24 LMM-T2I models in Section \ref{modelss}. We then elaborate on the subjective experiments in Section \ref{exp}, including the annotation dimension, criteria, interface and management. In addition, we provide an in-depth analysis of the EvalMi-50K database, including MOS distributions and model performance comparisons across the 20 tasks in Section \ref{analyze}. We outline the loss functions used in the training process for the LMM4LMM model in Section \ref{loss}. Details on the evaluation criteria and algorithms are also included in Section \ref{implementation}. Finally, we provide more performance comparisons between our model and other metrics in Section \ref{comparison}.

%% file: sec2/2_prompt.tex
\section{T2I Task-specific Challenge Define}
\label{tasks}
In this study, we systematically investigate the capabilities of text-to-image (T2I) generation models through a comprehensive evaluation framework. We focus on 20 distinct tasks that vary in complexity and require diverse compositional skills, as detailed in Table \ref{prompt} with their corresponding subcategories, keywords, and example prompts. These tasks are carefully designed to assess different aspects of model performance, ranging from basic object rendering to complex spatial and attribute understanding, as shown in Figures \ref{task1}-\ref{task5}. Below, we provide an overview of the main task categories and their associated challenges.
\input{tables2/prompt}
\begin{itemize}
\item \textbf{Single class}: evaluates a model's ability to generate a single instance of a specified object class. The challenge lies in producing high-fidelity representations that maintain essential class-specific features without additional contextual constraints.

\item \textbf{Two class}: evaluates a model's capacity to simultaneously render two distinct object classes within a single image. This task introduces the challenge of maintaining object integrity while managing inter-object relationships. The complexity increases when considering potential occlusions, relative scaling, and basic spatial arrangements between the two objects.

\item \textbf{Counting}: evaluates a model's ability to generate a specific number of objects in a scene. The challenge includes numerical understanding and managing multiple instances without overlap or spatial issues, especially for larger numbers.

\item \textbf{Colors}: evaluates a model's proficiency in associating specific color attributes with generated objects. The challenge lies in accurately binding color properties to target objects while maintaining object integrity and distinguishing foreground objects from background elements. 

\item \textbf{Position}: evaluates a model's capability to render two objects with specified positional relationships. The challenge encompasses not only object generation but also the accurate representation of specific spatial relationships (\textit{e.g.}, above, below, left of, right of). This requires precise control over object arrangement while maintaining their identities.

\item \textbf{Shapes}: evaluates a model's ability to generate objects with specific geometric shapes (\textit{e.g.}, spherical, rectangular, triangular, star) while preserving their recognizability. This tests the ability to abstract representations of real-world objects and express them in other shapes.

\item \textbf{Texture}: evaluates a model's capability to render objects with specific surface textures and material properties (\textit{e.g.}, metallic, wooden, glass). The challenge lies in creating realistic textures that match the object's properties and lighting conditions.

\item \textbf{Scene}: evaluates a model's ability to create complex scenes with multiple naturally composed elements in a specific environment (\textit{e.g.}, beach, forest, kitchen). The challenge is to ensure all objects and backgrounds are contextually relevant and spatially consistent, evaluating the model's holistic scene understanding.

\item \textbf{Style}: evaluates a model's proficiency in generating images in specific artistic styles (\textit{e.g.}, watercolor, oil painting, cartoon). The challenge is to mimic the style's visual characteristics while keeping objects and scenes recognizable, testing the model's ability to apply abstract stylistic concepts consistently.

\item \textbf{OCR (Optical Character Recognition)}: evaluates a model's capability to generate readable text within images, such as words or short sentences. The challenge is to make the text visually coherent with the image and machine-readable by OCR systems, testing the model's understanding of typography and text integration.

\item \textbf{HOI (Human-Object Interaction)}: evaluates a model's ability to generate realistic interactions between humans and objects, ensuring the actions are physically plausible. The challenge is to create recognizable humans and objects while maintaining natural spatial and logical relationships.
\item \textbf{Human}: evaluates a model's ability to generate human figures with specific occupational attire, unique accessories, and hairstyles. The challenge lies in creating realistic and coherent human representations while maintaining consistency across these attributes. 

\item \textbf{Emotion}: evaluates a model's ability to convey specific emotions or moods, either through human facial expressions (\textit{e.g.}, happiness, sadness) or through the overall atmosphere of a scene (\textit{e.g.}, serene, love). This evaluates the model's understanding of emotional cues and its ability to translate abstract emotions into visuals.

\item \textbf{Linguistic Structure}: evaluates a model's ability to interpret and render linguistic structures involving negation (\textit{e.g.}, “without,” “no”). The challenge is to generate images that accurately reflect the absence of specified objects or features (\textit{e.g.}, a “classroom without people”) while maintaining scene integrity. This tests the model's comprehension of negative constructs.

\item \textbf{View}: evaluates a model's ability to generate images from specific viewpoints (\textit{e.g.}, first-person, third-person, side view). The challenge is to maintain correct spatial orientation, scale, and proportion across perspectives, testing the model's understanding of spatial geometry.

\item \textbf{World Knowledge}: evaluates a model's knowledge of real-world landmarks, historical sites (\textit{e.g.}, the Great Wall, Eiffel Tower, Great Pyramid), and the physical appearances of famous individuals (\textit{e.g.}, Albert Einstein).  The challenge lies in creating content that accurately aligns with people's perceptions of famous landmarks and the physical appearances of well-known individuals. 

\item \textbf{Face}: evaluates a model's ability to generate human faces with specific features (\textit{e.g.}, face shape, nose structure, hairstyle). The challenge is to create realistic and diverse facial representations while maintaining feature consistency, and testing the model's understanding of facial anatomy.

\item \textbf{Imagination}: evaluates a model's ability to generate imaginative scenes that combine elements from different categories or depict impossible scenarios in the real world (\textit{e.g.}, a “cat wearing a chef's hat cooking in a kitchen”). The challenge is to balance creativity with visual plausibility, evaluating the model's capacity for creative thinking and novel concept synthesis.

\item \textbf{Time \& Light}: evaluates a model's ability to generate images that accurately depict different times of day (\textit{e.g.}, morning, evening) and lighting conditions (\textit{e.g.}, sunlight, dim light). The challenge is to adjust brightness, color temperature, shadows, and reflections appropriately and test the model's understanding of time-based lighting dynamics and its ability to visually represent them.

\item \textbf{Complex}: is designed by combining simpler task components, such as color recognition, object counting, and shape identification, into more intricate and multifaceted challenges.  These tasks require models to integrate and execute multiple simple tasks simultaneously within a single image. Below are some combined forms of complex tasks along with corresponding examples:
\begin{itemize}
\item[(1)] \textbf{Counting + Color + Shapes + Scene}: A photo of [number] [color] [class] [action] in a [shape] [scene].
\textbf{Example}: \textit{A photo of two white dogs swimming in a triangle-shaped swimming pool}.

       \item[(2)] \textbf{Counting + Color + Shapes + Texture}: A photo of [number] [color] [texture] [shape] [class].  
\textbf{Example}:\textit{ A photo of two brown wooden rectangular books}.

\item[(3)] \textbf{HOI + Color + Shape + Texture}: A photo of [human action] a [color] [texture] [shape] [object].  
\textbf{Example}: \textit{A photo of people opening a yellow wooden triangle box}.

\item[(4)] \textbf{Style + Color + Position}: A [style] image of a [color1] [class1] [position] a [color2] [class2].  
\textbf{Example}: \textit{A cartoon image of a yellow dog to the left of a white cat}.

\item[(5)] \textbf{Style + OCR + Color}: A [style] image of [color] text “\texttt{[content]}”.  
\textbf{Example}: \textit{An oil painting of red text “\texttt{CONGRATULATIONS}”}.

\item[(6)] \textbf{OCR + Color + Single Class}: A photo of [color1] text “\texttt{[content]}” on a [color2] [class].  
\textbf{Example}: \textit{A photo of green text “\texttt{Happy Birthday}” on a pink cake}.

\item[(7)] \textbf{Counting + Shapes + Two Classes}: A photo of [number1] [shape1] [class1] and [number2] [shape2] [class2].  
\textbf{Example}: \textit{A photo of six spherical balls and three rectangular cups}.

\item[(8)] \textbf{Counting + Color + Two Classes}: A photo of [number1] [color1] [class1] and [number2] [color2] [class2].  
\textbf{Example}: \textit{A photo of six red books and four blue pens}.

\item[(9)] \textbf{View + World Knowledge}: A [view] of [famous landmark].  
\textbf{Example}: \textit{An aerial view of the Great Wall}.

\item[(10)] \textbf{Human + Emotion}: A [human description] [action] with [emotion].  
\textbf{Example}: \textit{A girl in a white blouse and navy skirt, wearing a red ribbon tie, smiles with excitement as she receives a trophy during a school award ceremony. Her long brown hair shines as she turns to the audience}.
    \end{itemize}
\end{itemize}

%% file: tables2/prompt.tex
\begin{table*}
\caption{Prompt categories with corresponding keywords and examples.
}
\label{prompt}
\renewcommand\arraystretch{1.2}
\resizebox{\textwidth}{!}{
    \begin{tabular}{llp{6cm}}
    \toprule
   Category   & Subcategory / Keywords  &  Prompt examples\\
    
    \midrule   
     Single Class &  person, bicycle, car, motorcycle, airplane, bus, train, truck, boat, traffic light, \dots &   A photo of a bench \\
     
     Two Class& bench \& sports, sheep \& dog, cow \& elephant, knife \& spoon, chair \& couch, \dots& A photo of a bench and a sports ball\\
     
     Counting&zero, one, two, three, four, five, six, seven, eight, nine, ten&A photo of three computer keyboards \\
     Colors&red, orange, yellow, green, blue, purple, pink, brown, black, white & A photo of a black donut\\
     
     Position&left of, right of, above, below & A photo of a bottle right of a train\\
     
     Shapes&circle, cylinder, sphere, star, triangle, rectangle, irregular, oval, linear, cone & A photo of a circle skateboard\\
     
     Texture&glass, cement, stone, rubber, fabric, ceramics, leather, metallic, wooden, plastic &A photo of a fabric model bicycle\\
     
     Scene& kitchen, living room, street, swimming pool, playground, waterfall, forest&A photo of in the forest \\
     
     Style&cartoon, realistic, oil painting, vintage, watercolor, line drawing &A vintage image of a tv remote \\
     
     OCR&``HELLO", ``STOP", ``SUCCESSFUL", ``Have a nice day", ``Enjoy life", ``Keep going", \dots &A photo of phrase ``Believe in yourself" \\
     
     HOI&hold a stop sign, operate an oven, peel an apple, lie on a bench, carry a book, \dots &A photo of people boarding a car \\
     
     Human&human, cloth, cloth-color, hair, hair-color & A man in a blue shirt smiles warmly, his curly black hair framing his face\\
     Emotion&happy, sadness, love, fear, surprise, anger, worry, neutrality &A dog is smiling with happy emotion. He finds a lot of delicious food \\
     Linguistic Structure& without, no, not& The garden has no flowers blooming. It is late in the winter\\
     View&close-up, ground view, aerial view, overhead view, first-person view, wide-angle view, \dots& An overhead view of a pickup truck with boxes in its flatbed \\
     World Knowledge&Great Wall, Great Pyramid, Ha Long Bay, Machu Picchu, Eiffel Tower, Grand Canyon, \dots& boats in Ha Long Bay\\
     
     Face&hair, mouth, emotion, eyes, necklace, cheeks, nose, skin &A face image with medium length hair, wearing necklace \\
     
     Imagination& ——&A panda is flying in the sky \\
     
     Time \& Light & \renewcommand\cellalign{t}\makecell[l]{time: sunset, early morning, night, midnight, midday, noon, dawn, \dots \\ light: fiery orange, golden, moonlight, silvery, misty, bright, crimson, \dots}& As the sun sets, fiery orange light streaks across the sky, casting a warm glow over the city skyline and the distant hills\\
     
     Complex& Counting + Color + Shapes + Scene, Style + Color + Position, Human + Emotion, \dots &A photo of four blue birds playing on a circle playground \\
      
        \bottomrule
    \end{tabular}
}
\centering
\end{table*}

%% file: sec2/3_models.tex
\section{Detailed Information of T2I Models }
\label{modelss}
\input{tables2/url}

\noindent 
{\bf Stable Diffusion v2.1}~\cite{Rombach_2022_CVPR} is a model designed for generating and modifying images based on text prompts. It is a Latent Diffusion Model that employs a fixed, pretrained text encoder (OpenCLIP-ViT/H~\cite{radford2021learning}). It is conditioned on the penultimate text embeddings of a CLIP ViT-H/14~\cite{radford2021learning} text encoder.  

\noindent 
{\bf i-Code-V3}~\cite{tang2023any} is a composable diffusion model capable of generating language, image, video, and audio from any input combination. It reuses Stable Diffusion 1.5’s structure and weights, leveraging large-scale datasets like LAION-400M to achieve high-quality multi-modal generation with strong cross-modal coherence.

\noindent 
{\bf Stable Diffusion XL (SDXL)}~\cite{podell2023sdxl} massively increases the UNet backbone size from Stable Diffusion v2~\cite{Rombach_2022_CVPR} and incorporates two text encoders. There is a second refinement model, which we do not use as it does not affect the composition of the image.

\noindent 
{\bf DALLE3}~\cite{betker2023improving}, developed by OpenAI, enhances spatial reasoning and improves the handling of complex prompts by leveraging advanced transformer architectures and refined training datasets, enabling the generation of highly detailed and contextually accurate images.

\noindent 
{\bf LLMGA}~\cite{xia2024llmga} enhances multimodal large language models (MLLMs) by generating detailed text prompts for Stable Diffusion (SD)~\cite{Rombach_2022_CVPR}, improving contextual understanding and reducing noise in generation. It leverages a diverse dataset for prompt refinement, image editing, and inpainting, enabling more precise and flexible image synthesis.

\noindent 
{\bf Kandinsky-3}~\cite{arkhipkin2024kandinsky} is a hybrid model combining diffusion and transformer architectures, which emphasizes artistic and abstract image generation. It is particularly effective for creating visually striking and imaginative compositions.

\noindent 
{\bf LWM}~\cite{liu2024world} is a multimodal autoregressive model trained on extensive video and language data. Using RingAttention, it efficiently handles long-sequence training, expanding context size up to 1M tokens, enabling strong language, image, and video understanding and generation.

\noindent 
{\bf Playground}~\cite{li2024playground} is designed for high-resolution and photorealistic outputs, which incorporates advanced noise scheduling and fine-tuning techniques. It is optimized for generating detailed and visually appealing images with minimal artifacts.

\noindent 
{\bf LaVi-Bridge}~\cite{zhao2024bridging} is designed for text-to-image diffusion models and serves as a bridge, which enables the integration of diverse pre-trained language models and generative vision models for text-to-image generation. By leveraging LoRA and adapters, it offers a flexible and plug-and-play approach without requiring modifications to the original weights of the language and vision models.

\noindent 
{\bf ELLA}~\cite{hu2024ella} is a method that enhances current text-to-image diffusion models with state-of-the-art large language models (LLMs) without requiring the training of LLMs or U-Net. We design a lightweight and adaptive Timestep-Aware Semantic Connector (TSC) to effectively condition the image generation process, ensuring comprehensive prompt understanding from the LLM. With ELLA, the diffusion model can generate high-fidelity and accurate images based on long, information-dense prompts.

\noindent 
{\bf Seed-xi}~\cite{ge2024seed} is a unified and versatile foundation model that can serve as a multimodal AI assistant in real-world applications. Through different instruction tuning, it can respond to various user needs by unifying multi-granularity comprehension and generation.

\noindent
{\bf PixArt-sigma}~\cite{chen2024pixart} is a Diffusion Transformer model (DiT) capable of directly generating images at 4K resolution. Representing a significant advancement over its predecessor, PixArt-alpha~\cite{chen2023pixartalphafasttrainingdiffusion}, it offers markedly higher image fidelity and improved alignment with text prompts. A key feature of PixArt-sigma~\cite{chen2024pixart} is its training efficiency.

\noindent 
{\bf LlamaGen}~\cite{sun2024autoregressive} applies the next-token prediction paradigm of large language models to image generation. By refining image tokenizers and training datasets, it surpasses diffusion models in class-conditional generation and maintains competitive text alignment in text-to-image synthesis.

\noindent 
{\bf Kolors}~\cite{kolors} is a large-scale latent diffusion model developed by the Kuaishou Kolors team for text-to-image generation. Trained on billions of text-image pairs, it outperforms both open-source and closed-source models in visual quality, complex semantic accuracy, and text rendering. Supporting both Chinese and English inputs, it excels at generating high-fidelity images while demonstrating strong performance in understanding Chinese-specific content.

\noindent 
{\bf Flux\_schnell}~\cite{flux2024} is a 12 billion parameter rectified flow transformer capable of generating images from text descriptions. Trained using latent adversarial diffusion distillation, it can generate high-quality images in only 1 to 4 steps.The model is very responsive and suitable for personal development

\noindent
{\bf OmniGen}~\cite{xiao2024omnigen} is a unified image generation model capable of producing a wide range of images from multi-modal prompts. It is designed to be simple, flexible, and easy to use. As a new diffusion model for unified image generation, it not only excels in text-to-image generation but also inherently supports various downstream tasks, such as image editing, subject-driven generation, and visual conditional generation.

\noindent 
{\bf EMU3}~\cite{wang2024emu3} is a multimodal model that leverages next-token prediction as its sole training paradigm. By tokenizing images, text, and videos into a unified discrete space, it enables a single Transformer to be trained from scratch on diverse multimodal sequences.  It streamlines the multimodal learning process, enhancing both efficiency and versatility in handling complex multimodal interactions.

\noindent 
{\bf Vila-u}~\cite{wu2024vila} is a unified foundation model for video, image, and language understanding and generation. Unlike traditional VLMs with separate modules, it employs a single autoregressive framework, simplifying architecture while achieving near state-of-the-art performance in both comprehension and generation.

\noindent 
{\bf Stable Diffusion 3.5 Large}~\cite{esser2024scaling} is a Multimodal Diffusion Transformer (MMDiT) text-to-image model that features improved performance in image quality, typography, complex prompt understanding, and resource-efficiency. It uses three fixed, pretrained text encoders, and with QK-normalization to improve training stability.

\noindent 
{\bf Show-o}~\cite{xie2024show} processes text tokens autoregressively with causal attention while handling image tokens using (discrete) denoising diffusion modeling via full attention. It then generates the desired output. Specifically, it is capable of performing image captioning, visual question answering, text-to-image generation, text-guided inpainting/extrapolation, and mixed-modality generation.

\noindent 
{\bf Janus}~\cite{wu2024janus} is a novel autoregressive framework that unifies multimodal understanding and generation. It addresses the limitations of previous approaches by decoupling visual encoding into separate pathways while still utilizing a single, unified transformer architecture for processing.

\noindent 
{\bf HART}~\cite{tang2024hart} introduces a hybrid tokenizer that enhances autoregressive (AR) models by improving image reconstruction quality and reducing training costs for high-resolution (1024px) image generation. It achieves this by decomposing the continuous latents from the autoencoder into two components: discrete tokens that capture the overall structure and continuous tokens that retain fine-grained residual details.

\noindent 
{\bf NOVA}~\cite{deng2024nova} is a model that enables autoregressive image/video generation with high efficiency. It reformulates the video generation problem as non-quantized autoregressive modeling of temporal frame-by-frame prediction and spatial set-by-set prediction. It generalizes well and enables diverse zero-shot generation abilities in one unified model.

\noindent 
{\bf Infinity}~\cite{Infinity} is a bitwise visual autoregressive model that adopts a novel token prediction framework with an infinite-vocabulary tokenizer and bitwise self-correction. By scaling the tokenizer vocabulary and transformer size concurrently, it enhances the model’s capacity for high-resolution image generation while maintaining fine-grained visual fidelity.

%% file: tables2/url.tex
\begin{table*}
\centering
  \caption{An overview and URLs of the adopted 24 text-to-image generation models.}
  \label{models}
    \renewcommand\arraystretch{1.3}
  \resizebox{\textwidth}{!}{\begin{tabular}{lcccccccl}
    \toprule
     Models&Type &Date& Resolution& URL\\
    \midrule 
    SD\_v2-1~\cite{Rombach_2022_CVPR} & Diff.& 2022.12&768$\times$768 & \url{https://huggingface.co/stabilityai/stable-diffusion-2-1} \\

    i-Code-V3~\cite{tang2023any} & Diff.&2023.05 & 256$\times$256& \url{https://github.com/microsoft/i-Code} \\
   
    SDXL\_base\_1~\cite{podell2023sdxl} &Diff. &2023.07 &1024$\times$1024 & \url{https://huggingface.co/stabilityai/stable-diffusion-xl-base-1.0} \\

    DALLE3~\cite{betker2023improving} & Diff.&2023.09 & 1024$\times$1024& \url{https://openai.com/index/dall-e-3} \\

    LLMGA~\cite{xia2024llmga} & Diff.&2023.11 & 1024$\times$1024& \url{https://github.com/dvlab-research/LLMGA} \\

    Kandinsky-3~\cite{arkhipkin2024kandinsky} &Diff. & 2023.12& 1024$\times$1024& \url{https://github.com/ai-forever/Kandinsky-3} \\

    LWM~\cite{liu2024world} &AR & 2024.01&256$\times$256 & \url{https://github.com/LargeWorldModel/LWM} \\
    
    Playground~\cite{li2024playground} &Diff. &2024.02 & 1024$\times$1024& \url{https://huggingface.co/playgroundai/playground-v2.5-1024px-aesthetic} \\

    LaVi-Bridge~\cite{zhao2024bridging} &Diff. &2024.03 &512$\times$512& \url{https://github.com/ShihaoZhaoZSH/LaVi-Bridge} \\

    ELLA~\cite{hu2024ella} &Diff. & 2024.03&512$\times$512 & \url{https://github.com/TencentQQGYLab/ELLA} \\

    Seed-xi~\cite{ge2024seed} &Diff. & 2024.04&1024$\times$1024 & \url{https://github.com/AILab-CVC/SEED-X} \\

    PixArt-sigma~\cite{chen2024pixart} &Diff. &2024.04 & 1024$\times$1024& \url{https://github.com/PixArt-alpha/PixArt-sigma} \\
    LlamaGen~\cite{sun2024autoregressive} &AR &2024.06 &256$\times$256 & \url{https://github.com/FoundationVision/LlamaGen} \\
    
        Kolors~\cite{kolors} &Diff. & 2024.07&1024$\times$1024 & \url{https://github.com/Kwai-Kolors/Kolors} \\
        
    Flux\_schnell~\cite{flux2024} &Diff. & 2024.08&1024$\times$1024 & \url{https://huggingface.co/black-forest-labs/FLUX.1-schnell} \\

    Omnigen~\cite{xiao2024omnigen} &Diff. & 2024.09& 1024$\times$1024& \url{https://github.com/VectorSpaceLab/OmniGen} \\

    EMU3~\cite{wang2024emu3} &AR &2024.09 &720$\times$720 & \url{https://github.com/baaivision/Emu} \\

    Vila-u~\cite{wu2024vila} &AR &2024.09 & 256$\times$256& \url{https://github.com/mit-han-lab/vila-u} \\

    SD3\_5\_large~\cite{esser2024scaling} &Diff. &2024.10 &1024$\times$1024 & \url{https://huggingface.co/stabilityai/stable-diffusion-3.5-large} \\

    Show-o~\cite{xie2024show} & AR+Diff.&2024.10 & 512$\times$512& \url{https://github.com/showlab/Show-o} \\
    
    Janus~\cite{wu2024janus} &AR & 2024.10& 384$\times$384& \url{https://github.com/deepseek-ai/Janus} \\
    
    Hart~\cite{tang2024hart} &AR & 2024.10&1024$\times$1024 & \url{https://github.com/mit-han-lab/hart} \\

    NOVA~\cite{deng2024nova} &AR & 2024.12&512$\times$512 & \url{https://github.com/baaivision/NOVA} \\

    Infinity~\cite{Infinity} &AR & 2024.12&1024$\times$1024 & \url{https://github.com/FoundationVision/Infinity} \\

    \bottomrule
  \end{tabular}}
\end{table*}

%% file: sec2/4_exp.tex
\section{More Details of Subjective Experiment}
\label{exp}
\subsection{Annotaion Dimension and Criteria}
To comprehensively assess the performance of AI-generated images (AIGIs), we propose a dual-dimensional evaluation framework that examines both perceptual quality and text-to-image (T2I) correspondence. This approach enables a thorough analysis of different aspects of image generation, providing a holistic understanding of a model's capabilities and limitations.

\begin{itemize}
\item \textbf{Perceptual quality} evaluates the visual characteristics and aesthetic appeal of generated images. This dimension focuses on multiple aspects of image quality, including \textbf{visual clarity} (the sharpness and resolution of image details), \textbf{naturalness} (the degree to which the image appears realistic and free from artifacts), \textbf{aesthetic appeal} (the composition, color harmony, and overall visual attractiveness), \textbf{structural coherence} (the logical consistency of spatial relationships and object proportions), and \textbf{authenticity} (whether the generated image is realistic). High-scoring images are characterized by exceptional clarity, vivid and well-balanced colors, and meticulous attention to detail, offering an immersive and visually striking experience. In contrast, low scores reflect images with blurriness, unnatural color tones, faded visuals, and a lack of clarity or detail. This dimension captures the foundational visual attributes that make an image aesthetically pleasing or distracting. For detailed criteria, refer to Figure \ref{sup_bz1}.

\item \textbf{Text-image correspondence} assesses the semantic alignment between the generated image and the input text prompt, including \textbf{content accuracy} (the presence and correct representation of described objects and elements), \textbf{contextual relevance} (the appropriate depiction of scenes and relationships between objects), \textbf{attribute fidelity} (the accurate representation of specific characteristics mentioned in the prompt), and \textbf{semantic consistency} (the logical coherence between visual elements and textual descriptions). Images with high scores perfectly match the descriptions in the prompt, accurately reflecting all elements with high fidelity. These images effectively translate textual information into visual content without mismatches. In contrast, images with lower scores exhibit inconsistencies, missing elements, or mismatched content. For detailed criteria, refer to Figure \ref{sup_bz2}.
\end{itemize}

\subsection{Significance of the Two Dimensions}
\input{figures2/dimension}
The dual-dimensional evaluation framework, which combines perception quality and T2I correspondence, is essential for addressing the inherent trade-offs and complementary aspects of AIGIs. While perception quality emphasizes the visual characteristics that contribute to an image's appeal and realism, T2I correspondence ensures that the generated content remains semantically faithful to the original textual description. Together, these dimensions provide a comprehensive assessment of both the aesthetic and functional aspects of image generation.
As illustrated in Figure \ref{dimension}, a high perception quality score alone does not guarantee semantic accuracy. For example, an image may exhibit exceptional visual quality, characterized by high resolution, vibrant colors, and meticulous detail, yet fail to accurately represent the specific objects, relationships, or attributes described in the text prompt. Conversely, an image may perfectly align with the textual description in terms of content and context but suffer from poor visual quality, such as low resolution, unnatural textures, or inconsistent lighting, which detracts from its overall appeal and usability.
The integration of both dimensions ensures that generated images achieve a balance between visual excellence and semantic fidelity. This holistic approach not only enhances the evaluation of generative models but also aligns with real-world applications where both image quality and content accuracy are critical. By considering both dimensions, the framework provides a more nuanced understanding of a model's strengths and weaknesses, facilitating targeted improvements in image generation systems.

\subsection{Annotation Interface}
To ensure a comprehensive and efficient image quality evaluation, we design two custom annotation interfaces tailored for different assessment tasks: simple task annotation and complex task annotation. The simple task annotation interface, shown in Figure \ref{ui1}, is a manual evaluation platform developed using the Python tkinter package, designed to facilitate MOS assessments. 
The experiment involves evaluating images based on two independent dimensions and answering a binary question related to a specific task-specific challenge. There are 20 task-specific challenges, including categories such as human, shape, scene, color, etc.
Each trial presents three images that correspond to the same prompt. These images are randomly selected from 24 different models. Importantly, participants are instructed to assign absolute scores to each image on the two predefined dimensions, rather than making relative comparisons between the images.
For each image, participants provide:
(1) Two separate scores representing the two evaluation dimensions.
(2) A binary response (yes/no) to indicate whether the image meets the specified challenge criterion.
 Meanwhile, the complex task annotation interface, is illustrated in Figure \ref{ui2}. The complex tasks are composed of multiple subtasks such as Number, Color, Shape, and Scene. Each subtask is evaluated independently with a yes/no response. The complex task is considered correct only if all its sub-tasks are correct. If any sub-task is incorrect, the entire complex task is marked as incorrect. To ensure uniformity and minimize resolution-related biases in image quality evaluation, all images displayed in this interface are cropped to a spatial resolution of 1024×1024 pixels.  Navigation options, such as “Previous” and “Next” streamline the workflow, enabling efficient annotation.
\input{figures2/ui1}
\input{figures2/ui2}
\subsection{Annotation Management}
To ensure ethical compliance and the quality of annotations, we implement a comprehensive process for the EvalMi-50K dataset. All participants are fully informed about the experiment's purpose, tasks, and ethical considerations. Each participant sign an informed consent agreement, granting permission for their subjective ratings to be used exclusively for noncommercial research purposes. The dataset, comprising 50,400 AIGIs alongside their corresponding prompts, has been publicly released under the CC BY 4.0 license, ensuring accessibility while adhering to ethical guidelines.
We ensure the exclusion of all inappropriate or NSFW content (textual or visual) through a rigorous manual review process during the image generation phase. This step ensures that the dataset remains suitable for academic and research use.
The annotation process is structured into two primary components: Mean Opinion Score (MOS) annotation and task-specific question-answering (QA) annotation. Each component is designed to evaluate images across 20 task-specific challenges, including color, position, shapes, view, and \textit{etc.}
The MOS annotation task involves 16 participants to rate each image on a 0-5 Likert scale, assessing both perception quality and T2I correspondence. The question-answering annotation task is similarly conducted with 16 participants, ensuring consistency in the evaluation process. In this task, participants are presented with a series of yes/no questions across the 20 task-specific challenges. To determine the final answer for each question, a majority voting mechanism is employed. This approach ensures that the final decision reflects the collective judgment of the participants, minimizing the impact of individual biases or errors. 

Prior to engaging in the annotation tasks, all participants undergoes a rigorous training process. As illustrated in Figures \ref{sup_bz1}-\ref{sup_bz2}, they are provided with detailed instructions and multiple standardized examples. To ensure a high level of understanding and consistency, a pre-test is conducted to evaluate participants' comprehension of the criteria and their alignment with the standard examples. Participants who do not meet the required accuracy threshold are excluded from further participation, ensuring that only well-prepared individuals contribute to the final dataset.
During the experiment, all evaluations are conducted in a controlled laboratory environment under normal indoor lighting conditions. Participants are seated at a comfortable viewing distance of approximately 60 cm from the screen to minimize visual strain and ensure consistent evaluation conditions. 
While individual preferences may naturally vary, the use of detailed explanations and standardized annotation criteria ensure a high degree of agreement among participants. This consensus is particularly evident in question-answering annotations, where majority voting effectively captures group preferences.  This rigorous and ethically sound annotation management strategy establishes EvalMi-50K as a robust and reliable resource for advancing research in image quality assessment. 

%% file: figures2/dimension.tex
\begin{figure}[!t]
	\centering
	\includegraphics[width=\linewidth]{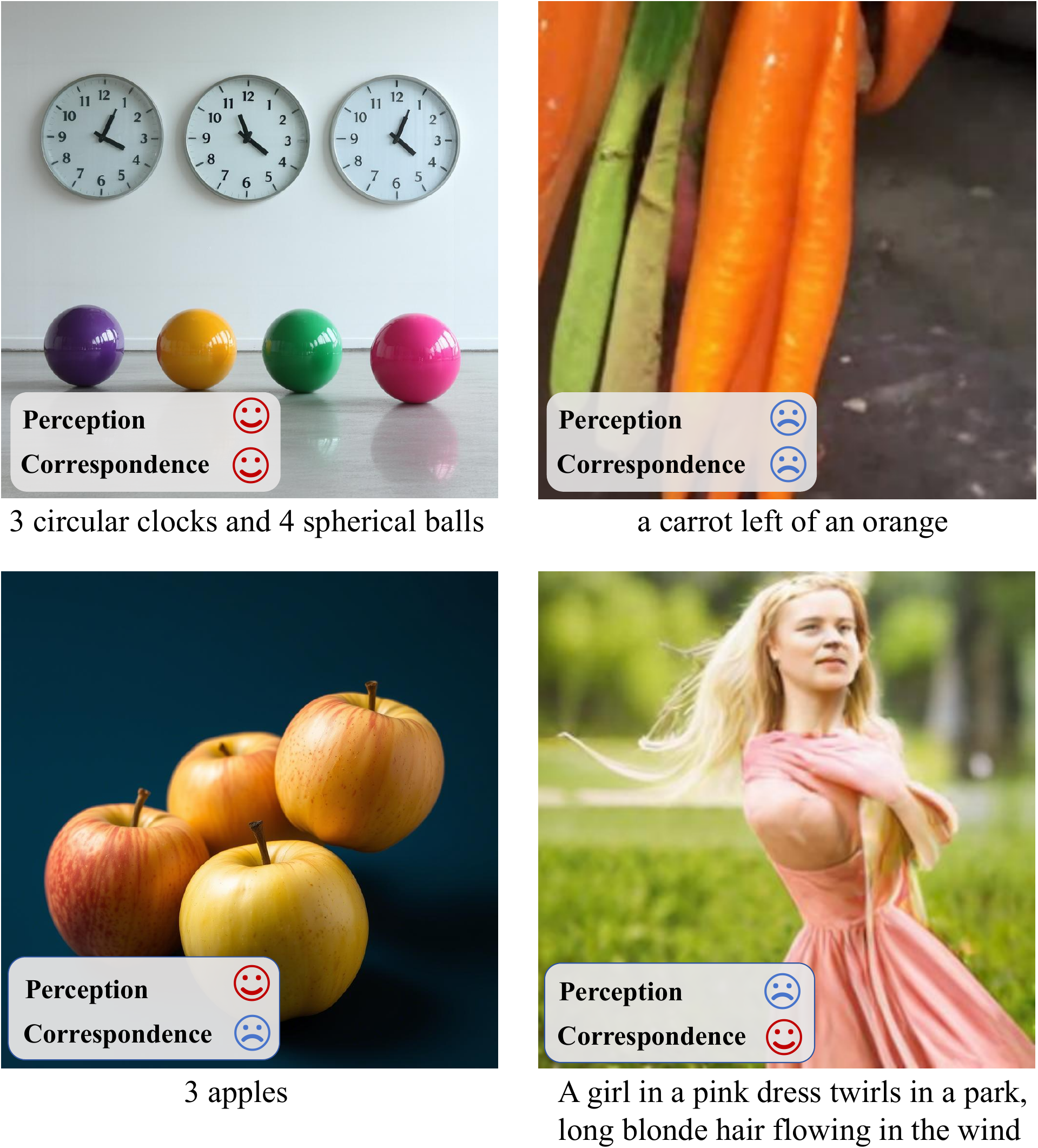}
 \vspace{-6mm}
	\caption{Illustration of the evaluation dimensions: perceptual quality and text-image correspondence, attached with examples with different subjective qualities.}
	\label{dimension}
\end{figure}

%% file: figures2/ui1.tex
\begin{figure*}[!t]
\vspace{-4mm}
	\centering
	\includegraphics[width=\linewidth]{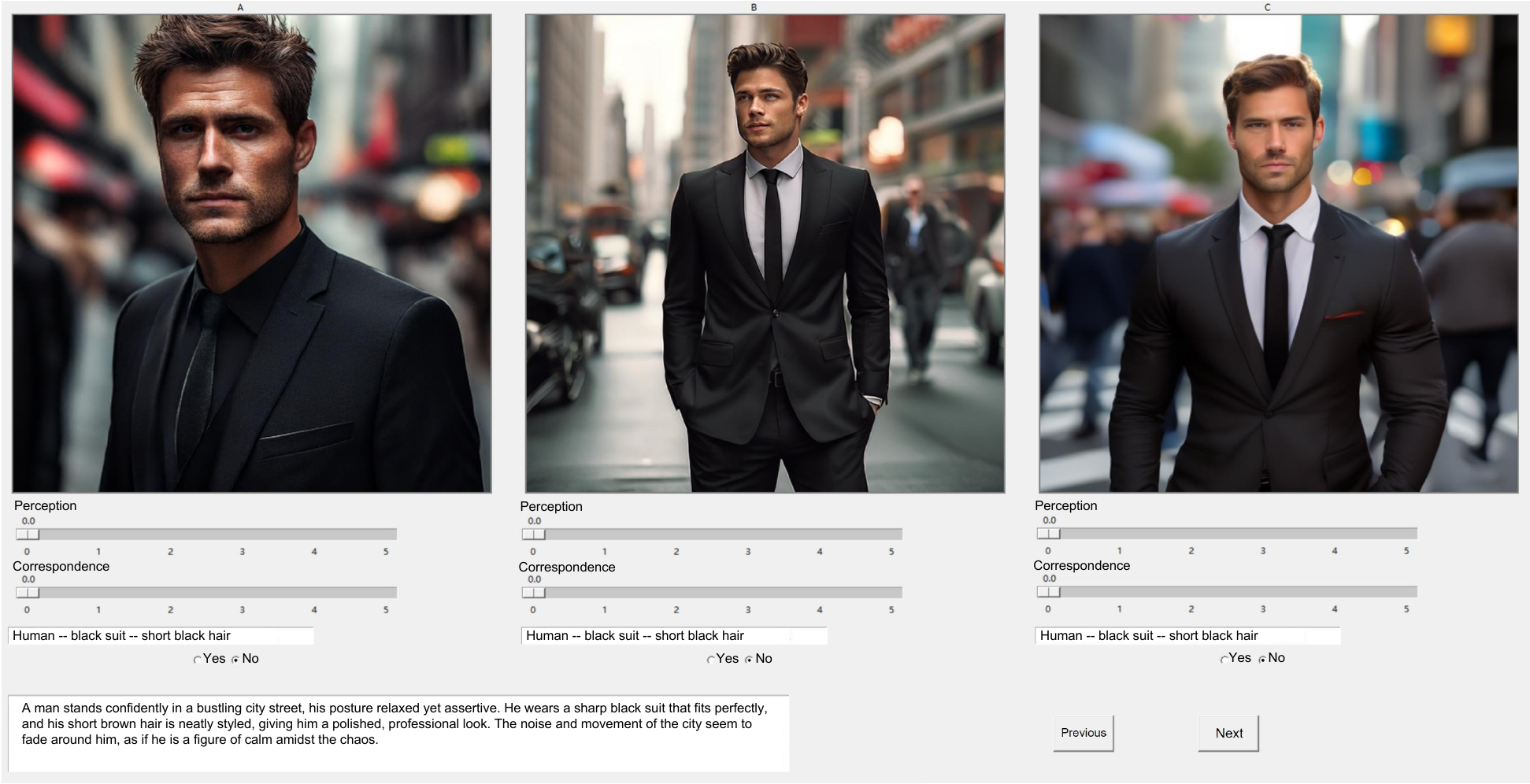}
 \vspace{-5mm}
	\caption{ An example of the simple task annotation interface for human evaluation. The subjects are instructed to rate two dimensions of AI-generated images, \textit{i.e.}, perception and text-image correspondence, and provide a binary (yes/no) response for a task-specific challenge. Each trial presents three images generated from 24 models for the same prompt, with absolute scoring applied independently to each image.}
	\label{ui1}
\end{figure*}

%% file: figures2/ui2.tex
\begin{figure*}[!t]
	\centering
	\includegraphics[width=\linewidth]{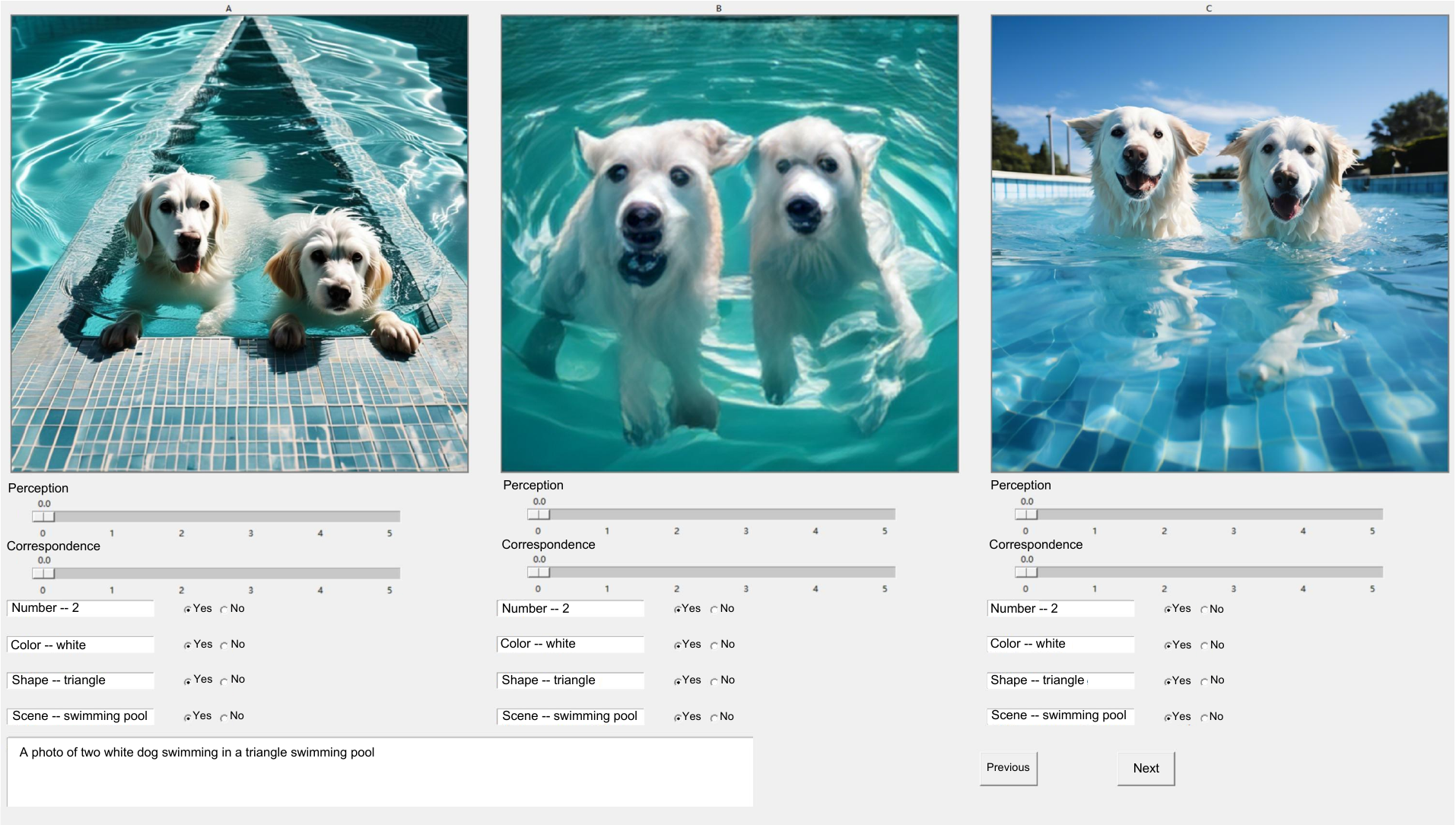}
 \vspace{-5mm}
	\caption{ An example of the complex task annotation interface, which extends the simple task evaluation by incorporating multiple sub-tasks (\textit{e.g.}, Number, Color, Shape, and Scene). The subjects are instructed to rate two dimensions of AI-generated images, \textit{i.e.}, perception and text-image correspondence, based on the given image and its prompt. Each sub-task is judged independently with a yes/no response. The complex task is considered correct only if all sub-tasks are correctly identified; if any sub-task is incorrect, the entire complex task is marked as incorrect. }
 \vspace{-3mm}
	\label{ui2}
\end{figure*}

%% file: sec2/5_analyze.tex
\section{More Analysis of EvalMi-50K Database}
\label{analyze}
\subsection{MOS Distribution across 20 Challenges}
\input{figures2/20task1}
As mentioned in the main text, we process and compute the valid subjective evaluation results, obtaining a total of 100,800 Mean Opinion Scores (MOSs) across two dimensions, along with QA accuracy. To better illustrate the generative capabilities of current T2I models in different prompt challenges, we categorize the computed MOSs data into 20 task categories and used the categorized data to plot histograms and kernel density curves (KDC) graphs, as shown in Figure \ref{sup_20task1}.
We can observe that the 24 T2I models we tested exhibit relatively poor text-image alignment in prompt challenges related to position, OCR, linguistic structures, and complexity, with MOSs primarily clustering around 30. In contrast, their performance in other prompt challenges is relatively better. The overall perception MOSs does not show significant differences across different prompt challenges, with scores generally concentrated at a higher level. However, models perform slightly worse in OCR, HOI, and Face-related prompt challenges, where lower MOSs appear more frequently compared to other prompt challenges.

\subsection{T2I Model Performance across 20 Challenges}
Tables \ref{mos1}-\ref{mos3} provide detailed performance comparisons of the 24 T2I models across 20 task-specific challenges on three types of human annotations: perception MOS, T2I correspondence MOS, and question-answering accuracy. 
For perception quality, as demonstrated in Table \ref{mos1} and Figure \ref{perception}-\ref{perception2}, models like Playground~\cite{li2024playground} stand out with the highest MOS and perform particularly well in categories such as “Colors,” “Shapes,” and “Scene”. These models excel in generating images that are visually appealing, realistic, and aesthetically pleasing.  
For T2I correspondence, as demonstrated in Table \ref{mos2} and Figure \ref{correspondence}-\ref{correspondence2}, SD3\_5\_large~\cite{esser2024scaling} leads the way, demonstrating strong alignment between the generated images and the textual descriptions, but has a relatively lower performance in perception quality. Conversely, models like Kolors~\cite{kolors} excel in perception quality, delivering high MOS scores, but can not perform as well in terms of T2I correspondence.
The contrasting trends in performance between perception quality and T2I correspondence emphasize the importance of evaluating both dimensions independently. While perception quality focuses on the visual aspects of the generated images, T2I correspondence measures how well the image aligns with the content described in the text prompt. This dual evaluation ensures a more comprehensive understanding of a model’s abilities, where one dimension evaluates aesthetic quality, and the other checks the accuracy of the image-text alignment.
In terms of task-specific accuracy, as demonstrated in Table \ref{mos3}, the ranking of models largely mirrors the performance in T2I correspondence. Since task-specific accuracy is inherently tied to T2I correspondence, models that excel in faithfully translating text into images also tend to perform well in answering specific questions related to those images. While task-specific accuracy provides binary (0/1) assessments based on task-specific queries, MOS offers continuous scoring that enables a more granular evaluation of the text-image correspondence, providing deeper insights into how accurately a model generates images in relation to the given prompt, beyond a simple binary judgment.

%% file: figures2/20task1.tex
\begin{figure*}[!t]
	\centering
	\includegraphics[width=1\linewidth]{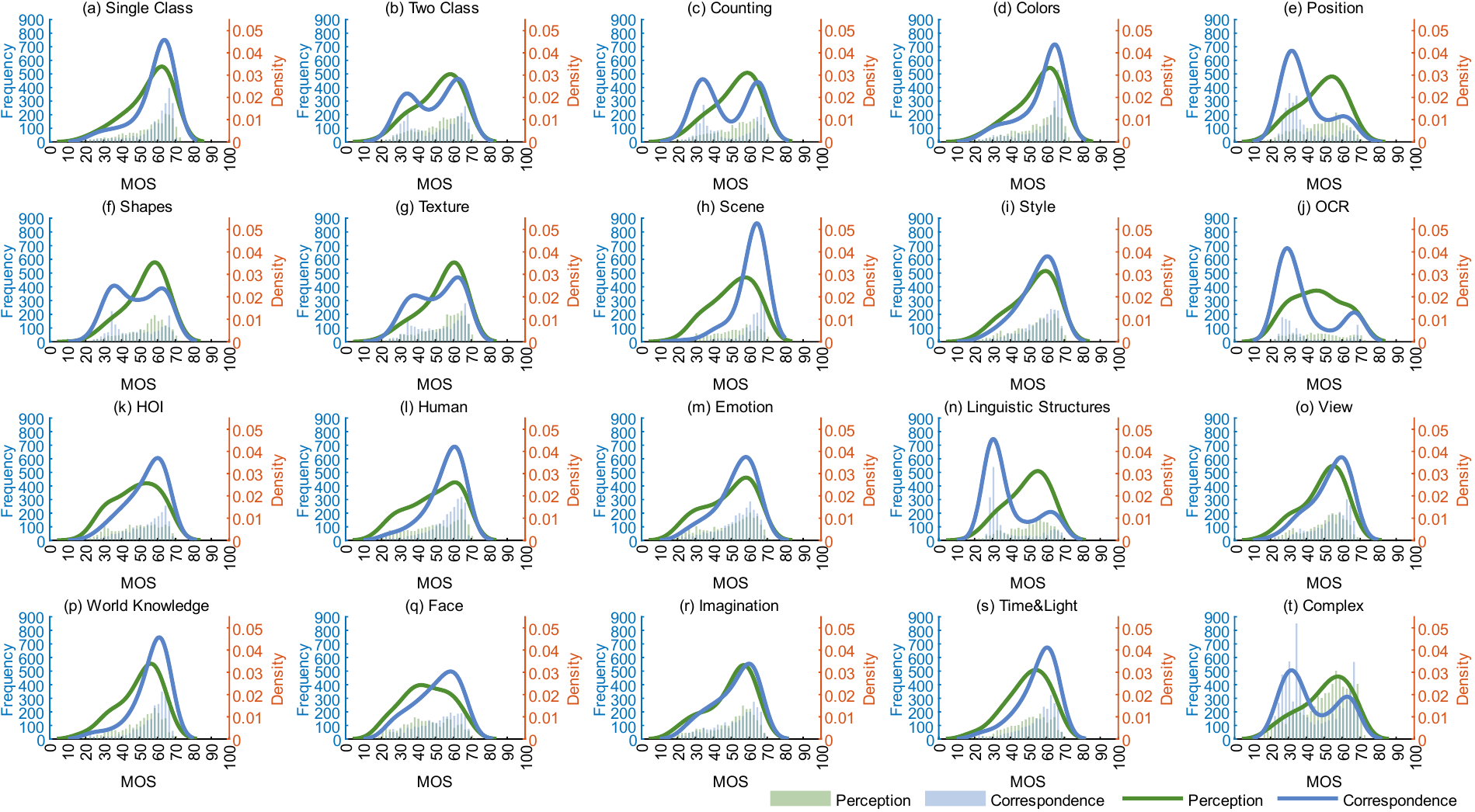}
 \vspace{-1mm}
	\caption{\textbf{Mean Opinion Score (MOS) distribution histograms and kernel density curves} of EvalMi-50K dataset. It includes two dimensions: Perception MOS and Correspondence MOS. Each dimension contains a total of 50,400 MOS values.}
	\label{sup_20task1}
\end{figure*}

%% file: sec2/6_loss.tex
\section{Details of Loss Function}
\label{loss}
The training process for LMM4LMM is divided into two progressive stages, each utilizing a specific loss function to target distinct objectives: language loss for instruction tuning, aligning visual and language features to give visual question answers across the 20 task-specific challenges, L1 loss for quality regression fine-tuning to generate accurate perception and correspondence scores.

\paragraph{(1) Instruction tuning  with language loss.}
In the first stage, we train the projector to align visual and language features using the standard language loss. This involves ensuring that the visual tokens extracted from the vision encoder correspond effectively to the language representations from the LLM. The language loss, calculated using a cross-entropy function, measures the model’s ability to predict the correct token given the prior context:
\begin{eqnarray}
\begin{aligned}
&\mathcal{L}_{\text{language}} = -\frac{1}{N} \sum_{i=1}^N \log P(y_{\text{label}} | y_{\text{pred}})
\end{aligned}
\end{eqnarray}
where $P(y_{\text{label}} | y_{\text{pred})}$ represents the probability assigned to the correct token $y_{\text{label}}$ by the model, $y_{\text{pred}}$ is the predicted token, and $N$ is the total number of tokens. By minimizing this loss, the model learns to generate coherent textual descriptions of image content, laying the foundation for subsequent stages.

\paragraph{(2) Refining quality scoring with L1 loss.}
Once the model can produce coherent descriptions of image content, the focus shifts to fine-tuning the quality regression module to output stable and precise numerical quality scores. The quality regression module takes the aligned visual tokens as input and predicts a quality score that reflects the overall image quality.
Using the EvalMi-50K, which contains human-annotated MOS for each image, the model is trained to align its predictions with human ratings. The training objective minimizes the difference between the predicted quality score $Q_{predict}$ and the ground-truth MOS  $Q_{label}$ using the L1 loss function:
\begin{eqnarray}
\begin{aligned}
\label{loss_function}
\mathcal{L}_{\text{MOS}} = \frac{1}{N} \sum_{i=1}^N \left| Q_{\text{predict}}(i) - Q_{\text{label}}(i) \right|
\end{aligned}
\end{eqnarray}
where $Q_{\text{predict}}(i)$ is the score predicted by the regressor $i$ and $Q_{\text{label}}(i)$ is the corresponding ground-truth MOS derived from subjective experiments, and $N$  is the number of images in the batch. This loss function ensures that the predicted scores remain consistent with human evaluations, enabling the model to accurately assess the quality of AI-generated images in numerical form.

%% file: sec2/7_methods.tex
\section{Implemention Details}
\label{implementation}

\subsection{Detailed Information of Evaluation Criteria}
\noindent
We adopt the widely used metrics in IQA literature \cite{sun2023blind, wu2023qalign,zhang2023liqe}: Spearman rank-order correlation coefficient (SRCC), Pearson linear correlation coefficient (PLCC), and Kendall’s Rank Correlation Coefficient (KRCC) as our evaluation criteria. SRCC quantifies the extent to which the ranks of two variables are related, which ranges from -1 to 1. Given $N$ action images, SRCC is computed as:
\begin{equation}
SRCC = 1 - \frac{{6\sum\nolimits_{n = 1}^N {{{({v_n} - {p_n})}^2}} }}{{N({N^2} - 1)}},
\end{equation}
where $v_n$ and $p_n$ denote the rank of the ground truth $y_n$ and the rank of predicted score ${\hat y_n}$ respectively. The higher the SRCC, the higher the monotonic correlation between ground truth and predicted score.
Similarly, PLCC measures the linear correlation between predicted scores and ground truth scores, which can be formulated as:
\begin{equation}
PLCC = \frac{{\sum\nolimits_{n = 1}^N {({y_n} - \bar y)({{\hat y}_n} - \bar {\hat y})} }}{{\sqrt {\sum\nolimits_{n = 1}^N {{{({y_n} - \bar y)}^2}} } \sqrt {\sum\nolimits_{n = 1}^N {{{({{\hat y}_n} - \bar {\hat y})}^2}} } }},
\end{equation}
where $\bar y$ and $\bar {\hat y}$ are the mean of ground truth and predicted score respectively.

\noindent
We also adopt the Kendall Rank Correlation Coefficient (KRCC) as an evaluation metric, which measures the ordinal association between two variables. For a pair of ranks $(v_i, p_i)$ and $(v_j, p_j)$, the pair is concordant if:
\begin{equation}
(v_i - v_j)(p_i - p_j) > 0,
\end{equation}
and discordant if $<  $ 0.
 Given $N$ AIGVs, KRCC is computed as:
\begin{equation}
KRCC =  \frac{{C - D}}{{\frac{1}{2}N(N-1)}},
\end{equation}
where $C$ and $D$ denote the number of concordant and discordant pairs, respectively.

\subsection{Detailed Information of Evaluation Methods}
\noindent
{\bf LLaVA-1.5}~\cite{liu2024improved} is an advanced Large Multimodal Model (LMM) framework designed for visual instruction tuning, aimed at improving multimodal understanding capabilities for general-purpose assistants. The model builds upon the LLaVA architecture and uses a simple fully-connected vision-language connector, making it more data-efficient. 

\noindent
{\bf LLaVA-NeXT}~\cite{liu2024llavanext} improves on LLaVA-1.5~\cite{liu2024improved} by increasing input image resolution and enhances visual detail, reasoning, and OCR capabilities. It also improves world knowledge and logical reasoning while maintaining LLaVA's minimalist design and data efficiency, using under 1M visual instruction tuning samples. 

\noindent
{\bf mPLUG-Owl3}~\cite{ye2024mplug} is a versatile multi-modal large language model designed to handle long image sequences, interleaved image-text, and lengthy video inputs. It introduces Hyper Attention blocks that efficiently integrate vision and language into a shared semantic space, allowing for the processing of extended multi-image scenarios. 

\noindent
{\bf MiniCPM-V2.6}~\cite{yao2024minicpm} is designed for deployment on end-side devices, addressing the challenges of running large models with significant computational costs. Key features include strong OCR capability, supporting high-resolution image perception, trustworthy behavior with low hallucination rates, and multilingual support for over 30 languages. 

\noindent
{\bf Qwen2-VL}~\cite{Qwen2-VL}
is an advanced large vision-language model designed to process images, videos, and text with dynamic resolution handling and multimodal rotary position embedding (M-RoPE). The model features strong capabilities in OCR, video comprehension, multilingual support, and robust agent functionalities for device operations. 

\noindent
{\bf Qwen2.5-VL}~\cite{Qwen2.5-VL}
 is the latest flagship model in the Qwen vision-language series, featuring significant improvements in visual recognition, object localization, document parsing, and long-video comprehension. Building on the Qwen2-VL architecture, it introduces key enhancements such as dynamic resolution processing for images and videos, absolute time encoding for temporal dynamics, and window attention to optimize inference efficiency. 
 
\noindent
{\bf Llama3.2-Vision}~\cite{meta2024llama}
 excels in image reasoning tasks, such as document-level understanding, chart and graph captioning, and visual grounding. These models can reason with images, such as answering questions based on graphs or maps, and generate captions that describe visual scenes. 
 
\noindent
{\bf DeepseekVL}~\cite{lu2024deepseekvl}
 leverages a hybrid vision encoder for efficient high-resolution image processing and a carefully balanced training strategy that integrates language model capabilities with vision tasks. By emphasizing diverse, real-world data and a use case taxonomy, DeepSeek-VL delivers superior performance in tasks like OCR, document parsing, and visual-grounding. 
 
\noindent
{\bf DeepseekVL2}~\cite{wu2024deepseekvl2mixtureofexpertsvisionlanguagemodels}
is an advanced series of mix-of-experts (MoE) vision language models. It introduces a dynamic tiling vision encoding strategy, allowing efficient processing of high-resolution images with varying aspect ratios, enhancing tasks like visual grounding and document analysis. It also leverages the Multi-head Latent Attention (MLA) mechanism for the language component, which reduces computational costs and improves inference efficiency. 

\noindent
{\bf CogAgent}~\cite{hong2024cogagentvisuallanguagemodel}
is designed to facilitate understanding and navigation of graphical user interfaces (GUIs). It utilizes both low and high-resolution image encoders to recognize small text and page elements. CogAgent excels in GUI tasks like navigation and decision-making. CogAgent’s innovative design includes a cross-attention branch to balance high-resolution inputs and computational efficiency. 

\noindent
{\bf InternVL2.5}~\cite{chen2024expanding}
 demonstrates strong performance in various benchmarks, including multi-discipline reasoning, document and video understanding, and multimodal hallucination detection. The model features enhanced vision encoders, larger dataset sizes, and improved test-time scaling.
 
\noindent
{\bf InternLM-XComposer}~\cite{internlmxcomposer}
excels at generating long-form content that integrates contextually relevant images, enhancing the engagement and immersion of the reading experience. It autonomously identifies optimal locations in the text for image placement and selects appropriate images from a large-scale database, ensuring contextual alignment. 

\noindent
{\bf CLIPScore}~\cite{hessel2021clipscore} is an image captioning metric, which is widely used to evaluate T2I/T2V models. It passes both the image and the candidate caption through their respective feature extractors, then computing the cosine similarity between the text and image embeddings.

\noindent
{\bf BLIPScore}~\cite{li2022blip} provides more advanced multi-modal feature extraction capabilities. Using the same methodology as CLIPScore ~\cite{hessel2021clipscore}, it computes the cosine similarity between the text and visual embeddings, but benefits from enhanced pre-training strategy, which is designed to better capture fine-grained relationships between text and visual content.

\noindent
{\bf ImageReward}~\cite{xu2023imagereward} builds upon the BLIP model~\cite{li2022blip} by introducing an additional MLP layer on top of BLIP’s output. Instead of directly computing a similarity score, the MLP generates a scalar value representing the preference for one image over another in comparative settings. 

\noindent
{\bf PickScore}~\cite{kirstain2023pick}
is a scoring function designed to predict human preferences in text-to-image generation. It was trained by fine-tuning CLIP-H on human preference data, aiming to maximize the probability of a preferred image being selected. PickScore exhibits strong correlation with human rankings, outperforming traditional metrics like FID and aesthetics predictors, and is recommended as a more reliable evaluation metric for text-to-image models.

\noindent
{\bf HPS}~\cite{wu2023human}
 is designed to improve text-to-image generation models by better aligning their outputs with human preferences. HPS is based on a fine-tuned CLIP model that accurately predicts human preferences over generated images. 
 
\noindent
{\bf VQAScore}~\cite{li2024evaluating}
 is designed to assess the alignment between generated images and text prompts, particularly for compositional text-to-visual generation tasks. It can be used in a black-box manner, requiring no fine-tuning or additional prompt decomposition. 
 
\noindent
{\bf FGA-BLIP2}~\cite{han2024evalmuse40kreliablefinegrainedbenchmark}
is a method for evaluating image-text alignment in T2I models, specifically designed to provide fine-grained analysis. It involves fine-tuning a vision-language model to produce alignment scores and element-level annotations for image-text pairs. This approach uses a variance-weighted optimization strategy to account for the diversity of images generated from specific prompts. 

\noindent
{\bf CNNIQA}~\cite{kang2014convolutional}
is a convolutional neural network (CNN) designed for no-reference image quality assessment (NR-IQA), which predicts the visual quality of distorted images without using reference images. Unlike traditional methods that rely on handcrafted features, CNNIQA directly learns discriminative features from raw image patches, allowing for a more efficient and effective image quality estimation. 

\noindent
{\bf DBCNN}~\cite{quality:DBCNN}
 is a deep bilinear convolutional neural network designed for blind image quality assessment, which handles both synthetic and authentic distortions. The model uses two specialized convolutional neural networks. The features from both CNNs are pooled bilinearly into a unified representation for quality prediction.
 
\noindent
{\bf HyperIQA}~\cite{su2020blindly}
aims at handling authentically distorted images. It addresses two main challenges: distortion diversity and content variation. The model is based on a self-adaptive hyper network that adjusts quality prediction parameters according to the image content, making the predictions more consistent with human perception. 

\noindent
{\bf TReS}~\cite{golestaneh2021no}
handles both synthetic and authentic distortions. It combines CNNs for capturing local image features with the self-attention mechanism to learn non-local features, addressing both local and global image quality aspects. The model also incorporates a relative ranking loss to enhance the correlation between subjective and objective scores by learning the relative quality ranking among images. 

\noindent
{\bf MUSIQ}~\cite{ke2021musiq}
leverages a patch-based multi-scale Transformer architecture to handle images of varying resolutions and aspect ratios without resizing or cropping. Unlike CNN-based models, which require fixed-size input, MUSIQ can process full-size images, extracting features at multiple scales to capture both fine-grained and global image quality details. The model introduces a unique hash-based 2D spatial embedding and scale embedding to effectively manage positional information across multi-scale inputs. 

\noindent
{\bf StairIQA}~\cite{sun2023blind}
employs a staircase structure that hierarchically integrates features from intermediate layers of a CNN, allowing it to leverage both low-level and high-level visual information for more effective quality assessment. Additionally, it introduces an Iterative Mixed Database Training (IMDT) strategy, which trains the model across multiple diverse databases to improve generalization and handle variations in image content and distortions. 

\noindent
{\bf Q-Align}~\cite{wu2023qalign} is a human-emulating syllabus designed to train large multimodal models for visual scoring tasks. It mimics the process of training human annotators by converting MOS into five text-defined rating levels. We used the officially pre-trained model and finetuned it on our EvalMi-50K.

\noindent
{\bf LIQE}~\cite{zhang2023liqe}
 integrates auxiliary tasks such as scene classification and distortion type identification to improve the quality prediction of in-the-wild images. It uses a textual template to describe the image's scene, distortion, and quality, using CLIP to compute the joint probability of these tasks.
\subsection{Question design for LLM-based models}
For LLM-based detection methods, we not only need to input the image to be evaluated, but also the corresponding prompt to guide the model to output the result we want. Three different questions need to be input for each image to be evaluated. When designing questions from the two dimensions of Perception and T2I correspondence, all images have a unified template, but to obtain the question-answer pair for an image, different questions need to be designed according to the challenge corresponding to the prompt used to generate the image. We have a total of 20 tasks, so there are 20 question models for this dimension. The specific question template is as follows:
\begin{itemize}
    \item \textbf{Perception}:
    Suppose you are now a volunteer for subjective quality evaluation of images and you are now required to rate the quality of the given images on a scale of 0-100. Results are accurate to the nearest digit. Answer only one score.
    \item \textbf{T2I Correspondence}:
Please rate the consistency between the image and the text description “\textless prompt \textgreater”. The rating scale is from 0 to 100, with higher scores for descriptions that include important content from the image and lower scores for descriptions that lack important content. Results are accurate to the nearest digit. Answer only a score.
    \item \textbf{Question-Answer Pairs}:

    \begin{itemize}
    \item[(1)] \textbf{Single class}: Does the image contain \textless class\_name \textgreater? Answer yes or no.
    \item[(2)] \textbf{Two class}: Does the image contain both \textless class1\_name \textgreater and \textless class2\_name \textgreater? Answer yes or no.
    \item[(3)] \textbf{Counting}: Does the image contain \textless class\_count \textgreater \textless class\_name \textgreater? Answer yes or no.
    \item[(4)] \textbf{Colors}: Does the image contain \textless class\_name \textgreater in the color of \textless class\_color \textgreater? Answer yes or no.
    \item[(5)] \textbf{Position}: Does the image contain both \textless class1\_name \textgreater and \textless class2\_name \textgreater, and are they positioned as described in “\textless prompt \textgreater”? Answer yes or no.
    \item[(6)] \textbf{Shapes}: Does the image contain a \textless class\_shape \textgreater \textless class\_name \textgreater? Answer yes or no.
    \item[(7)] \textbf{Texture}: Does the image contain a \textless class\_texture \textgreater \textless class\_name \textgreater? Answer yes or no.
    \item[(8)] \textbf{Scene}: Does the image depict a \textless scene\_name \textgreater scene? Answer yes or no.
    \item[(9)] \textbf{Style}: Is the style of the image \textless style\_name \textgreater? Answer yes or no.
    \item[(10)] \textbf{OCR (Optical Character Recognition)}: Does the image contain the text “\textless OCR \textgreater” with all letters correct? Answer yes or no.
    \item[(11)] \textbf{HOI (Human-Object Interaction)}: Does the image contain both a person and \textless object\_name \textgreater, and is the person's action \textless verb\_ing \textgreater? Answer yes or no.
    \item[(12)] \textbf{Human}: Do the appearance, hairstyle, accessories, and profession of the person in the image match the description in “\textless prompt \textgreater”? Answer yes or no.
    \item[(13)] \textbf{Emotion}: If there is a person in the image, is their emotion \textless emotion\_class \textgreater? If there is no person, does the overall mood of the image convey \textless emotion\_class \textgreater? Answer yes or no.
    \item[(14)] \textbf{Linguistic Structure}: Does the scene depicted in the image exclude \textless class\_name \textgreater? Answer yes or no.
    \item[(15)] \textbf{View}: Is the perspective shown in the image \textless view\_class \textgreater? Answer yes or no.
    \item[(16)] \textbf{World Knowledge}: Does the image contain a famous landmark or celebrity \textless knowledge\_class \textgreater? Answer yes or no.
    \item[(17)] \textbf{Face}: Does the face in the image have \textless first\_body\_part
 \textgreater \textless first\_shape\_or\_color\textgreater and\textless second\_body\_part \textgreater  \textless second\_shape\_or\_color \textgreater? Answer yes or no.
    \item[(18)] \textbf{Imagination}: Does the image content show imaginative elements, and does it match the description in “\textless prompt \textgreater”? Answer yes or no.
    \item[(19)] \textbf{Time \& Light}: Does the image depict the time \textless time\_class \textgreater with sunlight appearing as \textless ligth\_class \textgreater? Answer yes or no.
    \item[(20)] \textbf{Complex}: The questions for a complex challenge are a combination of the questions for the 19 individual challenges described above. For example, for a complex challenge consisting of a combination of task 1, task 2, etc., the question template is: Are the text descriptions of the pictures: \textless task1\_question \textgreater, \textless task1\_question \textgreater \dots all correct? Answer yes or no.
    \end{itemize}
\end{itemize}
The content in “\textless  \textgreater” in the above question template needs to be determined based on the specific prompt content.

%% file: sec2/8_results.tex
\section{More Results Comparisons}
\label{comparison}
As shown in Table \ref{geneval}, we further launch comparisons of the alignment between different metric results and human annotations in evaluating T2I model performance. We compare the performance of GenEval~\cite{ghosh2023geneval}, Grounding-DINO~\cite{liu2024grounding}, and our model across five tasks. Since GenEval~\cite{ghosh2023geneval} evaluates models using only these specific dimensions, we focus on tasks that align with GenEval’s capabilities to ensure a fair comparison. 
 GenEval~\cite{ghosh2023geneval} evaluates object detection using Mask2Former \cite{cheng2022masked}, which is part of the MMDetection \cite{chen2019mmdetection} toolbox from OpenMMLab, providing robust detection of objects and their relative positioning.  For the counting task, Mask2Former \cite{cheng2022masked} is paired with a higher confidence threshold (0.9) to improve human agreement.  Additionally, a heuristic method is used to evaluate the relative positioning of objects based on their bounding box coordinates, classifying objects as “left”, “right”, “above”, or “below” one another if they meet a minimum distance threshold.
For color classification, GenEval~\cite{ghosh2023geneval} utilizes the CLIP ViT-L/14 \cite{radford2021learning} model for zero-shot color classification, where each object’s bounding box is cropped to improve accuracy by removing the background.

To further explore the performance of detection models on these tasks, we replace Mask2Former \cite{cheng2022masked} with Grounding-DINO~\cite{liu2024grounding} and use the InternVL2.5-38B~\cite{chen2024expanding} model for color classification. While this improves counting and position tasks due to Grounding-DINO's enhanced detection, GenEval still outperforms on color, single-class, and two-class tasks.  This is likely due to differences in detection model threshold settings and highlights the limitations of using detection models as a backbone for tasks such as counting and position, which may require more specialized methods.
In contrast, our model, which combines LMM for comprehensive evaluation, outperforms both GenEval~\cite{ghosh2023geneval} and Grounding-DINO~\cite{liu2024grounding} in all tasks.
Unlike GenEval~\cite{ghosh2023geneval}, which relies on a combination of multiple models to handle different tasks, our approach is an \textit{\textbf{all-in-one}} solution that integrates various capabilities into a single framework. This unified design allows for more consistent and efficient performance across tasks, as it avoids the potential inconsistencies and complexities that arise from combining multiple specialized models. Our model demonstrates superior task-specific accuracy, achieving higher human agreement and better overall performance across all tasks, showcasing the advantage of our integrated approach over traditional detection-based methods and multi-model systems.

%% file: sec2/9_figures.tex
\input{tables2/MOS1}
\input{tables2/MOS2}
\input{tables2/MOS3}
\input{tables2/results}

\input{figures2/bz1}
\input{figures2/bz2}
\input{figures2/perception}
\input{figures2/perception2}
\input{figures2/correspondence1}
\input{figures2/correspondence2}
\include{figures2/task1}
\include{figures2/task2}
\include{figures2/task3}
\include{figures2/task4}
\include{figures2/task5}

%% file: tables2/MOS1.tex
\begin{table*}[tbph]
\vspace{-6mm}
\centering
\renewcommand\arraystretch{1}
\caption{Performance comparisons of T2I Models on human-annotated perception MOS.
}
\vspace{-3mm}
   \resizebox{\linewidth}{!}{\begin{tabular}{l||ccccccccccccccccccccc:c}
  \Xhline{1px}
 Models
&$\text{Single}$&$\text{Two Class}$&$\text{Counting}$&$\text{Colors}$&$\text{Position}$ &$\text{Shapes}$&$\text{Texture}$&$\text{Scene}$&$\text{Style}$&$\text{OCR}$&$\text{HOI}$&$\text{Human}$&$\text{Emotion}$&$\text{Linguistic}$&$\text{View}$&$\text{Knowledge}$&$\text{Face}$&$\text{Imagination}$&$\text{Time\&Light}$&$\text{Complex}$&Overall&Rank\\
    \hline
     Playground~\cite{li2024playground} &63.56&61.78&62.20&64.19&58.84&62.86&63.40&63.34&61.98&55.54&61.09&63.80&61.74&60.29&58.96&61.77&59.66&61.97&61.82&61.76&61.64&1\\ 
     
     Kolors~\cite{kolors}&63.47&61.51&61.58&63.96&59.59&61.92&61.93&61.40&62.53&53.01&59.75&62.50&62.18&59.73&59.34&61.64&59.27&60.67&60.96&61.47&61.14&2\\
     
     Infinity~\cite{Infinity}&65.31&60.68&61.67&65.02&58.02&60.60&62.20&63.76&61.10&64.78&59.73&61.82&60.54&59.26&57.81&58.34&56.53&61.17&59.36&61.65&60.86&3\\
    
    Flux\_schnell~\cite{flux2024} &65.17&62.99&61.82&63.05&59.66&62.46&63.28&65.45&57.71&65.83&63.31&62.78&60.27&59.50&58.92&58.96&48.76&59.31&52.26&63.05&60.63&4\\
    
    SD3\_5\_large~\cite{esser2024scaling} &64.37&62.48&61.71&63.93&58.39&60.70&61.62&57.26&59.57&65.27&56.56&60.71&57.18&57.64&57.29&57.47&49.87&60.48&52.95&62.40&59.50&5\\
    
     DALLE3~\cite{betker2023improving} &63.01&62.72&60.32&62.09&57.93&60.79&63.32&55.06&63.46&67.73&58.76&59.88&59.22&56.37&58.05&57.60&45.88&57.82&54.96&61.81&59.34&6\\
     
   Omnigen~\cite{xiao2024omnigen}&63.47&60.03&59.28&61.53&55.72&59.13&60.02&60.25&57.98&57.82&58.07&63.87&58.90&57.22&56.79&58.87&60.89&57.65&55.48&59.11&59.12&7\\
    
   Kandinsky-3~\cite{arkhipkin2024kandinsky}& 59.78&55.50&58.03&60.32&54.19&60.74&60.47&60.47&61.56&52.61&58.25&58.83&57.24&56.92&56.24&57.56&62.60&56.73&58.74&57.53&58.21&8\\

   PixArt-sigma~\cite{chen2024pixart}
&60.89&56.52&57.87&58.88&53.52&59.72&60.37&60.26&59.19&48.37&55.30&58.85&57.35&57.06&56.67&56.46&54.27&59.71&59.91&56.57&57.43&9 \\

   EMU3~\cite{wang2024emu3}& 57.08&53.58&53.53&54.56&50.78&54.74&55.73&57.55&57.23&43.02&53.72&57.73&54.56&54.37&52.81&54.83&56.08&54.19&55.81&52.72&54.29&10\\

   SDXL\_base\_1~\cite{podell2023sdxl}&59.34&56.33&57.49&59.84&52.48&56.53&57.72&51.87&54.79&49.90&50.76&53.56&49.53&50.07&53.05&54.38&41.55&52.46&50.96&54.81&53.50&11\\
   
 Show-o~\cite{xie2024show}&60.81&57.33&59.30&60.59&53.37&58.74&60.58&52.74&53.91&41.50&47.96&46.43&45.46&50.20&50.35&51.64&37.21&52.20&45.91&54.53&52.31&12\\
 
 Seed-xi~\cite{ge2024seed}& 55.06&46.23&52.50&54.60&45.34&55.15&55.74&52.62&53.99&49.19&46.27&47.45&50.63&49.20&52.09&54.50&42.58&53.15&52.10&49.97&50.73&13\\
 
    NOVA~\cite{deng2024nova}& 56.81&54.23&52.95&57.69&50.65&54.41&56.36&49.77&57.76&31.76&45.43&43.87&47.44&48.77&46.98&49.36&48.58&53.43&47.53&50.85&50.69&14\\
  
   LaVi-Bridge~\cite{zhao2024bridging}&56.13&52.18&52.74&54.03&45.96&54.44&54.04&52.53&56.12&39.37&51.42&46.85&50.60&50.38&48.25&48.11&45.04&51.62&50.38&49.95&50.56&15\\
   Hart~\cite{tang2024hart}&52.12&48.76&49.54&53.19&46.97&50.65&51.14&47.09&52.50&39.89&42.12&50.06&50.95&48.31&49.08&50.27&53.21&53.99&53.90&48.06&49.80&16\\
   
   LLMGA~\cite{xia2024llmga}&53.30&52.14&52.41&54.92&47.61&54.17&54.94&50.95&53.03&50.28&43.30&42.90&46.14&49.46&49.40&49.74&39.68&47.69&49.26&44.50&48.67&17\\

   SD\_v2-1~\cite{Rombach_2022_CVPR}&55.85&49.76&53.15&56.41&44.87&52.90&51.38&48.46&43.40&42.72&44.30&47.00&42.28&48.90&50.17&50.99&35.35&39.82&47.13&48.87&47.68&18\\

   ELLA~\cite{hu2024ella}&48.78&46.21&46.21&50.75&43.48&49.55&52.33&41.93&40.25&38.11&41.99&43.38&42.24&42.70&43.44&40.32&31.11&42.76&45.97&49.63&44.61&19\\

   Janus~\cite{wu2024janus}&42.57&40.18&37.82&41.99&36.58&41.00&41.06&38.91&40.53&26.47&33.63&30.69&33.34&34.77&36.89&37.97&29.81&34.26&40.89&37.24&36.98&20\\
   
    i-Code-V3~\cite{tang2023any}&42.58&36.07&37.27&37.96&30.44&40.41&39.92&35.61&38.71&33.44&32.81&30.63&29.93&36.84&32.27&32.23&34.45&29.49&35.50&33.48&34.70&21\\
    
   Vila-u~\cite{wu2024vila}&38.74&32.54&33.44&38.15&29.88&38.00&35.26&33.24&40.05&27.61&28.72&27.20&32.37&33.12&32.64&34.38&37.29&34.44&39.99&31.38&33.80&22\\

   LlamaGen~\cite{sun2024autoregressive}&33.86&30.90&33.12&33.96&27.89&34.17&35.17&32.53&33.29&27.81&29.46&29.05&26.52&33.19&29.34&32.96&21.72&28.77&27.59&27.04&29.96&23\\
   LWM~\cite{liu2024world}&35.11&29.55&32.08&32.68&25.82&36.12&33.10&30.87&30.95&34.21&29.23&24.33&24.15&30.64&26.00&26.17&29.18&22.50&29.44&26.89&28.88&24\\
    \Xhline{1px}
  \end{tabular}}\label{mos1}

 \vspace{-2mm}
\end{table*}

%% file: tables2/MOS2.tex
\begin{table*}[tbph]
\vspace{-2mm}
\centering
\renewcommand\arraystretch{1}
\caption{Performance comparisons of T2I Models on human-annotated correspondence MOS.
}
\vspace{-3mm}
   \resizebox{\linewidth}{!}{\begin{tabular}{l||ccccccccccccccccccccc:c}
  \Xhline{1px}
 Models
&$\text{Single}$&$\text{Two Class}$&$\text{Counting}$&$\text{Colors}$&$\text{Position}$ &$\text{Shapes}$&$\text{Texture}$&$\text{Scene}$&$\text{Style}$&$\text{OCR}$&$\text{HOI}$&$\text{Human}$&$\text{Emotion}$&$\text{Linguistic}$&$\text{View}$&$\text{Knowledge}$&$\text{Face}$&$\text{Imagination}$&$\text{Time\&Light}$&$\text{Complex}$&Overall&Rank\\
    \hline
    SD3\_5\_large~\cite{esser2024scaling} &64.96&62.10&60.32&63.79&45.58&55.56&60.48&64.10&60.29&65.66&60.20&62.59&58.47&38.73&56.59&61.72&55.66&59.39&58.17&57.75&58.35&1\\

    Flux\_schnell~\cite{flux2024} &64.88&63.19&58.71&58.32&49.32&53.16&57.91&66.11&59.31&65.29&62.68&62.64&59.72&35.54&58.40&62.69&56.94&61.40&56.68&56.03&58.10&2 \\

     DALLE3~\cite{betker2023improving} &64.57&63.50&55.28&63.60&48.29&54.85&58.75&62.87&57.25&63.84&61.07&62.64&61.16&40.73&60.85&62.67&53.79&62.01&59.34&53.22&57.97&3\\
    Infinity~\cite{Infinity}&65.42&57.82&56.31&63.99&45.19&48.36&54.79&65.92&61.22&60.64&59.73&62.46&59.23&35.71&57.32&62.65&58.95&59.66&61.75&55.56&57.43&4\\
    
     Playground~\cite{li2024playground} &63.59&59.00&55.04&62.99&39.47&54.98&58.50&64.98&60.74&38.42&60.42&62.27&60.67&42.52&57.39&63.72&59.77&58.87&62.38&46.53&56.06&5\\ 
     
Omnigen~\cite{xiao2024omnigen}&64.09&58.38&50.47&60.11&46.09&50.75&51.29&65.97&55.68&55.58&58.92&62.99&58.15&39.50&56.32&62.32&61.13&57.04&60.60&50.59&55.81&6\\
    
    PixArt-sigma~\cite{chen2024pixart}
&62.22&52.68&52.77&60.94&40.56&50.34&59.35&64.25&62.16&31.97&57.76&61.70&58.27&38.33&56.96&61.65&58.70&59.68&62.06&46.65&54.72&7 \\

 Show-o~\cite{xie2024show}&63.12&57.86&59.20&62.27&44.73&52.25&55.37&63.32&57.28&30.69&56.10&58.04&54.03&39.29&55.75&59.03&50.80&56.71&55.17&50.66&54.21&8\\

Kolors~\cite{kolors}&63.71&56.02&53.28&59.78&39.79&51.03&50.65&62.75&55.40&39.84&54.83&60.36&58.42&35.45&55.44&62.09&55.32&56.97&61.45&46.01&53.53&9\\
     
NOVA~\cite{deng2024nova}& 59.65&55.74&53.29&60.80&40.38&53.12&55.61&61.70&58.29&27.09&56.62&57.01&55.00&38.47&54.32&58.26&55.60&56.23&54.55&45.59&52.73&10\\

SDXL\_base\_1~\cite{podell2023sdxl}&61.38&52.38&48.85&60.34&39.16&50.52&54.82&62.24&59.17&44.22&57.69&58.78&53.92&41.05&54.75&60.28&49.12&53.54&56.90&42.91&52.23&11\\

  EMU3~\cite{wang2024emu3}& 58.40&48.65&44.81&57.77&38.21&47.31&50.93&63.35&56.42&32.13&55.50&59.78&55.39&38.94&54.04&59.47&55.87&54.61&59.02&40.79&50.97&12\\
    
  Seed-xi~\cite{ge2024seed}& 58.52&46.69&44.94&59.55&39.25&51.09&53.58&63.53&57.93&34.77&55.63&53.93&55.31&39.40&55.25&59.51&48.98&56.33&57.77&41.32&50.96&13\\
    
  Hart~\cite{tang2024hart}&55.22&45.66&45.89&57.49&38.51&47.00&51.51&59.73&55.46&27.44&51.14&57.91&55.19&38.94&53.85&56.30&57.59&54.85&59.83&41.71&50.30&14\\
   
    LaVi-Bridge~\cite{zhao2024bridging}&60.12&50.33&49.29&59.82&34.70&48.43&52.16&63.31&57.69&27.12&56.24&56.95&54.66&37.88&52.94&54.45&50.32&53.63&58.55&39.07&50.19&15\\
     ELLA~\cite{hu2024ella}&56.15&46.30&47.71&58.26&38.95&48.10&52.11&60.67&49.89&31.52&51.98&56.18&51.15&36.08&51.61&53.65&41.37&50.46&53.69&46.17&49.07&16\\ 
     
   Kandinsky-3~\cite{arkhipkin2024kandinsky}& 58.71&42.14&46.52&53.22&34.37&50.46&45.89&58.91&50.42&30.55&52.94&55.30&54.10&36.87&52.78&58.61&56.54&54.77&57.56&35.43&48.37&17\\

   SD\_v2-1~\cite{Rombach_2022_CVPR}&59.86&46.84&46.62&58.48&34.31&49.30&50.78&60.46&52.24&34.64&53.37&54.08&47.89&44.59&54.01&56.84&42.22&42.88&50.81&38.50&47.96&18\\
 
Janus~\cite{wu2024janus}&50.90&44.38&39.35&56.09&46.98&42.13&43.46&58.68&51.88&26.38&44.85&49.30&45.30&40.04&50.66&50.96&43.03&42.27&50.38&42.06&45.94&19\\
    
   Vila-u~\cite{wu2024vila}&47.69&37.88&39.35&51.55&33.31&43.94&41.47&51.54&50.91&26.34&43.77&48.24&46.44&40.08&48.16&48.69&49.04&43.81&53.18&34.68&43.47&20\\   
   LLMGA~\cite{xia2024llmga}&53.92&40.69&36.84&49.10&32.85&41.44&41.88&60.43&50.41&38.52&42.97&41.36&46.74&43.92&47.98&51.45&42.37&46.45&51.73&32.54&43.43&21\\
     
    i-Code-V3~\cite{tang2023any}&49.24&34.35&39.11&48.28&28.87&41.66&42.94&56.29&47.69&29.26&43.76&44.32&38.01&40.94&42.70&42.83&39.89&33.45&43.31&30.85&39.80&22\\
   LlamaGen~\cite{sun2024autoregressive}&44.06&35.45&37.81&46.75&31.34&38.34&40.49&49.64&44.38&27.37&43.08&43.30&36.59&38.08&43.23&44.96&27.54&34.75&35.42&29.77&37.73&23\\
   LWM~\cite{liu2024world}&43.55&32.29&34.56&43.80&28.10&40.23&36.69&46.71&42.15&35.79&38.64&35.38&32.62&36.44&37.06&35.08&36.62&28.22&35.75&29.35&35.46&24\\
    \Xhline{1px}
  \end{tabular}}\label{mos2}

 \vspace{-2mm}
\end{table*}

%% file: tables2/MOS3.tex
\begin{table*}[tbph]
\vspace{-2mm}
\centering
\renewcommand\arraystretch{1}
\caption{Performance comparisons of T2I Models on human-annotated task-specific accuracy.
}
\vspace{-3mm}
   \resizebox{\linewidth}{!}{\begin{tabular}{l||ccccccccccccccccccccc:c}
  \Xhline{1px}
 Models
&$\text{Single}$&$\text{Two Class}$&$\text{Counting}$&$\text{Colors}$&$\text{Position}$ &$\text{Shapes}$&$\text{Texture}$&$\text{Scene}$&$\text{Style}$&$\text{OCR}$&$\text{HOI}$&$\text{Human}$&$\text{Emotion}$&$\text{Linguistic}$&$\text{View}$&$\text{Knowledge}$&$\text{Face}$&$\text{Imagination}$&$\text{Time\&Light}$&$\text{Complex}$&Overall&Rank\\
    \hline
    SD3\_5\_large~\cite{esser2024scaling} &98.89&94.50&82.22&91.35&36.94&68.75&86.00&96.97&88.66&94.00&91.86&98.10&90.09&23.81&75.25&97.00&76.79&82.88&87.13&78.42&81.43&1
\\

    Flux\_schnell~\cite{flux2024} &95.56&94.50&74.44&76.92&48.65&60.00&77.00&98.48&89.69&94.00&95.35&95.24&90.99&11.90&84.16&100.00&83.04&96.40&83.17&71.92&80.29&2
 \\

     DALLE3~\cite{betker2023improving} &100.00&94.50&66.67&93.27&47.75&70.00&83.00&95.45&73.20&90.00&89.53&97.14&95.50&29.76&92.08&97.00&70.54&94.59&90.10&64.73&80.24&3
\\
    Infinity~\cite{Infinity}&100.00&78.90&67.78&90.38&37.84&47.50&67.00&100.00&92.78&80.00&89.53&97.14&89.19&13.10&83.17&99.00&83.93&88.29&93.07&71.23&78.10&4
\\
    
     Playground~\cite{li2024playground} &94.44&84.40&65.56&94.23&22.52&68.75&78.00&98.48&88.66&12.00&89.53&97.14&95.50&33.33&82.18&100.00&86.61&85.59&93.07&41.10&73.86&5
\\ 
     
Omnigen~\cite{xiao2024omnigen}&96.67&83.49&50.00&79.81&39.64&56.25&55.00&100.00&75.26&68.00&87.21&97.14&84.68&23.81&78.22&97.00&91.96&78.38&93.07&56.51&73.29&6
\\

     Show-o~\cite{xie2024show}&96.67&82.57&78.89&89.42&36.94&58.75&70.00&95.45&80.41&0.00&82.56&93.33&79.28&25.00&80.20&93.00&63.39&83.78&80.20&57.88&71.71&7
\\
     
    PixArt-sigma~\cite{chen2024pixart}
&88.89&62.39&60.00&86.54&23.42&50.00&81.00&93.94&96.91&2.00&82.56&99.05&85.59&21.43&84.16&97.00&87.50&88.29&96.04&43.15&70.71&8
 \\

NOVA~\cite{deng2024nova}& 91.11&76.15&64.44&89.42&25.23&67.50&72.00&93.94&84.54&0.00&82.56&91.43&79.28&20.24&78.22&93.00&80.36&82.88&69.31&41.78&68.19&9
\\

Kolors~\cite{kolors}&95.56&71.56&60.00&82.69&23.42&53.75&54.00&90.91&74.23&14.00&68.60&89.52&83.78&13.10&69.31&95.00&71.43&77.48&92.08&40.75&65.05&10
\\
     
SDXL\_base\_1~\cite{podell2023sdxl}&92.22&64.22&43.33&87.50&20.72&56.25&69.00&93.94&89.69&30.00&83.72&92.38&72.97&28.57&77.23&97.00&46.43&63.96&82.18&33.56&63.67&11
\\

  Seed-xi~\cite{ge2024seed}& 87.78&42.20&34.44&87.50&16.22&61.25&65.00&100.00&87.63&4.00&87.21&78.10&83.78&25.00&79.21&93.00&49.11&82.88&87.13&27.05&61.43&12
\\
  
  EMU3~\cite{wang2024emu3}& 83.33&47.71&34.44&80.77&16.22&41.25&53.00&96.97&79.38&2.00&75.58&91.43&77.48&25.00&71.29&96.00&77.68&73.87&89.11&25.68&59.90&13
\\
     Hart~\cite{tang2024hart}&72.22&38.53&38.89&75.00&19.82&38.75&56.00&87.88&79.38&0.00&61.63&84.76&84.68&25.00&78.22&86.00&83.04&71.17&94.06&31.51&59.29&14
\\
   
    LaVi-Bridge~\cite{zhao2024bridging}&87.78&59.63&50.00&87.50&9.91&51.25&56.00&96.97&79.38&0.00&79.07&85.71&78.38&20.24&74.26&75.00&57.14&68.47&91.09&23.29&59.10&15
\\
     ELLA~\cite{hu2024ella}&75.56&42.20&46.67&83.65&19.82&52.50&61.00&96.97&58.76&4.00&61.63&84.76&66.67&16.67&64.36&73.00&33.04&51.35&68.32&44.86&54.90&16\\ 
     
   Kandinsky-3~\cite{arkhipkin2024kandinsky}& 81.11&25.69&41.11&61.54&9.01&52.50&38.00&81.82&48.45&0.00&60.47&74.29&70.27&16.67&69.31&91.00&73.21&70.27&79.21&12.67&50.14&17\\

   SD\_v2-1~\cite{Rombach_2022_CVPR}&88.89&46.79&41.11&82.69&8.11&53.75&52.00&89.39&62.89&4.00&69.77&74.29&39.64&42.86&75.25&87.00&25.00&19.82&51.49&21.58&48.86&18\\
 
Janus~\cite{wu2024janus}&61.11&29.36&18.89&77.88&51.35&26.25&23.00&87.88&65.98&0.00&38.37&54.29&31.53&26.19&67.33&63.00&34.82&17.12&59.41&33.56&42.95&19\\

    LLMGA~\cite{xia2024llmga}&70.00&28.44&13.33&51.92&4.50&30.00&25.00&87.88&59.79&26.00&34.88&37.14&50.45&39.29&54.46&68.00&25.89&40.54&62.38&10.27&37.67&20\\
    
   Vila-u~\cite{wu2024vila}&48.89&11.01&27.78&62.50&9.01&35.00&19.00&56.06&63.92&0.00&32.56&50.48&36.04&30.95&53.47&49.00&47.32&19.82&71.29&14.04&35.24&21\\    
     
    i-Code-V3~\cite{tang2023any}&60.00&8.26&22.22&54.81&0.00&28.75&34.00&78.79&47.42&2.00&26.74&36.19&11.71&34.52&32.67&27.00&19.64&3.60&23.76&5.48&25.00&22\\
   LlamaGen~\cite{sun2024autoregressive}&36.67&8.26&18.89&51.92&8.11&16.25&27.00&46.97&32.99&0.00&36.05&32.38&11.71&25.00&42.57&39.00&0.00&5.41&12.87&6.85&21.19&23\\
   LWM~\cite{liu2024world}&36.67&1.83&11.11&38.46&1.80&25.00&14.00&53.03&24.74&18.00&12.79&17.14&3.60&17.86&25.74&18.00&12.50&0.00&14.85&5.14&15.48&24\\
    \Xhline{1px}
  \end{tabular}}\label{mos3}

 \vspace{-2mm}
\end{table*}

%% file: tables2/results.tex
\begin{table*}[tbph]
\vspace{-2mm}
\centering
\renewcommand\arraystretch{0.9}
\caption{Comparisons of the alignment between different metric results and human annotations in evaluating T2I model performance.
}
\vspace{-3mm}
   \resizebox{\linewidth}{!}{\begin{tabular}{l||c:>{\columncolor{mycolor_gray}}ccc|c:>{\columncolor{mycolor_gray}}ccc|c:>{\columncolor{mycolor_gray}}ccc|c:>{\columncolor{mycolor_gray}}ccc|c:>{\columncolor{mycolor_gray}}ccc}
  \Xhline{1px}
    Dimension  &\multicolumn{4}{c|}{\textbf{Single Class}}&\multicolumn{4}{c|}{\textbf{Two Class}}&\multicolumn{4}{c|}{\textbf{Counting}}&\multicolumn{4}{c|}{\textbf{Position}}&\multicolumn{4}{c}{\textbf{Color}}\\
  \cmidrule(lr){2-5}  \cmidrule(lr){6-9} \cmidrule(lr){10-13} \cmidrule(lr){14-17}\cmidrule(lr){18-21}
 Models
&$\text{Human}$&$\text{Ours}$&$\text{GenEval}$&$\text{G-dino}$&$\text{Human}$ &$\text{Ours}$&$\text{GenEval}$&$\text{G-dino}$&$\text{Human}$&$\text{Ours}$&$\text{GenEval}$&$\text{G-dino}$&$\text{Human}$&$\text{Ours}$&$\text{GenEval}$&$\text{G-dino}$&$\text{Human}$&$\text{Ours}$&$\text{GenEval}$&$\text{Qwen2.5}$\\
    \midrule
    SD3\_5\_large~\cite{esser2024scaling} & 98.89 & 93.94&  100.0 & 100.0 &94.50  & 93.94 & 91.92 & 95.96 & 82.22 &76.25  & 75.00  &  70.00  & 36.94& 35.00&22.00 & 49.00  & 91.35  &90.43&84.04 & 84.29\\
    Flux\_schnell~\cite{flux2024} & 95.56   & 96.25   & 100.0 &100.0&94.50& 95.96 & 89.90 & 97.98 &74.44 &73.75& 72.50 & 62.50 & 48.65 &49.00 &28.00& 62.50 & 76.92 &78.72&77.66 &88.00\\
     DALLE3~\cite{betker2023improving} & 100.0  & 100.0 & 98.75 & 100.0 & 94.50&  93.94 & 84.85 &91.92  & 66.67 & 58.75 & 45.00  & 51.25  & 47.75 & 50.00& 45.00& 54.00 & 93.27 & 92.55&77.66 & 86.67\\
     Infinity~\cite{Infinity}& 100.0 &100.0 & 100.0 &100.0 & 78.90& 78.79 &  71.72 & 82.83 & 67.78 & 66.25 & 60.00 & 56.25 & 37.84 & 38.00 & 29.00& 51.00 & 90.38 & 89.36 & 82.98 & 89.57\\
    Playground~\cite{li2024playground} & 94.44 & 94.68 & 98.75 &100.0 & 84.40& 85.86 & 71.72 & 91.92  & 65.56 & 63.75 & 50.00 &  65.00 & 22.52 & 25.00 & 8.00& 36.00 & 94.23 &94.68&82.98&94.21 \\ 
   Omnigen~\cite{xiao2024omnigen}
    &96.67&98.75 &100.0&100.0 &83.49&86.87&82.83& 91.92&50.00 &46.25&46.25&52.00&39.64&40.00&27.00&52.00&79.81&79.79&75.53&76.84\\
     Show-o~\cite{xie2024show}&96.67 &97.50 &98.75&100.0&82.57   &82.83 &79.80 &92.93&78.89&82.50&71.25 &67.50 &36.94 &39.00&29.00&58.00&89.42 &90.43&76.60&88.42\\
     PixArt-sigma~\cite{chen2024pixart}
&88.89&91.25&98.75&100.0&62.39&64.65&68.69&86.87&60.00&56.25&50.00 &53.75 &23.42 &28.00&11.00&43.00&86.54&88.30&80.85 &80.00 \\
    NOVA~\cite{deng2024nova}& 91.11 & 91.25 &98.75   & 100.0&76.15 & 79.80 & 80.81 &92.93&64.44 & 58.75 &25.23 &56.25 &25.23 
  & 27.00&13.00&51.00&89.42&88.30&84.04&87.50\\
  Kolors~\cite{kolors}& 95.56 & 95.00 & 97.50 &100.0 & 71.56 & 70.71  & 69.70& 81.82 & 60.00 & 58.75& 45.00 & 53.75  & 23.42&28.00 &14.00 & 36.00 & 82.69 & 82.98 &78.72 &83.48\\
   
    SDXL\_base\_1~\cite{podell2023sdxl}&92.22 &92.50  &98.75 &100.0&64.22&63.64 &63.64&81.82 &43.33&43.75 &43.75 &42.50 &20.72&24.00&12.00&38.00&87.50&87.23&86.17&88.82\\
   Seed-xi~\cite{ge2024seed}& 87.78 & 90.00 & 97.50  &100.0&42.20 & 42.43 &63.64 &89.90&34.44 &33.75&35.00 &30.00 &16.22 &20.00&17.00&38.00&87.50 & 87.23& 89.36 &79.09\\
   EMU3~\cite{wang2024emu3}& 83.33   & 87.50  & 95.00 &100.0 & 47.71&  50.51 &61.62 &84.85&34.44 &32.50 &31.25 &32.50 &16.22 & 18.00&11.00&47.00&80.77&84.04&77.66&83.16\\
   Hart~\cite{tang2024hart}&72.22&72.50&96.25&100.0&38.53& 42.43 &53.54 &85.86 &38.89 &38.75 &33.75&38.75 &19.82 &25.00&12.00&43.00&75.00&78.72&80.85&62.86\\
   
   LaVi-Bridge~\cite{zhao2024bridging}& 87.78 & 88.75  & 97.50 & 100.0& 59.63& 60.61 &61.62 &78.79 &50.00 &47.50 &41.25 & 46.25&9.91&13.00&4.00&30.00&87.50&85.11&84.04&87.06\\
   ELLA~\cite{hu2024ella}&75.56&78.75&90.00&100.0&42.20&45.45 &32.32 &74.74&46.67 &42.50&12.50 &47.50 &19.82 &24.00&6.00&37.00&83.65&86.17&63.83&86.67\\
   
   Kandinsky-3~\cite{arkhipkin2024kandinsky}& 81.11  & 80.00 & 96.25 &100.0&25.69 & 31.31 &36.36 &54.55&41.11 &43.75&37.50 &47.50 &9.01 &11.00&9.00&31.00&61.54&63.83&63.83&61.38\\
   
   SD\_v2-1~\cite{Rombach_2022_CVPR}&88.89&88.75&95.00&100.0&46.79&51.52 &49.49 &78.79&41.11 &41.25&38.75 &45.00 &8.11 &13.00&7.00&33.00&82.69&84.04&81.91&80.00\\
   
   Janus~\cite{wu2024janus}&61.11&68.75&92.50&100.0&29.36&32.32 &67.68 &77.78&18.89 &15.00&21.25 &30.00 &51.35 &57.00&51.00&69.00&77.88&81.91&86.17&76.25\\
   LLMGA~\cite{xia2024llmga}&70.00&75.00&90.00&97.50&28.44&32.32 &32.32 &52.53&13.33 &13.75&12.50 &8.75 &4.50 &7.00&6.00&25.00&51.92&52.13&63.83&61.11\\
   
   Vila-u~\cite{wu2024vila}&48.89&47.50&85.00&98.75&11.01&12.12 &41.41 &69.70&27.78 &27.50&27.50 &28.75 &9.01 &10.00&3.00&33.00&62.50&63.83&74.47&68.18\\
   
   i-Code-V3~\cite{tang2023any}&60.00&62.50&86.25&100.0&8.26 &6.07 &15.15 &40.00&22.22&21.25&23.75 &27.50&0.00&0.00 &2.00&14.00&54.81&55.32&62.77&53.85\\
   LlamaGen~\cite{sun2024autoregressive}&36.67&38.75&75.00&100.0&8.26&9.10 &26.26 &70.71&18.89&23.75&25.00 &27.50 &8.11 &8.00&9.00&32.00&51.92&54.26&53.19&50.42\\
   LWM~\cite{liu2024world}&36.67&45.00&18.75&100.0&1.83&1.01 &1.01 &53.54&11.11 &8.75&0.00 &18.75 &1.80 &2.00&0.00&18.00&38.46&42.55&0.00&35.56\\
   \hline
   SRCC   to  human $\uparrow$& - & \textbf{\red{0.982}}&  \textbf{\blue{0.921}}& 0.346 &-& \textbf{\red{0.995}}& \textbf{\blue{0.936}} & 0.879 & -&\textbf{\red{0.990}}&0.865&\textbf{\blue{0.966}}&-&\textbf{\red{0.987}}&0.887 & \textbf{\blue{0.926}}&-&\textbf{\red{0.984}}&0.670&\textbf{\blue{0.848}}\\
   RMSE to human $\downarrow$& -  & \textbf{\red{3.14}} & \textbf{\blue{16.84}}& 26.74 &-& \textbf{\red{2.55}} & \textbf{\blue{13.37}} & 32.08 &-&\textbf{\red{3.30}}&13.27&\textbf{\blue{7.47}}&-&\textbf{\red{2.98}}&\textbf{\blue{9.13}}&19.29&-&\textbf{\red{2.00}}&11.36&\textbf{\blue{5.19}}\\
    \Xhline{1px}
  \end{tabular}}\label{geneval}

 \vspace{-2mm}
\end{table*}

%% file: figures2/bz1.tex
\begin{figure*}[!t]
\vspace{-5mm}
	\centering
	\includegraphics[width=0.89\linewidth]{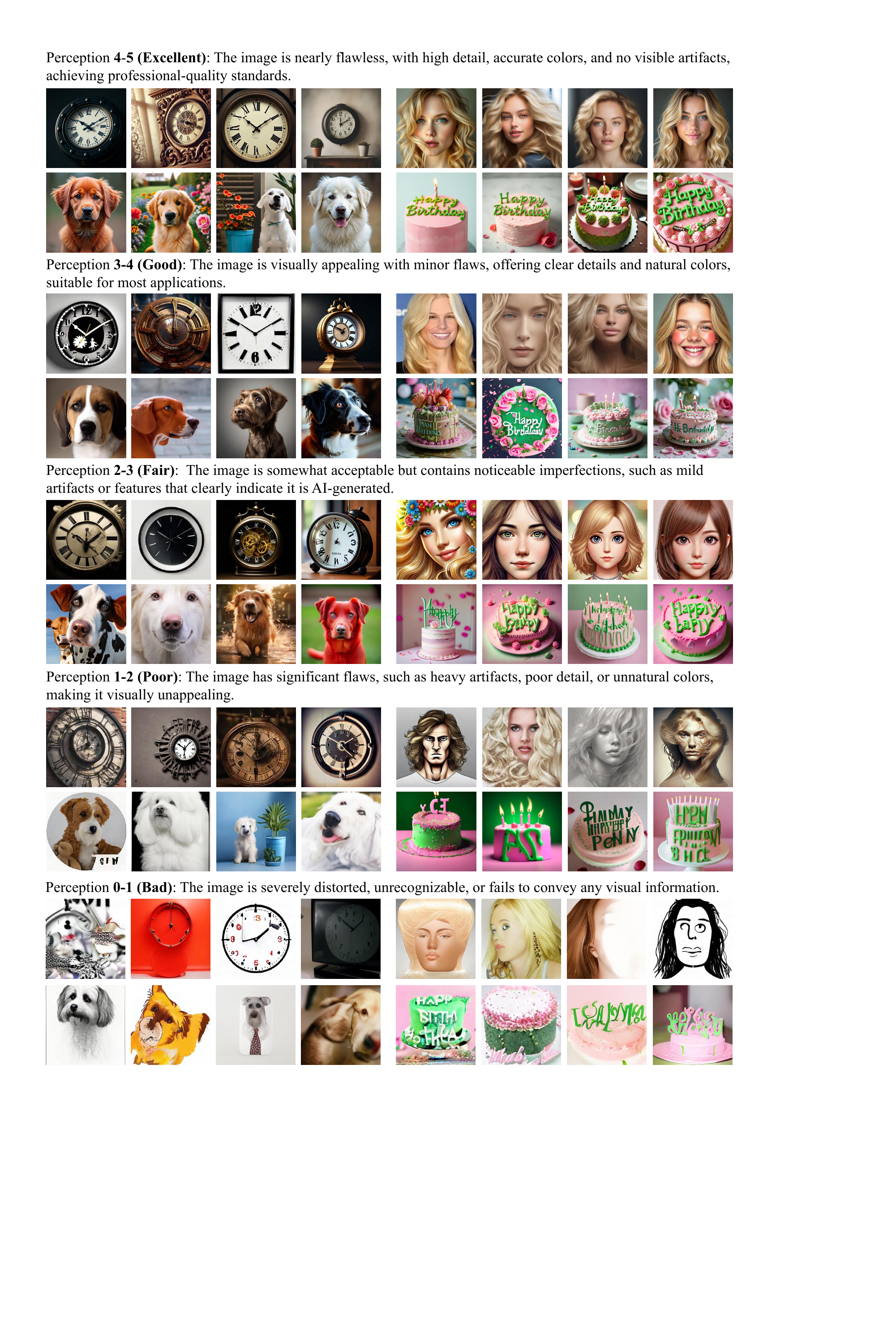}
 \vspace{-2mm}
	\caption{Instructions and examples for manual evaluation of \textbf{perception}.}
 \vspace{-3mm}
	\label{sup_bz1}
\end{figure*}

%% file: figures2/bz2.tex
\begin{figure*}[!t]
\vspace{-5mm}
	\centering
	\includegraphics[width=0.9\linewidth]{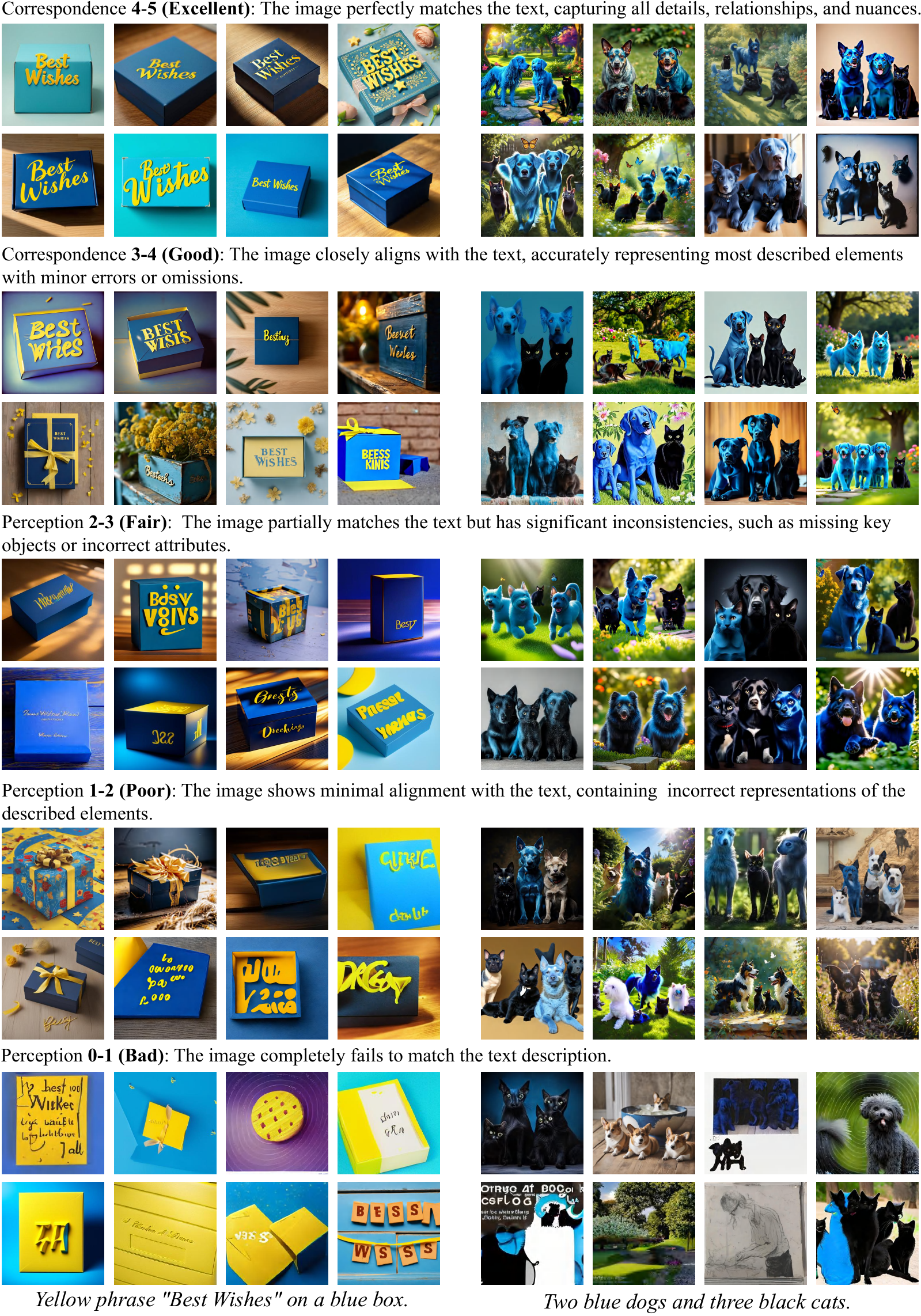}
 \vspace{-1mm}
	\caption{Instructions and examples for manual evaluation of \textbf{T2I correspondence}. Prompt (left): yellow phrase ``Best Wishes" on a blue box. Prompt (right): two blue dogs and three black cats.}
 \vspace{-3mm}
	\label{sup_bz2}
\end{figure*}

%% file: figures2/perception.tex
\begin{figure*}[!t]
\vspace{-5mm}
	\centering
	\includegraphics[width=\linewidth]{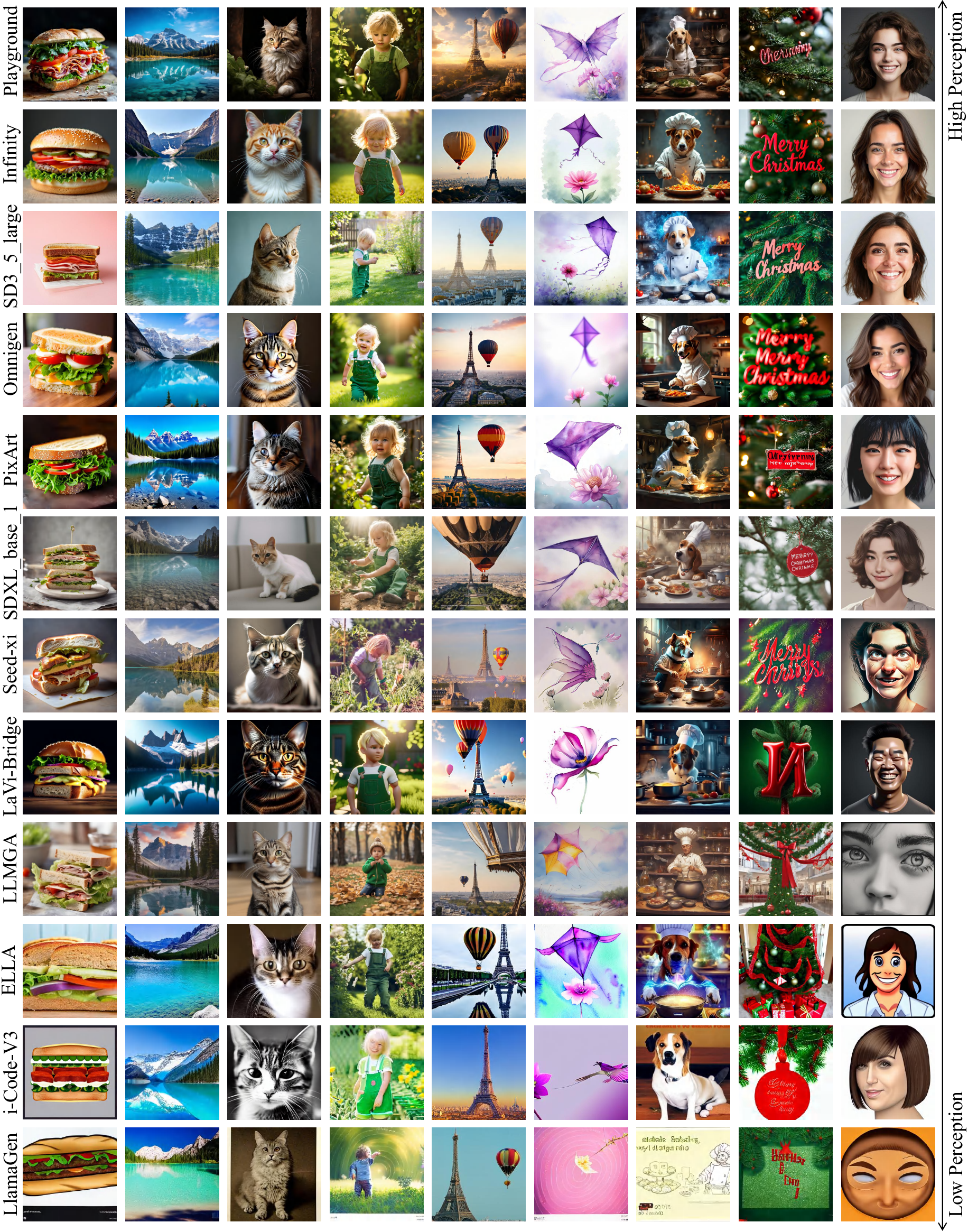}
 \vspace{-4mm}
	\caption{Visualization of generated images in the EvalMi-50K: sort by average \textbf{perception} quality of T2I models from high to low.}
 \vspace{-3mm}
	\label{perception}
\end{figure*}

%% file: figures2/perception2.tex
\begin{figure*}[!t]
\vspace{-5mm}
	\centering
	\includegraphics[width=\linewidth]{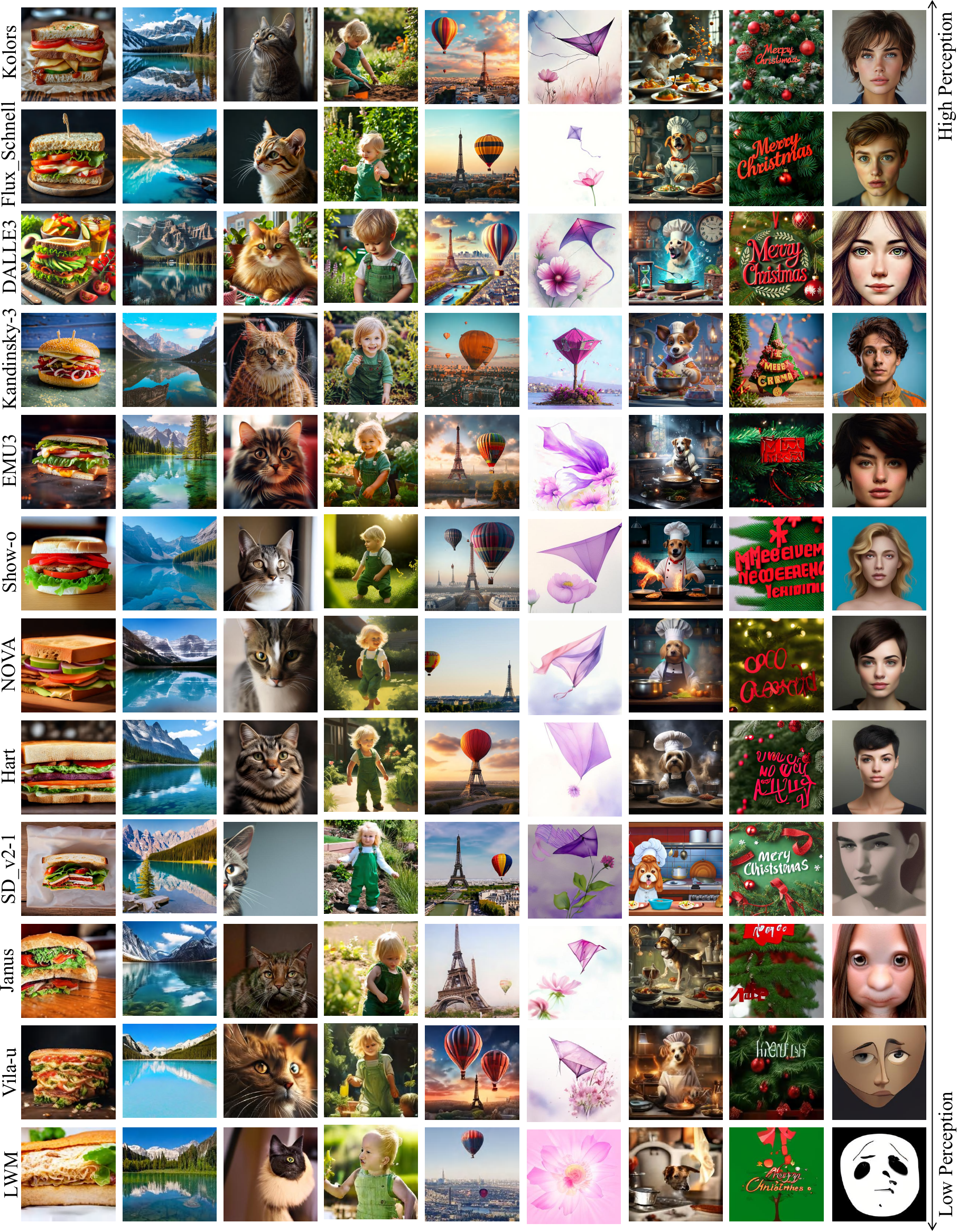}
 \vspace{-4mm}
	\caption{Visualization of generated images in the EvalMi-50K: sort by average \textbf{perception} quality of T2I models from high to low.}
 \vspace{-3mm}
	\label{perception2}
\end{figure*}

%% file: figures2/correspondence1.tex
\begin{figure*}[!t]
\vspace{-5mm}
	\centering
	\includegraphics[width=0.95\linewidth]{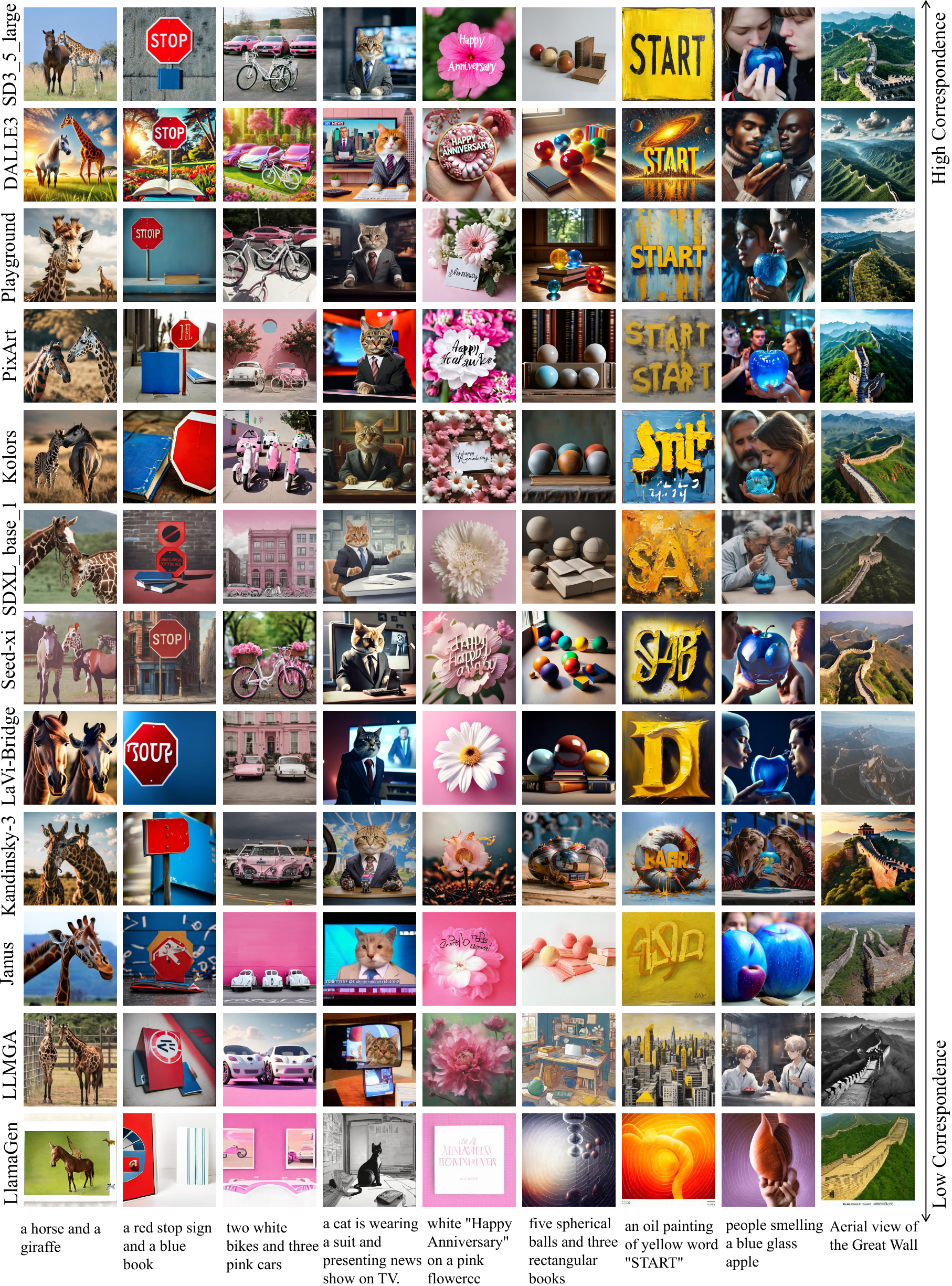}
 \vspace{-2mm}
	\caption{Visualization of generated videos in the EvalMi-50K: sort by average \textbf{T2I correspondence} of T2I models from high to low.}
 \vspace{-3mm}
	\label{correspondence}
\end{figure*}

%% file: figures2/correspondence2.tex
\begin{figure*}[!t]
\vspace{-5mm}
	\centering
	\includegraphics[width=0.95\linewidth]{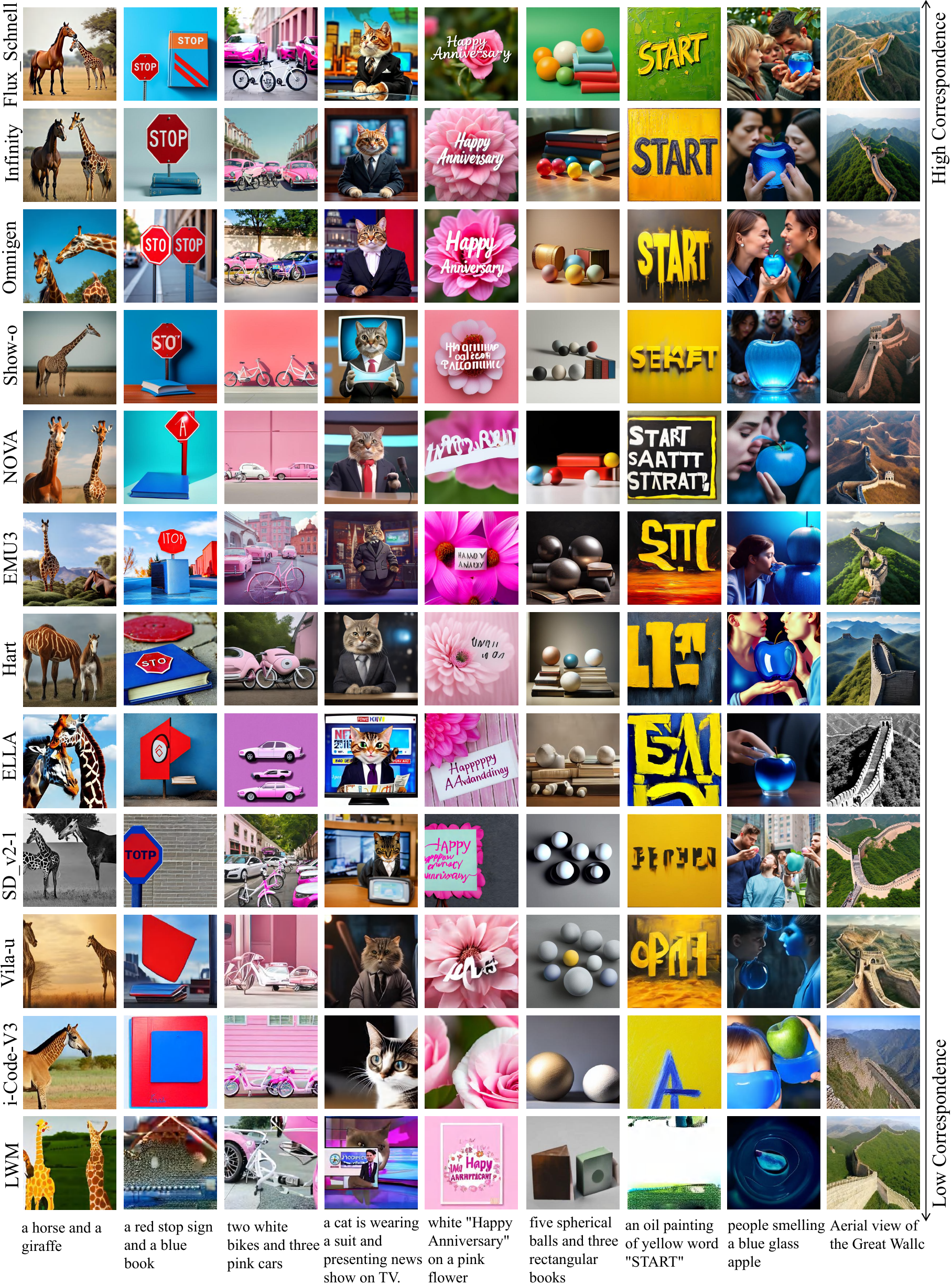}
 \vspace{-2mm}
	\caption{Visualization of generated videos in the EvalMi-50K: sort by average \textbf{T2I correspondence} of T2I models from high to low.}
 \vspace{-3mm}
	\label{correspondence2}
\end{figure*}

%% file: figures2/task1.tex
\begin{figure*}[!t]
\vspace{-5mm}
	\centering
	\includegraphics[width=0.89\linewidth]{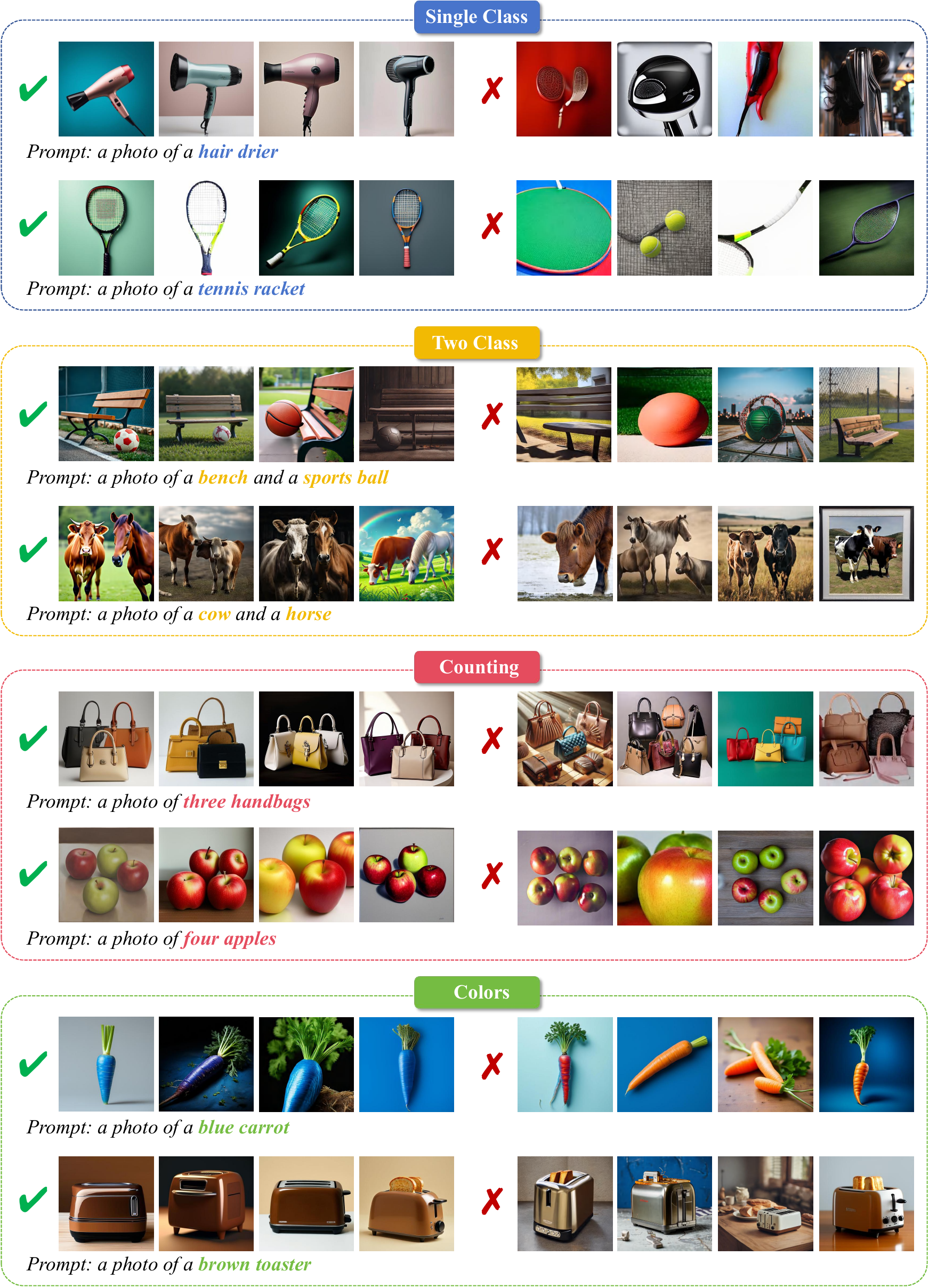}
 \vspace{-2mm}
	\caption{Examples for different task-specific challenges.}
 \vspace{-3mm}
	\label{task1}
\end{figure*}

%% file: figures2/task2.tex
\begin{figure*}[!t]
\vspace{-5mm}
	\centering
	\includegraphics[width=0.89\linewidth]{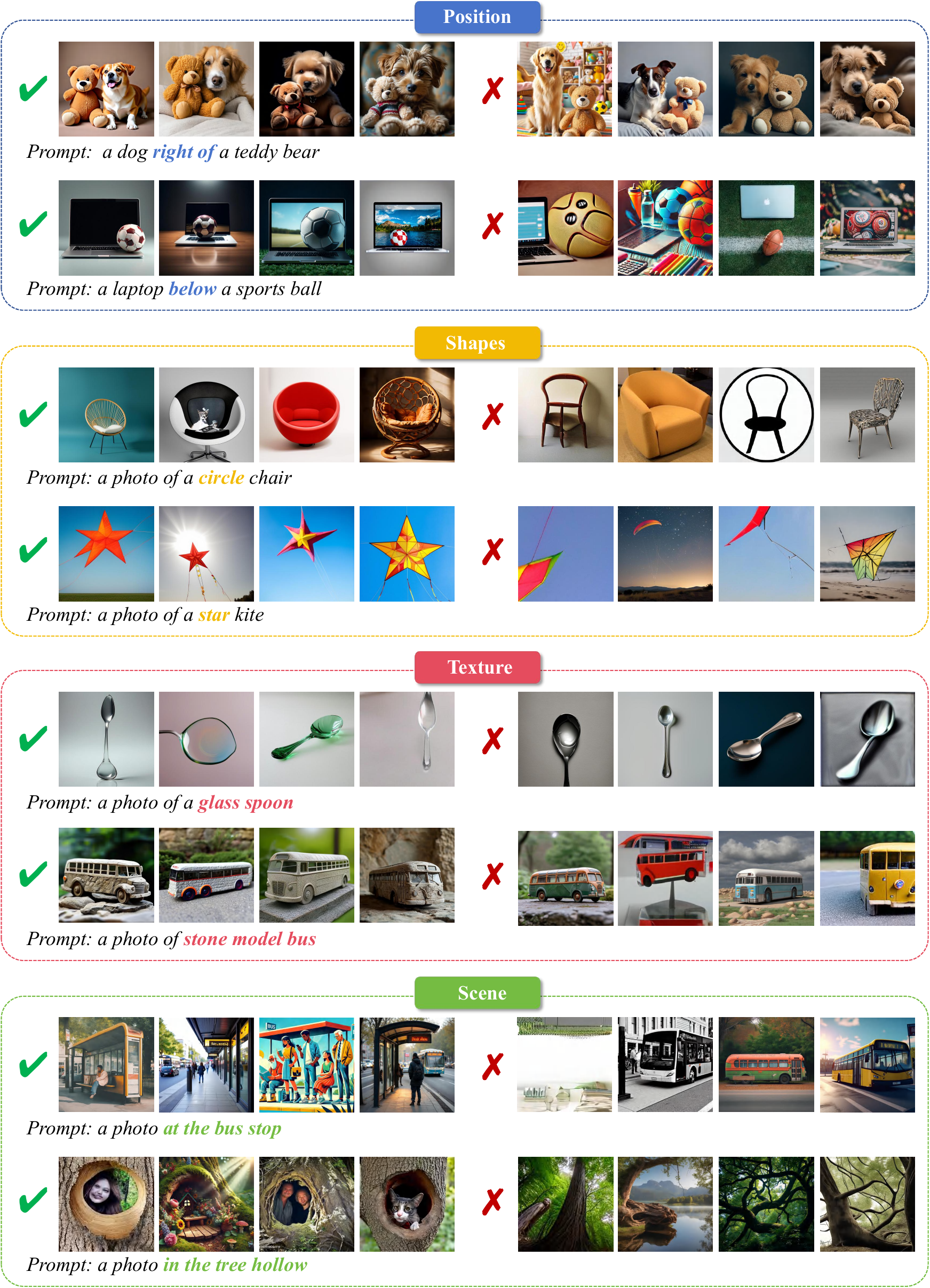}
 \vspace{-2mm}
	\caption{Examples for different task-specific challenges.}
 \vspace{-3mm}
	\label{task2}
\end{figure*}

%% file: figures2/task3.tex
\begin{figure*}[!t]
\vspace{-5mm}
	\centering
	\includegraphics[width=0.89\linewidth]{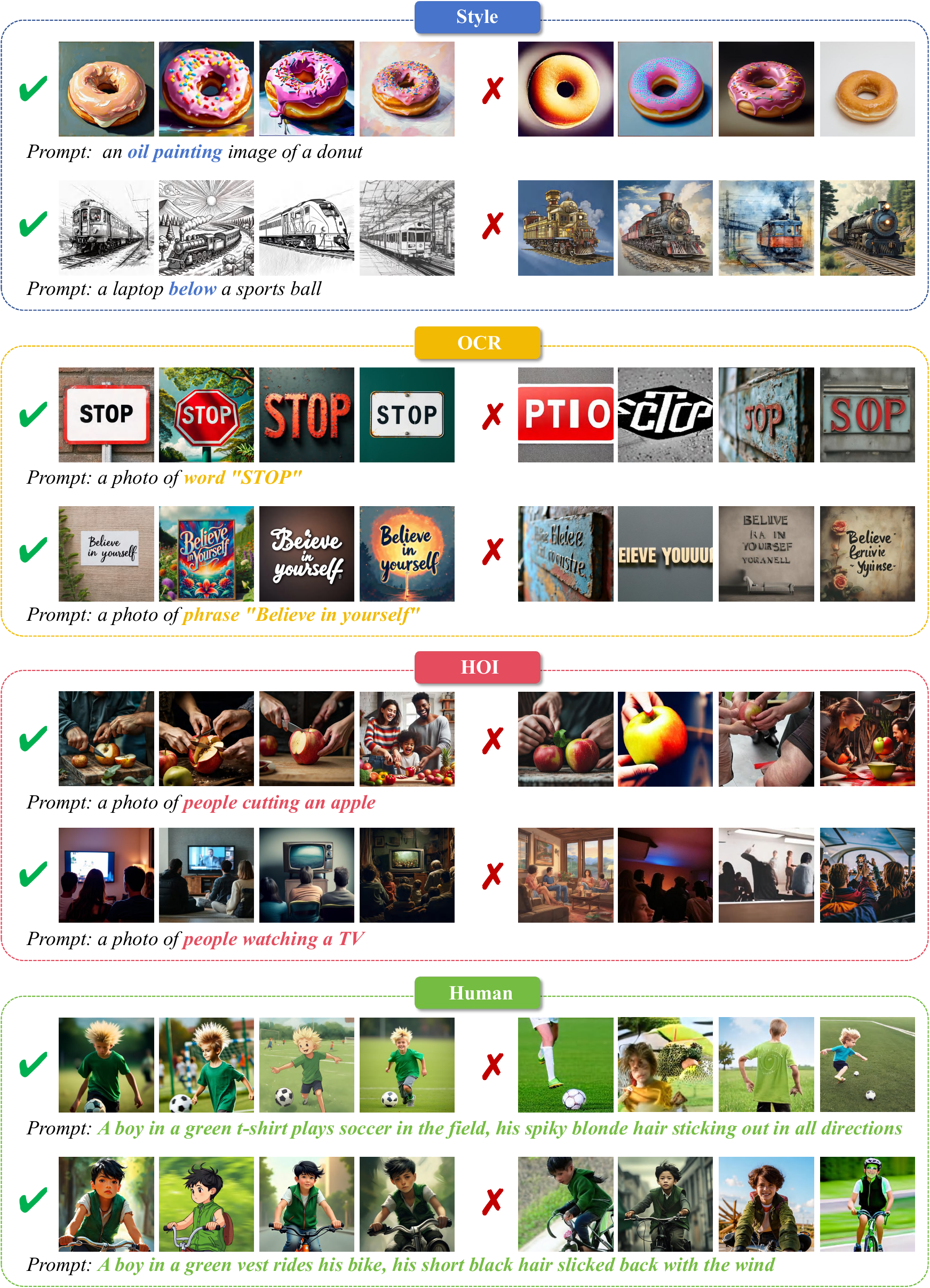}
 \vspace{-2mm}
	\caption{Examples for different task-specific challenges.}
 \vspace{-3mm}
	\label{task3}
\end{figure*}

%% file: figures2/task4.tex
\begin{figure*}[!t]
\vspace{-5mm}
	\centering
	\includegraphics[width=0.89\linewidth]{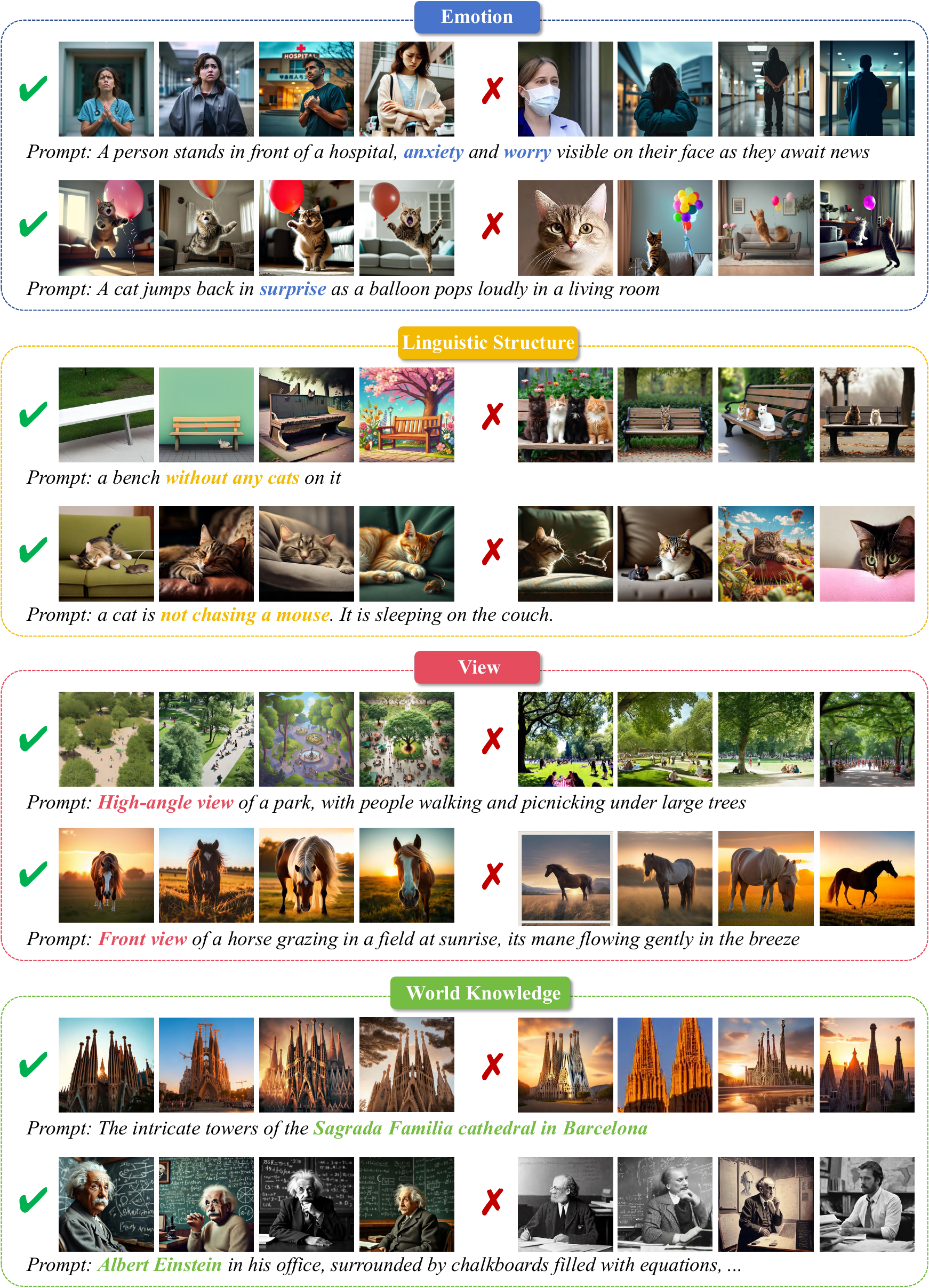}
 \vspace{-2mm}
	\caption{Examples for different task-specific challenges.}
 \vspace{-3mm}
	\label{task4}
\end{figure*}

%% file: figures2/task5.tex
\begin{figure*}[!t]
\vspace{-5mm}
	\centering
	\includegraphics[width=0.89\linewidth]{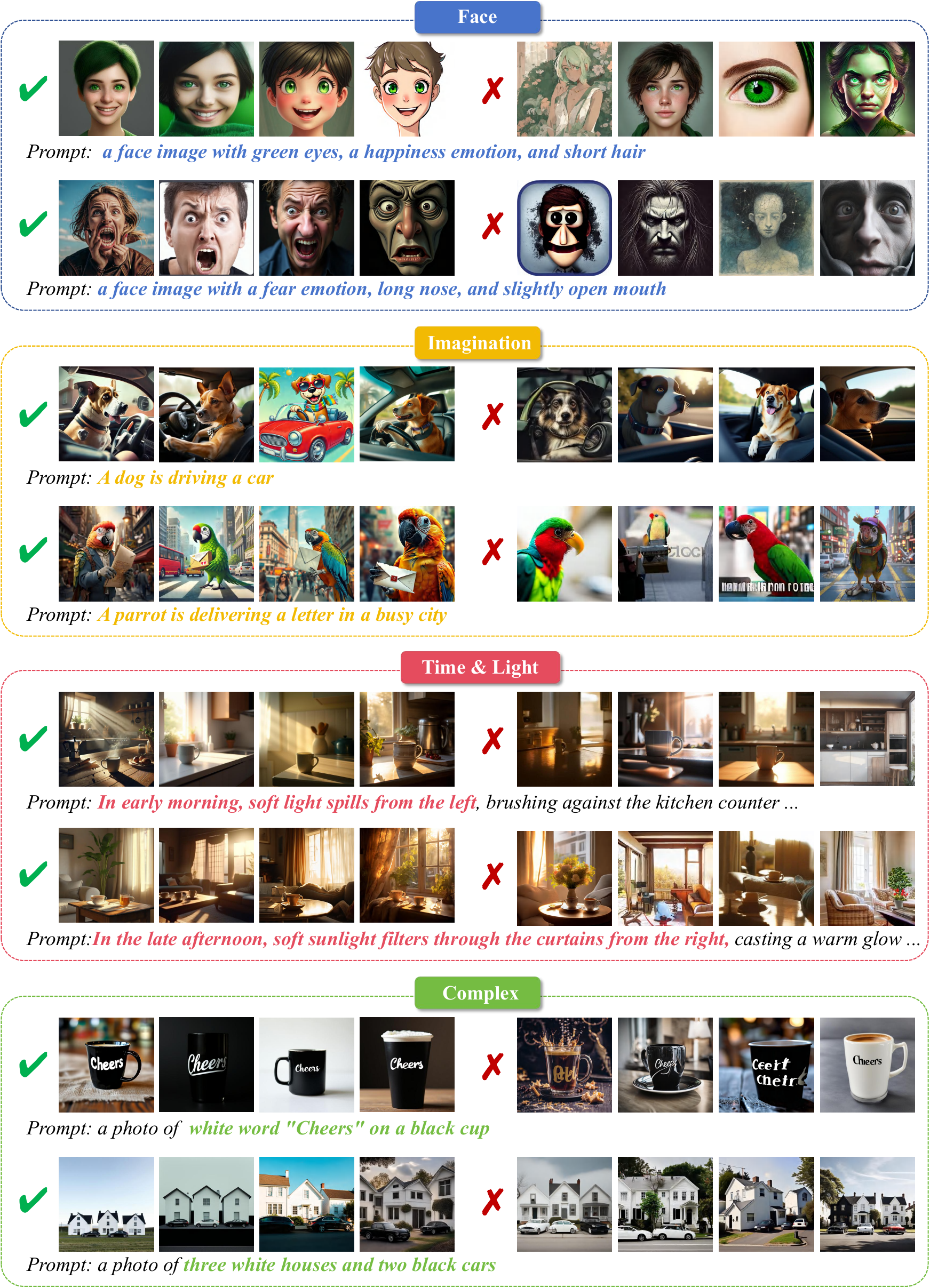}
 \vspace{-2mm}
	\caption{Examples for different task-specific challenges.}
 \vspace{-3mm}
	\label{task5}
\end{figure*}